\pdfoutput=1
\documentclass[10pt,twocolumn,letterpaper]{article}
\usepackage[pagenumbers]{edpreprint}
\usepackage{graphicx}
\usepackage[T1]{fontenc}
\usepackage[utf8]{inputenc}
\usepackage{microtype}
\usepackage{mathtools}
\usepackage{array}
\usepackage{tabularx}
\usepackage{tikz}
\definecolor{linkblue}{rgb}{0.21,0.49,0.74}
\usepackage{colortbl}
\definecolor{rankA}{HTML}{F4A09A}%
\definecolor{rankB}{HTML}{F9C98E}%
\definecolor{rankC}{HTML}{FCEE9C}%
\newcommand{\rA}[1]{\cellcolor{rankA}#1}
\newcommand{\rB}[1]{\cellcolor{rankB}#1}
\newcommand{\rC}[1]{\cellcolor{rankC}#1}
\usepackage[pagebackref,breaklinks,colorlinks,allcolors=linkblue]{hyperref}
\hypersetup{pdfauthor={Karthik Sundar Senthilkumar, Christopher Thirgood, Simon Hadfield},pdftitle={EventDiffusion: Learning Native Event Stream Distribution from Videos}}

\title{EventDiffusion: Learning Native Event Stream Distribution from Videos}
\author{\begin{tabular}{@{}c@{\hspace{2em}}c@{\hspace{2em}}c@{}}
Karthik Sundar Senthilkumar$^{*}$ & Christopher Thirgood$^{*}$ & Simon Hadfield\\
{\tt\small ks02450@surrey.ac.uk} & {\tt\small ct00659@surrey.ac.uk} & {\tt\small s.hadfield@surrey.ac.uk}\\[4pt]
\multicolumn{3}{c}{University of Surrey}
\end{tabular}}

\begin{document}
\twocolumn[{%
\renewcommand\twocolumn[1][]{#1}%
\maketitle
\captionsetup{type=figure}
\centering
\includegraphics[width=\textwidth]{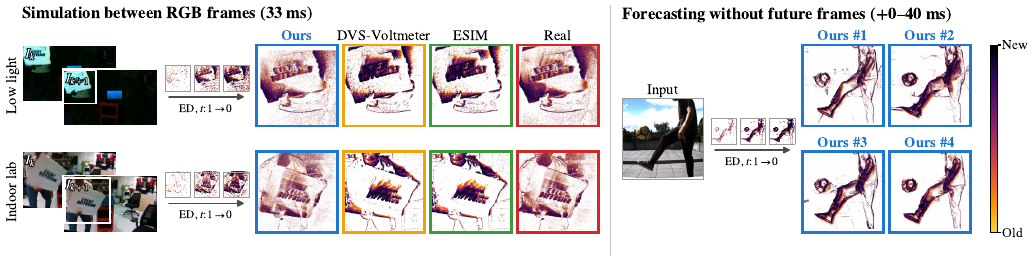}
\caption{EventDiffusion generates asynchronous event streams between RGB frames and beyond the last frame. \textbf{Left:} simulation between frames $I_k$ and $I_{k+1}$. Pixels in the boxed region are coloured by their latest event time in the 33\,ms interval. Above the first arrow: EAD's reverse diffusion of event counts in that region. \textbf{Right:} four forecasts of the first interval after $I_0$ on BS-ERGB, coloured by event time.}
\label{fig:teaser}
}]
{\renewcommand{\thefootnote}{*}\footnotetext{Equal contribution.}}
\begin{abstract}
Event cameras capture asynchronous brightness changes at high temporal
resolution, but scarce recordings constrain data-driven research.
Sparse RGB observations cannot uniquely determine intervening events since identical frames may correspond to streams with different activity, polarities, and timing.
Existing simulators largely rely on prescribed sensor dynamics and estimated brightness trajectories, whose assumptions limit the diversity and fidelity of plausible event streams.
We introduce EventDiffusion, a hierarchical generative framework that directly learns the conditional distribution of native asynchronous event streams from real recordings.
EventDiffusion decomposes generation into
three probabilistic processes:
Event Activity Diffusion (EAD) learns stochastic
discrete event counts,
Event Spatial Allocation (ESA) distributes these
counts to individual pixels while preserving cardinality,
and Event Timing Diffusion (ETD) learns continuous
asynchronous timestamp distributions.
Beyond generating events between RGB frames, EventDiffusion forecasts individual future events without requiring future RGB observations.
The resulting streams can be aggregated at user-selected temporal resolutions while preserving their native timestamps.
We also introduce CEFS, an event-stream realism metric combining event activity and polarity with multiscale space-time structure and temporal fidelity.
Across datasets, EventDiffusion outperforms existing simulators and demonstrates robustness across cameras and resolutions.
\end{abstract}

\section{Introduction}
\label{sec:intro}

Event cameras asynchronously report local log-intensity changes exceeding a contrast threshold.
Their microsecond-scale temporal resolution and high dynamic range suit fast motion and challenging illumination~\cite{gallego2022survey}.
However, acquiring large, synchronized RGB-event datasets is costly, motivating realistic event simulation.

Generating events from sparse RGB observations is fundamentally underdetermined.
Distinct unobserved motion and visibility trajectories can produce identical frames but different event activity, polarities, and timestamps.
Existing simulators address this ambiguity through prescribed sensor dynamics, interpolated or reconstructed brightness trajectories, and learned aggregated event representations~\cite{rebecq2018esim,hu2021v2e, lin2022voltmeter,zhang2024v2ce,thirgood2026tides}.
While pragmatic, these assumptions are often too simplistic for real scenes, making sim-to-real transfer challenging.

This motivates modelling a distribution over event streams rather than committing to a single realization. The same formulation also supports forecasting. Given only past observations, the model must predict not just future event activity, but also where events occur and when they fire, without access to future RGB frames.

To this end, we introduce EventDiffusion, a hierarchical generative framework that learns a conditional distribution over native asynchronous events from paired RGB and event-camera recordings.
Event generation involves predicting discrete event activity, allocating it across sensor pixels, and modelling continuous firing times. EventDiffusion addresses these through Event Activity Diffusion,
Event Spatial Allocation, and Event Timing Diffusion, respectively. This factorization preserves event counts and polarities while learning the spatial and temporal distributions of individual measurements.
Our approach also allows us to forecast complete asynchronous event streams directly from a single RGB image. Rather than predicting a future event frame or binned representation, it samples individual events and timestamps, allowing the same forecast to be examined at different temporal resolutions.
Evaluating event streams requires measuring activity and timing as well as spatial structure. PECS~\cite{han2024pecs} can overlook excess events near real ones, while EQS~\cite{chanda2025eqs} is insensitive to some count errors. For example, an empty output receives an EQS of $0.94$ or higher, despite missing all real activity. We introduce CEFS, a decomposable metric that compares count, polarity, space-time allocation, and timing against a paired recording. It provides a common evaluation for EventDiffusion and both learned and physics-based simulators.
We evaluate EventDiffusion under native and sparsely sampled RGB inputs, against simulators that also receive interpolated frames. Our experiments assess event fidelity, temporal statistics, and generalization across DSEC~\cite{gehrig2021dsec}, EDS~\cite{hidalgo2022eds}, BS-ERGB~\cite{tulyakov2022timelenspp}, and CEAR~\cite{zhu2024cear}.
We also evaluate forecasting against real future event streams.
Our contributions are:
\begin{enumerate}
    \item The first learned diffusion model of the RGB-conditioned distribution over asynchronous event streams.
    \item A hierarchical generation approach that models integer event counts, allocates them to pixels while preserving their totals, and generates continuous timestamps.
    \item Causal Forecasting of native asynchronous event streams, generating individual future events and their timestamps without access to future RGB observations.
    \item CEFS, a decomposable,
    multiscale metric of event count, polarity,
    spatial, temporal, and joint space--time fidelity.
\end{enumerate}

\section{Related Work}
\label{sec:related}

\subsection{Reconstruction-Based Event Simulation}
Many event simulators apply explicit sensor-response models to brightness trajectories. ESIM~\cite{rebecq2018esim} samples scene intensity through motion-adaptive rendering. For recorded video, VID2E~\cite{gehrig2020vid2e} interpolates intermediate RGB frames before simulation, but its events depend on the interpolation quality and a single interpretation of the unobserved scene.
Subsequent methods refine the sensor response to a
supplied brightness trajectory.
v2e~\cite{hu2021v2e} incorporates finite bandwidth, threshold variation, leakage, and noise, while DVS-Voltmeter~\cite{lin2022voltmeter} models stochastic pixel-voltage dynamics.
Characterization-based simulators~\cite{joubert2021characterized} incorporate measured hardware behaviour, and SENPI~\cite{greene2025senpi}
provides a differentiable sensor simulation framework.
FracEvent~\cite{chen2026fracevent} further models fractional-relaxation voltage dynamics with retained pixel memory and continuous threshold-crossing times.
These approaches improve sensor realism and temporal fidelity.
Scene-based methods improve the underlying brightness trajectory. PECS~\cite{han2024pecs} models optical and multispectral image formation from 3D scenes, while TIDES~\cite{thirgood2026tides} reconstructs dynamic Gaussian scenes to derive continuous-time intensity changes with visibility and sensor readout effects. Both rely on a specified or reconstructed scene and sensor response. EventDiffusion instead learns event-stream distributions from paired observations without scene reconstruction or intermediate brightness trajectories.
\begin{figure*}[t]
\centering
\makebox[\textwidth]{\includegraphics[width=\linewidth]{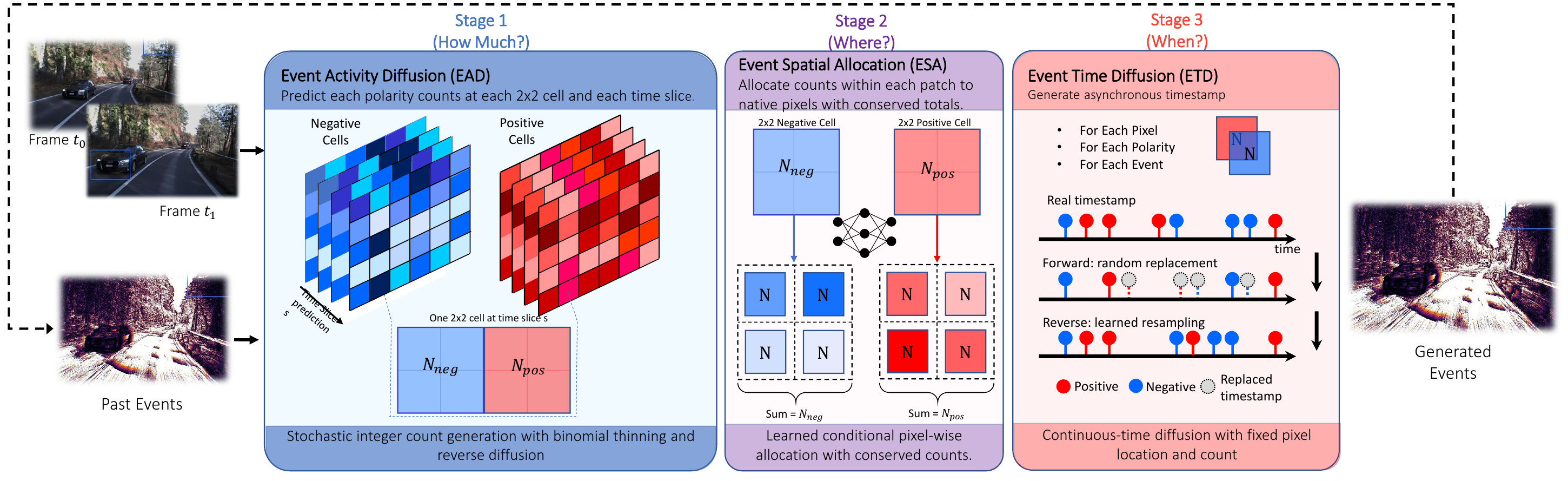}}
\caption{Three conditional models, one event stream. RGB enters EAD and ESA directly. ETD sees their activity/allocation outputs and event context, not a separate RGB encoder. Generated events are then given as the past events for the next pair of frames.}
\label{fig:ed_architecture}
\end{figure*}

\subsection{Learned Event Representations}
Learning-based methods differ in their output representations. EventGAN~\cite{zhu2019eventgan} learns image-conditioned event volumes using adversarial and consistency losses. V2CE~\cite{zhang2024v2ce} synthesizes event voxels before recovering continuous timestamps, while V2V~\cite{lou2025v2v} generates compact voxels for efficient storage and training.

Other generative methods broaden the conditioning modalities. Text-to-Events~\cite{ott2024texttoevents} combines diffusion with an event-frame autoencoder, while ControlEvents~\cite{hu2026controlevents} transfers image-diffusion priors to semantic and pose conditioning. Both use temporally aggregated representations. Texvent~\cite{wang2026texvent} generates video before interpolation and physics-based simulation to produce asynchronous events. These methods do not learn event timing conditioned on allocated event marks.

Diffusion has also been applied directly to event
clouds, albeit under different generation objectives.
Zhang \emph{et al.}~\cite{zhang2024completion}
formulate sparse event completion as diffusion-based
spatiotemporal point-cloud refinement, reconstructing
missing events from partial recordings.
Unlike RGB-conditioned simulation, this setting
assumes observed event measurements and targets
recovery of an incomplete event stream.
More generally, Point Set Diffusion~\cite{luedke2025psd} models random finite
point sets through thinning and noise superposition,
supporting variable-cardinality generation.
CountsDiff~\cite{soatto2026countsdiff} develops
diffusion processes on the natural numbers for
discrete count distributions.
Event-camera simulation requires integrating these mechanisms with polarity-aware activity, count-conserving spatial allocation, and fine-grained temporal dependencies. EventDiffusion addresses these coupled requirements through a hierarchical event-specific generative model.

Event forecasting predicts future event measurements from past observations.
E-Motion~\cite{wu2024emotion} adapts video diffusion to forecast future event representations from event history.
E-TIDE~\cite{sen2026etide} instead uses a lightweight deterministic network to predict temporally binned, polarity-separated event occurrence maps.
These approaches forecast aggregated representations, whereas EventDiffusion generates individual events with locations, polarities, and timestamps.

\subsection{Measuring Asynchronous Event Fidelity}
PECS~\cite{han2024pecs} compares events in space, time, and polarity using Chamfer and Gaussian distances. Its many-to-one matching can mask timing, count, and structural errors. EQS~\cite{chanda2025eqs} compares event-trained transformer features but depends on temporally aggregated inputs. TIDES~\cite{thirgood2026tides} adds timestamp-artifact diagnostics.
No prior work has proposed a unified criterion covering all aspects of event realism.
We introduce CEFS to evaluate these aspects of event fidelity, using reference-derived scale weights to reduce the influence of unreliable resolutions.

\section{Method}
\label{sec:ed_method}
As shown in Figure~\ref{fig:ed_architecture}, EventDiffusion models events
through a factorized distribution over plausible event streams. Rather than
jointly generating heterogeneous event attributes, our framework decomposes generation into
three conditional processes. First, EAD predicts stochastic event activity, next ESA resolves
this activity to individual sensor pixels, and finally ETD generates their asynchronous timestamps.

\subsection{Event Activity Diffusion}
\label{sec:ed_ead}

The number of events within a slice is always integer-valued.
EAD therefore models the joint event activity tensor through discrete count
diffusion \cite{soatto2026countsdiff}. Between two RGB observations at times $a$ and $b$,
we divide the interval $[a,b)$ into $S$ equal temporal slices of duration $\Delta=(b-a)/S$.
Let $A_{sc\sigma}$ denote the nonnegative integer event count in slice
$s\in\{1,\ldots,S\}$, spatial cell $c$, and polarity $\sigma$.
Each cell contains a spatial patch of sensor pixels, and $\sigma=0,1$ denotes
OFF and ON events, respectively. We refer to the full count tensor as $A$.

The forward diffusion process thins the counts, keeping each event independently with probability $\pi_{\rm EAD}(\xi)$,
\begin{equation}
 \begin{aligned}
 X_{\xi,sc\sigma}
 &\sim\operatorname{Binomial}
 (A_{sc\sigma},\pi_{\rm EAD}(\xi)),\\
 \pi_{\rm EAD}(\xi)
 &=\cos^2\!\left(\frac{\pi\xi}{2}\right),
 \end{aligned}
 \label{eq:fire_forward}
\end{equation}
where $\xi \in [0,1]$ denotes diffusion step and $X_\xi$ is the corrupted count tensor at that step.
Unlike the case of Gaussian Diffusion, this preserves the nonnegative integer counts \cite{soatto2026countsdiff}.

The reverse process of EAD is a denoising process to predict the removed activity count $Y_\xi=A-X_\xi$.
The denoiser predicts $Y_\xi$ across all slices jointly through the use of convolutions and attention
across the slices. We provide several conditioning signals to the denoising process, such as
the RGB features from an encoder, optional generated event history preceding the interval, valid sensor support, slice position and
duration, and optional camera information. The predicted expected missing count $\widehat Y_{\xi,sc\sigma}$ is scaled based on the slice duration as
\begin{equation}
\widehat Y_{\xi,sc\sigma}
=
(1-\pi_{\rm EAD}(\xi))
\Delta\,\lambda_{d_{\rm cam}}
\exp\left[
f_\phi(X_\xi,\text{cond})_{sc\sigma}
\right].
\label{eq:fire_rate}
\end{equation}
Here $\lambda_{d_{\rm cam}}$ is the camera's learned
baseline event rate, and $\text{cond}$ denotes the conditioning inputs to
the denoising network $f_\phi$, with learned parameters $\phi$.
We train the denoiser using a count reconstruction error:
\begin{equation}
 \ell_{\rm count}(\hat y,y)
 =\hat y-y\log\hat y+y\log y-y,
\label{eq:fire_err}
\end{equation}
which penalizes differences between
predicted and actual missing counts $\hat y$ and $y$.

To enable consistency across different experimental settings, this error is rescaled by the ratio of the slice duration against a reference duration ($\Delta_{\rm ref}=50\,\mathrm{ms}$), masked to the supervised cells via $m_c$ and normalised by $\mathcal Z$ across cells, polarities, and slices.
The resulting loss is then
\begin{equation}
\begin{aligned}
 \tilde\ell_{\xi,sc\sigma}
 &=\frac{\Delta_{\mathrm{ref}}}{\Delta}\,
 \ell_{\rm count}(\widehat Y_{\xi,sc\sigma},Y_{\xi,sc\sigma}),\\
 \mathcal L_{\mathrm{EAD}}
 &=\mathbb E_\xi\!\left[
 \frac{-\pi'_{\rm EAD}(\xi)}{\mathcal Z}
 \sum_{s,c,\sigma} m_c\,\tilde\ell_{\xi,sc\sigma}\right],
\end{aligned}
\label{eq:fire_loss}
\end{equation}
where the schedule derivative $-\pi'_{\rm EAD}(\xi)$
weights the diffusion levels.

At inference, EAD reverses the thinning process,
starting from an empty count map and progressively
sampling integer-valued event activity.
The resulting ON/OFF counts are passed to ESA
for count-preserving pixel allocation.

\subsection{Event Spatial Allocation}
\label{sec:ed_esa}
EAD estimates the coarse event activity counts but leaves detailed
spatial structure unresolved. ESA learns the conditional
per-pixel allocation within a spatial cell, while preserving polarity-specific
counts exactly.

To model the spatial clumping and dispersion of real event streams, we use a Conway--Maxwell-binomial (CMB) distribution~\cite{kadane2016comb}:
\begin{equation}
 p_{\mathrm{ESA}}(m\mid n,\eta,\kappa)=
 \frac{\binom{n}{m}^{\kappa}\exp(\eta m)}
 {\sum_{j=0}^{n}\binom{n}{j}^{\kappa}\exp(\eta j)},
 \label{eq:ed_cmb}
\end{equation}
where $n$ is the number of events being split, $m \in \{0, 1, \dots n\}$ denotes
the count of the events that follow one side of a split, which then ensures that
the other side of the split gets $n - m$ events. This formulation helps to
preserve the total count of events in each cell. $\eta$ biases the split towards the first side and $\kappa$ controls
how uneven the split can be. To distribute the events to the pixels within a spatial cell of size $w\times w$, we apply this split hierarchically over $2\log_2 w$ levels, halving the region at each level. The split parameters are estimated by a fully convolutional U-Net conditioned on the RGB frames.

Neighbouring cells need to have each other's context to preserve spatial structure such as edges. Therefore,
we provide the placed events in the neighbouring cells as conditioning to the ESA network. During training, a random set of cells is given to the network as
already allocated, and we calculate the loss over the rest. More formally, for a slice $s$, cell $c$ and polarity $\sigma$,
$A_{sc\sigma}$ denotes the total counts of events. In each training iteration, the allocations of
a random set of cells $\mathcal R$ are given to the network and the remaining cells' allocations are predicted. Let $\ell_{sc\sigma}$ denote the sum
of the log probabilities of the remaining cells, under Eq. \ref{eq:ed_cmb}, then the loss used to train our network is the
negative log likelihood per scored event,
\begin{equation}
 \mathcal L_{\mathrm{ESA}}
 =\mathbb E_{\mathcal R}\!\left[
 -\frac{1}{\mathcal Z_{\rm ESA}}\sum_{c,\sigma}
 m_c\,(1-r_c)\,\ell_{sc\sigma}\right],
 \label{eq:ed_esa_loss}
\end{equation}
where $r_c = 1$ for the cells that were already allocated, $m_c$ is the cell mask of Eq.~\eqref{eq:fire_loss}, and $\mathcal Z_{\rm ESA}$ is the number of scored events in the minibatch.

\subsection{Event Timing Diffusion}
\label{sec:ed_etd}
ESA defines where events occur spatially, but their temporal distribution is still highly complex. To ensure that we can model the refractory period, burstiness, and other quirks of
real event streams, we learn a joint distribution of all the event timestamps within a spatial patch and slice. Learning this joint distribution in one pass is hard, because each event's plausible times depend on the times of the others. Instead, we corrupt the timestamps step by step towards uniform noise and learn to reverse the corruption. At each reverse step, the model decides which events already hold plausible times, keeps them, and redraws the rest. Constraints between events, such as a refractory gap after a pixel's last event, are then imposed against events already in place.

To this end, ETD keeps the pixel and polarity of each event from ESA and samples only its
normalized time within the slice. Following Point Set
Diffusion~\cite{luedke2025psd}, the forward process replaces each real time
$\tau_{0,i}\in[0,1)$ with a uniform time drawn once per event, and replaces more
of them at later diffusion steps:
\begin{equation}
 \begin{aligned}
 \tau_{\xi,i}&=z_{\xi,i}\,\tau_{0,i}+\big(1-z_{\xi,i}\big)\,U_i,\\
 z_{\xi,i}&\sim\operatorname{Bernoulli}\big(\pi_{\rm EAD}(\xi)\big),
 \end{aligned}
 \label{eq:ed_etd_forward}
\end{equation}
where $\xi\in\{1/T,\dots,1\}$ is the diffusion step, $U_i\sim\mathcal U[0,1)$ is
the event's noise time, $z_{\xi,i}=1$ marks an event that holds its real
timestamp, and $\pi_{\rm EAD}$ is the cosine schedule of Eq.~\eqref{eq:fire_forward}.
A reverse step from $\xi$ to $\xi'$ undoes part of this corruption. A
network predicts the probability $q_i$ that event $i$ holds its real time, and
we sample $\widehat z_i\sim\operatorname{Bernoulli}(q_i)$ for every event.
Events with $\widehat z_i=1$ are judged real and keep their times. Each event
with $\widehat z_i=0$ is judged noise and is selected for timestamp replacement
with probability $\beta_{\xi\to\xi'}$, where
$\beta_{\xi\to\xi'}=\big(\pi_{\rm EAD}(\xi')-\pi_{\rm EAD}(\xi)\big)/\big(1-\pi_{\rm EAD}(\xi)\big)$.
Events not selected for replacement keep their current times.

For each selected event, we sample a replacement timestamp from a learned
density $p_i$. This density has to reproduce three kinds of timing structure. First, the activity of a patch varies smoothly over
time, often periodically, for example under lamp flicker.
$\rho^{\rm b}_i$ models this rhythm with a log-density represented as a bounded Fourier series. Second, events at nearby pixels fire in temporally correlated bursts as edges pass. Therefore, we attend to the tokens in the patch and in the same slice, and pick the $K$ events with the highest attention scores. We refer to these as the anchor events ($\mathcal A_i$). We then add a logistic kernel $\rho_{ij}^a$ to these anchors, centered at the
current time $\tau_j$ of the event (where $\tau_j$ denotes
the corrupted time $\tau_{\xi,j}$ in training, and the current time in the
reverse chain at inference). This increases the probabilities of the events happening at the same timestamp as their anchors.

The third sensor property we attempt to model is that due to the nature of their readout mechanism, event camera
sensors often give events along a row or column band the same microsecond timestamp, even when
they lie far apart spatially. Similar to the anchor events, we thus attend to the events in the same sensor rows and columns
$\mathcal C_i$. For each event $l\in\mathcal C_i$, we add a uniform density $\rho^{\rm t}_l$ over its
microsecond bin (taken from $\tau_l$) weighted by the attention softmax $\gamma_{il}$. This
increases the probability that events share the same microsecond.

Mixing these
three temporal behaviour components gives
\begin{equation}
 \begin{aligned}
 \tilde p_i(\tau)={}&\omega^{\rm b}_i\,\rho^{\rm b}_i(\tau)
 +\sum_{j\in\mathcal A_i}\omega^{\rm a}_{ij}\,\rho^{\rm a}_{ij}(\tau)\\
 &+\omega^{\rm t}_i\sum_{l\in\mathcal C_i}\gamma_{il}\,\rho^{\rm t}_l(\tau),
 \end{aligned}
 \label{eq:ed_density}
\end{equation}
where the nonnegative weights $\omega^{\rm b}_i$, $\omega^{\rm a}_{ij}$ and
$\omega^{\rm t}_i$ of the three components sum to one.
This mixture helps model temporal and spatial correlations that encourage firing at particular timestamps. The sensor also has a final temporal behaviour that suppresses particular timestamps.

A pixel that has just fired rarely fires
again during its refractory period, so the density function should be suppressed
near the times of its other events. Let $\mathcal P_i$ denote the other events (of either polarity) at
the pixel of event $i$. Around the current time $\tau_m$ of each such event
$m\in\mathcal P_i$, we suppress the probability density by a fraction
$h_{im}(\tau)=\alpha_{im}\exp[-(|\tau-\tau_m|\Delta/\vartheta_{im})^{\nu_{im}}]$,
where $\alpha_{im}$ controls the suppression strength, $\vartheta_{im}$ its
temporal extent, and $\nu_{im}$ its sharpness. The slice duration $\Delta$ converts the normalized timestamp difference
into physical time.
For each event $i$, a learned network predicts two sets of suppression parameters
$(\alpha,\vartheta,\nu)$, one for other events at the pixel with the same polarity
and one for those with the opposite polarity.
We weigh each such refractory suppression by $v_m\in[0,1]$, our confidence in the time of
event $m$. Combining this across all events
gives the same-pixel inhibition factor
\begin{equation}
 g_i(\tau)=\prod_{m\in\mathcal P_i}
 \Big(1-v_m\,h_{im}(\tau)\Big).
 \label{eq:ed_inhibition}
\end{equation}
Putting all of these behaviours together, we sample each
event's new time from $p_i(\tau)=\tilde p_i(\tau)\,g_i(\tau)/Z_i$, where $Z_i$
normalizes $p_i$.

To implement this, a transformer predicts $q_i$ and the parameters of $\tilde p_i$ and $g_i$ for
all events of a patch at once. Event tokens also carry neighbour features, the time gaps to other events at the same pixel. We train it on real events with corrupted times, combining
the binary cross-entropy $\ell_{\rm BCE}$ of the judgments with the negative
log-likelihood of the corrupted events' real times,
\begin{equation}
 \begin{aligned}
 \mathcal L_{\rm ETD}=\mathbb E_{\xi}\Big[
 &\frac{1}{N}\sum_{i}\ell_{\rm BCE}\big(q_i,z_{\xi,i}\big)\\
 &-\frac{1}{N_0}\sum_{i:\,z_{\xi,i}=0}\log p_i(\tau_{0,i})\Big],
 \end{aligned}
 \label{eq:ed_etd_loss}
\end{equation}
where $N$ is the total number of events, $N_0$ is the number whose timestamps
were corrupted ($z_{\xi,i}=0$), and the expectation is over uniformly sampled
diffusion steps $\xi$. We use techniques such as Diffusion Forcing~\cite{chen2024diffusionforcing} when training
to improve consistency when forecasting.  More details about inference
and training are provided in the supplementary materials.

\section{Measuring Event-Stream Fidelity}
\label{sec:cefs}
Event-stream fidelity depends jointly on activity, polarity, spatial structure, and timing, regardless of how the stream is generated. As discussed in Sec.~\ref{sec:related}, existing geometric and feature-based metrics capture complementary aspects of this fidelity. We introduce the Composite Event Fidelity Score (CEFS) to evaluate these properties together while retaining interpretable components.

CEFS compares generated events with a paired real recording over the same interval and sensor area, without count matching or realignment. Let $g$ and $r$ index the generated and recorded streams, respectively, as used in the component definitions below. CEFS requires no trained network or fitting to generated events. Scale weights depend only on the recording, and the two coefficients are fixed using corrupted real recordings (supplement). We use CEFS alongside existing metrics to evaluate EventDiffusion and the simulator baselines.

With $N_e^\sigma$ the number of polarity-$\sigma$ events of stream
$e\in\{r,g\}$ and $N_e=N_e^0+N_e^1$,
\begin{align}
D_{\rm count} &= 1-\exp\!\left[-\tfrac12\left(\log_2(N_g/N_r)\right)^2\right],\nonumber\\
D_{\rm pol} &= \left|\frac{N_r^1}{N_r}-\frac{N_g^1}{N_g}\right|.
\label{eq:cefs_activity}
\end{align}
$D_{\rm count}$ penalizes over- and under-generation symmetrically: generating twice or half the true number of events gives the same discrepancy.
$D_{\rm pol}$ measures the difference between the streams' ON-event fractions. Both discrepancies are zero for agreement.

For each polarity, events are trilinearly splatted into space--time
histograms at a fixed set of bin sizes $k$ (Sec.~\ref{sec:exp_setting}).
Each histogram is normalized to a distribution $p^e_{k,\sigma}$.
The streams are then compared by the
squared Hellinger distance
\begin{equation}
h_{k,\sigma}=1-\textstyle\sum_j\sqrt{p^r_{k,\sigma,j}\,p^g_{k,\sigma,j}} .
\label{eq:cefs_structure_unweighted}
\end{equation}
In a sparse window, fine bins hold so few events that even a second
recording of the scene would not match due to sensor noise. Bin sizes at which perturbed
copies of the real recording (random halves, block resamples, sub-pixel and $10\,\mu$s
shifts) already disagree with $r$ therefore receive low weight
$\widetilde w_k$ (supplement). Weighting each polarity by its share of $r$ gives
\begin{equation}
D_{\rm ST}=\Big[\textstyle\sum_k\widetilde w_k\sum_{\sigma}
\frac{N_r^\sigma}{N_r}h_{k,\sigma}\Big]^2 .
\label{eq:cefs_structure}
\end{equation}

$D_{\rm async}$ quantizes the timestamps of both streams at several clock resolutions and
penalizes timestamp sharing in $g$ beyond that in $r$ (i.e., excess
batching).
$D_{\rm temp}$ measures misplaced activity within the window as the
one-dimensional Wasserstein distance between the per-polarity event-time
distributions, divided by the window length plus $5\,$ms (supplement).

The discrepancies overlap, since a misplaced burst raises both $D_{\rm ST}$
and $D_{\rm temp}$.
To reduce the effect of double counting, we therefore keep the larger
discrepancy and add a bounded share of the smaller,
\begin{equation}
B_\lambda(x,y)=\max(x,y)+\lambda\min(x,y)\,[1-\max(x,y)],
\label{eq:cefs_blambda}
\end{equation}
first over timing and then with structure,
\begin{align}
D_{\rm timing}&=B_{\lambda_T}(D_{\rm async},D_{\rm temp}),\nonumber\\
D_{\rm core}&=B_{\lambda_C}(D_{\rm ST},D_{\rm timing}),\nonumber\\
\mathrm{CEFS}&=(1-D_{\rm count})(1-D_{\rm pol})(1-D_{\rm core}).
\label{eq:cefs_final}
\end{align}
CEFS lies in $[0,1]$ and is $1$ for an identical stream.
We also report shape fidelity $S=(1-D_{\rm pol})(1-D_{\rm core})$, so that
$\mathrm{CEFS}=(1-D_{\rm count})S$ separates how many events are produced from how they
are distributed. CEFS will be released as a standalone package to support evaluation of future event generation models.

\section{Experiments}
\label{sec:experiments}

\subsection{Experimental Settings}
\label{sec:exp_setting}
We evaluate test sets in DSEC~\cite{gehrig2021dsec}, EDS~\cite{hidalgo2022eds},
BS-ERGB~\cite{tulyakov2022timelenspp}, and CEAR~\cite{zhu2024cear}
with recording-level splits. All the baselines are calibrated separately for each dataset.
We evaluate two versions of EventDiffusion. In EventDiffusion-G, each of EAD, ESA and ETD is a single checkpoint shared by all datasets and, in EventDiffusion-FT, all three are fine-tuned for each dataset.
All methods in the simulation experiments use only RGB frames as inputs.
We compare Native RGB frames, SSM-10$\times$
(the same frames plus fixed $10\times$ Super SloMo~\cite{jiang2018superslomo} interpolation), and
Gap-5$\times$ (every fifth native frame) on identical reference-event
windows. Preprocessing, dataset eligibility, and input requirements are in
the supplement. We report CEFS (Sec.~\ref{sec:cefs}), EQS~\cite{chanda2025eqs}, and
PECS distances, alongside temporal diagnostics; full metric definitions are in
the supplement.
CEFS uses spatial bin sizes of $\{1,2,4,8,16\}$\,px and temporal bin sizes of
$\{10,50,100,500\}\,\mu$s and $\{1,5\}$\,ms (30 combinations).
Unless stated otherwise, scores are computed on the test windows
over complete valid $32{\times}32$\,px patches. Each sequence is averaged over
its windows, then each dataset is averaged over its sequences. In the result
tables, red, orange and yellow cells mark
the first, second and third best result in each column.
All models are trained on a single NVIDIA RTX 3090 GPU
using Adam for 100K iterations each.
EAD, ESA, and ETD are trained independently using their
respective objectives, with self-conditioning fine-tuning
to mitigate exposure bias.
Further implementation details are provided in the supplementary material.

\subsection{Event-Stream Fidelity under Native RGB}
\label{sec:main_results}
Both checkpoints lead Table~\ref{tab:gt_fidelity} in CEFS on all four datasets.
EventDiffusion operates at the tested cameras' native resolutions without a fixed full-frame output size, whereas V2CE uses resolution-dependent preprocessing. CEFS exposes count and timing errors that EQS and PECS can obscure. Table~\ref{tab:cadence} provides the matched-input comparison, holding reference events fixed. We use native RGB for EventDiffusion because interpolated frames degraded its performance compared with training on downsampled images.
\begin{table*}[t]
\centering\scriptsize
\setlength{\tabcolsep}{1.5pt}
\caption{Event fidelity. Simulators use $10\times$ Super SloMo inputs; V2CE and both EventDiffusion checkpoints use native RGB frames. EventDiffusion: mean ${\pm}$ standard deviation over three sampling seeds. P-CD: PECS Chamfer distance.}
\label{tab:gt_fidelity}
\begin{tabular*}{\textwidth}{@{\extracolsep{\fill}}l*{12}{c}@{}}
\toprule
& \multicolumn{3}{c}{\textbf{DSEC}}
& \multicolumn{3}{c}{\textbf{EDS}}
& \multicolumn{3}{c}{\textbf{BS-ERGB}}
& \multicolumn{3}{c}{\textbf{CEAR}} \\
\cmidrule(lr){2-4}\cmidrule(lr){5-7}\cmidrule(lr){8-10}\cmidrule(lr){11-13}
\textbf{Simulator}
& CEFS$\uparrow$ & EQS$\uparrow$ & P-CD$\downarrow$
& CEFS$\uparrow$ & EQS$\uparrow$ & P-CD$\downarrow$
& CEFS$\uparrow$ & EQS$\uparrow$ & P-CD$\downarrow$
& CEFS$\uparrow$ & EQS$\uparrow$ & P-CD$\downarrow$ \\
\midrule
ESIM
& 0.575 & \rC{0.977} & \rC{3.74} & 0.618 & 0.983 & \rC{6.29} & \rC{0.472} & \rB{0.985} & \rC{5.18} & 0.363 & 0.971 & 13.2 \\
v2e
& 0.324 & 0.976 & 9.77 & 0.310 & 0.983 & 31.8 & 0.302 & 0.981 & 11.4 & 0.133 & 0.964 & 37.8 \\
IEBCS
& 0.415 & 0.973 & 9.38 & 0.404 & 0.980 & 27.0 & 0.364 & 0.976 & 9.29 & 0.182 & 0.963 & 37.5 \\
DVS-Voltmeter
& \rC{0.576} & 0.974 & 3.86 & 0.611 & 0.981 & 6.40 & 0.468 & 0.980 & 5.46 & 0.394 & \rC{0.974} & \rC{12.5} \\
V2CE
& 0.502 & 0.971 & 4.77 & \rC{0.634} & \rB{0.991} & 6.72 & 0.456 & \rC{0.983} & 10.0 & \rC{0.461} & 0.969 & 14.8 \\
FracEvent
& 0.554 & 0.974 & 3.85 & 0.553 & 0.979 & 6.38 & 0.451 & 0.980 & 5.30 & 0.325 & 0.967 & 13.0 \\
\midrule
\textbf{EventDiffusion-G}
& \rB{0.636{\tiny$\pm$0.002}} & \rB{0.989{\tiny$\pm$0.000}} & \rA{3.56{\tiny$\pm$0.02}} & \rB{0.635{\tiny$\pm$0.001}} & \rC{0.989{\tiny$\pm$0.000}} & \rB{6.05{\tiny$\pm$0.02}} & \rB{0.550{\tiny$\pm$0.003}} & 0.980{\tiny$\pm$0.001} & \rB{5.16{\tiny$\pm$0.06}} & \rB{0.476{\tiny$\pm$0.004}} & \rB{0.984{\tiny$\pm$0.000}} & \rA{12.2{\tiny$\pm$0.1}} \\
\textbf{EventDiffusion-FT}
& \rA{0.684{\tiny$\pm$0.003}} & \rA{0.991{\tiny$\pm$0.000}} & \rB{3.62{\tiny$\pm$0.03}} & \rA{0.658{\tiny$\pm$0.002}} & \rA{0.992{\tiny$\pm$0.000}} & \rA{5.77{\tiny$\pm$0.03}} & \rA{0.592{\tiny$\pm$0.002}} & \rA{0.990{\tiny$\pm$0.000}} & \rA{4.88{\tiny$\pm$0.05}} & \rA{0.509{\tiny$\pm$0.003}} & \rA{0.986{\tiny$\pm$0.000}} & \rA{12.2{\tiny$\pm$0.0}} \\
\bottomrule
\end{tabular*}
\end{table*}

\begin{table}[t]
\centering\scriptsize
\setlength{\tabcolsep}{1.5pt}
\caption{Native RGB and Gap-5$\times$ (every fifth frame kept), mean over four datasets; EventDiffusion $\pm$ std over three seeds. CD: PECS Chamfer distance.}
\label{tab:cadence}
\begin{tabular*}{\columnwidth}{@{\extracolsep{\fill}}l*{6}{c}@{}}
\toprule
& \multicolumn{3}{c}{\textbf{Native RGB}} & \multicolumn{3}{c}{\textbf{Gap-5$\times$}} \\
\cmidrule(lr){2-4}\cmidrule(lr){5-7}
\textbf{Method} & CD$\downarrow$ & EQS$\uparrow$ & CEFS$\uparrow$
& CD$\downarrow$ & EQS$\uparrow$ & CEFS$\uparrow$ \\
\midrule
ESIM & \rC{7.15} & 0.971 & 0.460 & 9.18 & 0.969 & \rC{0.289} \\
v2e & 32.4 & 0.966 & 0.141 & 79.8 & 0.940 & 0.034 \\
IEBCS & 39.7 & 0.944 & 0.062 & 48.8 & 0.930 & 0.009 \\
DVS-Voltmeter & 7.18 & 0.968 & 0.461 & 9.31 & 0.964 & 0.288 \\
V2CE & 9.09 & \rC{0.978} & \rC{0.513} & 11.9 & \rC{0.971} & 0.280 \\
FracEvent & 7.35 & 0.969 & 0.422 & \rC{9.05} & 0.966 & \rC{0.289} \\
\midrule
\textbf{EventDiffusion-G} & \rB{6.73} & \rB{0.985} & \rB{0.574} & \rB{7.58} & \rB{0.978} & \rB{0.363} \\
& {\tiny$\pm$0.03} & {\tiny$\pm$0.000} & {\tiny$\pm$0.002} & {\tiny$\pm$0.02} & {\tiny$\pm$0.000} & {\tiny$\pm$0.002} \\
\textbf{EventDiffusion-FT} & \rA{6.60} & \rA{0.989} & \rA{0.611} & \rA{7.45} & \rA{0.983} & \rA{0.425} \\
& {\tiny$\pm$0.01} & {\tiny$\pm$0.000} & {\tiny$\pm$0.001} & {\tiny$\pm$0.01} & {\tiny$\pm$0.000} & {\tiny$\pm$0.002} \\
\bottomrule
\end{tabular*}
\end{table}

Under matched native RGB, EventDiffusion-FT and EventDiffusion-G achieve mean CEFS of $0.611$ and $0.574$, compared with $0.513$ for V2CE, the strongest baseline, and $0.460$ for ESIM. The shared EventDiffusion-G checkpoint also exceeds the strongest baseline without dataset-specific fine-tuning. SSM-10$\times$ gives ESIM and DVS-Voltmeter a denser brightness trajectory, while Gap-5$\times$ interpolation can misplace crossings and lose motion. V2CE is better with native-RGB inputs compared to interpolated frames.
Under Gap-5$\times$, EventDiffusion-FT still leads on all three metrics, with a CEFS of $0.425$ against $0.289$ for the best baselines. Its CEFS falls by about $30\%$ from native RGB, similar to FracEvent and less than ESIM, DVS-Voltmeter and V2CE ($37$--$45\%$). EventDiffusion-G is second on all three. This degradation comes from under-generated activity on several datasets and lost spatial structure on BS-ERGB.

Table~\ref{tab:timing} compares timestamp statistics against each camera's
paired recording rather than rewarding the absence of sensor artifacts.
Because real recordings can contain quantization and bursts, matching the
sensor's own statistics is more informative than merely avoiding timestamp
ties.
\begin{table}[t]
\centering\footnotesize
\setlength{\tabcolsep}{3pt}
\caption{Timing fidelity, native RGB, mean over four datasets; EventDiffusion $\pm$ std over three seeds.}
\label{tab:timing}
\begin{tabularx}{\columnwidth}{@{}>{\raggedright\arraybackslash}Xccc@{}}
\toprule
\textbf{Method}
& \shortstack{\textbf{Same-ts}\\$|\Delta_{\rm GT}|\downarrow$}
& \shortstack{\textbf{ISI-spike}\\$|\Delta_{\rm GT}|\downarrow$}
& \shortstack{\textbf{Fano}\\$|\log$ ratio$|\downarrow$} \\
\midrule
ESIM & 0.094 & 0.014 & 1.19 \\
v2e & 0.297 & \rC{0.009} & 8.26 \\
DVS-Voltmeter & 0.089 & 0.012 & \rC{1.18} \\
V2CE & \rC{0.078} & 0.012 & 2.84 \\
FracEvent & 0.232 & 0.017 & 1.29 \\
\midrule
\textbf{EventDiffusion-G} & \rA{0.0519{\tiny$\pm$0.0005}} & \rA{0.00558{\tiny$\pm$0.00025}} & \rB{0.904{\tiny$\pm$0.011}} \\
\textbf{EventDiffusion-FT} & \rB{0.0545{\tiny$\pm$0.0002}} & \rB{0.00854{\tiny$\pm$0.00042}} & \rA{0.840{\tiny$\pm$0.003}} \\
\bottomrule
\end{tabularx}
\vspace{-4mm}
\end{table}

\begin{figure}[t]
\centering
\includegraphics[width=0.8\columnwidth]{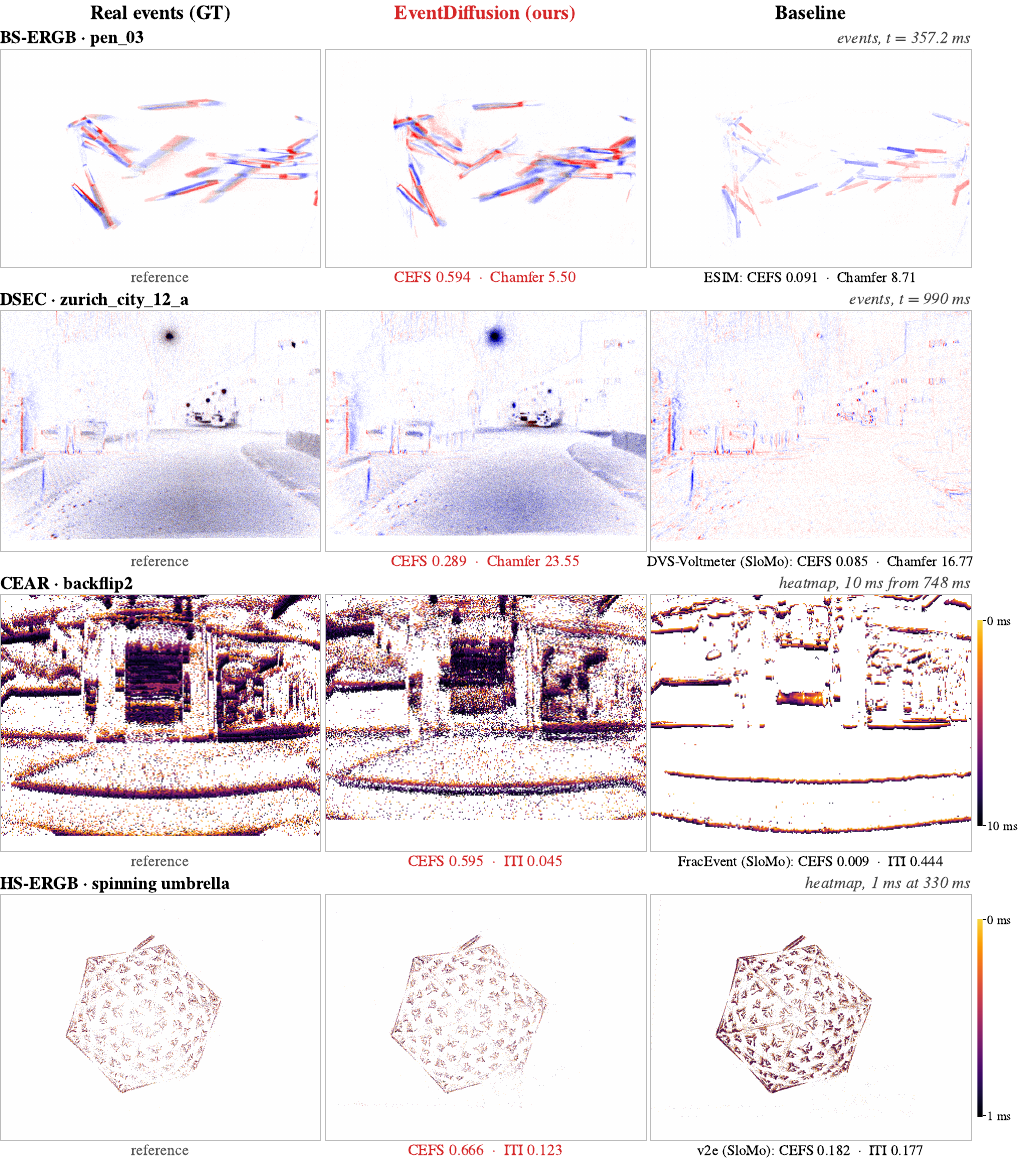}
\caption{Visual comparison between the baselines from binned
event frames and heatmapped event frames by timestamp}
\label{fig:qualitative_vis}
\end{figure}

EventDiffusion leads on all three diagnostics. ESIM and DVS-Voltmeter produce
continuous timestamps, but do not match sensor-specific inter-event intervals
and burst statistics as closely. ETD instead learns timing conditioned on
allocated events, including cross-pixel dependencies and same-pixel
inhibition. Fig.~\ref{fig:qualitative_vis} shows the structural and timing improvements over the baselines on four test recordings.
More qualitative results are provided in the supplementary.

\subsection{Asynchronous Event-Stream Forecasting}
\label{sec:forecasting_results}
We evaluate forecasting events over $[a,b)$ using a range of historical conditioning signals available at or before $a$: a single RGB frame, the preceding 20\,ms of events, multiple past RGB frames, and combinations of these. More details are provided in the supplementary material.

\subsubsection{Conditioning and horizon.}
Table~\ref{tab:forecast_conditioning} compares conditioning inputs across forecast horizons. The RGB and RGB-plus-event settings isolate the effect of recent events, while past RGB frames provide motion information.
Past events as the conditioning input give the best scores over short horizons (Table~\ref{tab:forecast_conditioning}) but fall off over later intervals, as the lower five-interval mean in Table~\ref{tab:forecast_results} shows. RGB frames reduce this decline over longer horizons.
To use past RGB frames, we first generate events between the past frames, then forecast from the current RGB frame and these generated events.
\begin{table}[t]
\centering\footnotesize
\caption{EventDiffusion's mean CEFS on BS-ERGB over the first 5--50\,ms of the forecast. CRPS: count CRPS of three forecasts over 25\,ms, pooled over sequences and divided by the real event count.}
\label{tab:forecast_conditioning}
\begin{tabular}{lccccc}
\toprule
Inputs & 5 ms & 10 ms & 20 ms & 50 ms & CRPS $\downarrow$ \\
\midrule
Single RGB & 0.290 & 0.313 & 0.328 & 0.324 & 0.485 \\
RGB + events & 0.449 & 0.461 & 0.458 & 0.407 & 0.365 \\
Past RGB & 0.409 & 0.427 & 0.430 & 0.395 & 0.388 \\
Past events & 0.524 & 0.534 & 0.522 & 0.434 & 0.250 \\
\bottomrule
\end{tabular}
\end{table}

\subsubsection{Comparison with forecasting baselines.}
Table~\ref{tab:forecast_results} groups the comparisons by input: a single RGB frame, past RGB frames, and event history. For event history, we compare against E-Motion~\cite{wu2024emotion}, which conditions on event-voxel frames, using its released checkpoint. For the RGB-prediction-plus-simulation baselines, SimVP~\cite{gao2022simvp,tan2022simvp} forecasts the future RGB frames from the two latest observed frames, which are then passed to an event simulator. We also report Cosmos Edge~\cite{nvidia2026cosmos3} as the video predictor, but it takes
5 RGB frames as input to generate the future frames.
CEFS applies only to native event streams, so E-Motion's voxel forecasts are scored with mIoU and F1 against the ground truth only. The evaluation procedure is detailed in the supplementary material.
EventDiffusion achieves the strongest performance in every comparison, including with only a single RGB frame or only past events as input. Voxel predictors such as E-Motion use fixed temporal aggregation, while RGB-prediction-plus-simulation pipelines depend on an estimated future brightness trajectory.
\begin{table}[t]
\centering\scriptsize
\caption{Forecasting comparison on the BS-ERGB test sequences. Scores are averaged across five frame
intervals (about 187\,ms). mIoU and F1 measure where events occur, in E-Motion's voxel
representation. CEFS applies only to native streams. }
\label{tab:forecast_results}
\resizebox{\columnwidth}{!}{%
\begin{tabular}{llcccc}
\toprule
Method & Input & Output & mIoU $\uparrow$ & F1 $\uparrow$ & CEFS $\uparrow$ \\
\midrule
\multicolumn{6}{l}{\textit{Single RGB frame}} \\
EventDiffusion & 1 RGB frame & Native stream & 0.259 & 0.390 & 0.262 \\
\midrule
\multicolumn{6}{l}{\textit{Past RGB frames}} \\
SimVP + ESIM & 2 RGB frames & Native stream & 0.230 & 0.359 & 0.133 \\
SimVP + DVS-Voltmeter & 2 RGB frames & Native stream & 0.238 & 0.369 & 0.139 \\
Cosmos Edge + ESIM & 5 RGB frames & Native stream & 0.167 & 0.274 & 0.127 \\
Cosmos Edge + DVS-Voltmeter & 5 RGB frames & Native stream & 0.170 & 0.277 & 0.132 \\
EventDiffusion & 2 RGB frames & Native stream & \rA{0.295} & \rA{0.433} & \rA{0.295} \\
\midrule
\multicolumn{6}{l}{\textit{Event history}} \\
E-Motion & Events (100\,ms) & Voxels & 0.221 & 0.329 & N/A \\
EventDiffusion & Events (20\,ms) & Native stream & \rC{0.274} & \rC{0.409} & \rC{0.273} \\
EventDiffusion & 1 RGB frame + events & Native stream & \rB{0.292} & \rB{0.426} & \rB{0.294} \\
\bottomrule
\end{tabular}%
}
\end{table}

\subsubsection{Native timing and temporal resolution.}
With identical forecasted events, ETD better matches the temporal statistics of the real stream than uniform or causal rate-based timestamp assignment, improving timing independently of count and location. The same forecast retains strong fidelity across temporal bin widths without resampling or changing its horizon. Detailed timing and aggregation results are in the supplement.
Figure~\ref{fig:forecast_qualitative} shows three EventDiffusion forecasts of the same input beside SimVP+ESIM, with each pixel coloured by the mean time of its events.
\begin{figure}[t]
\centering
\IfFileExists{figures/forecast_qualitative.pdf}{
  \includegraphics[width=0.9\columnwidth]{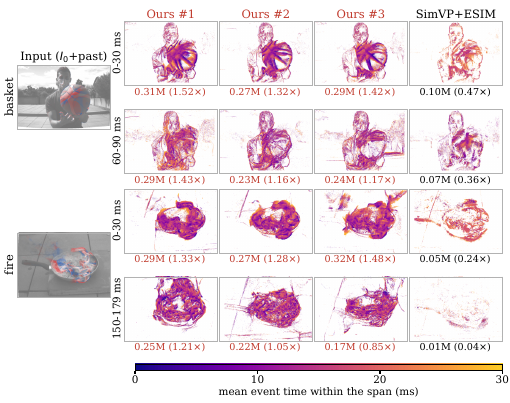}
}{}
\caption{Forecasting on BS-ERGB from $I_0$ and the preceding events (left): three EventDiffusion samples and
SimVP+ESIM over 30\,ms spans of the 180\,ms forecast (basket 60--90\,ms, fire the last 30\,ms). Pixels are coloured by
mean event time in the span (white: none); below, the event count and its ratio to the real stream's.}
\label{fig:forecast_qualitative}
\end{figure}

Analysis of Downstream Tasks and visualizations are provided in the supplementary material.

\subsection{EventDiffusion Ablations}
\label{sec:ablations}
Table~\ref{tab:abl_components} examines EAD and ETD. In EAD, generated event history improves count fidelity and temporal placement, indicating that previous activity helps predict how many events occur and when. Removing camera information has the largest effect on counts, raising $D_{\rm count}$ from $0.109$ to $0.191$. This suggests that camera conditioning helps the shared model adapt its event activity across cameras.

In ETD, removing neighbour features has the largest effect on CEFS and Fano error, showing the value of same-pixel timing gaps. Same-pixel inhibition improves timing fidelity, while anchor events help reproduce shared timestamps. Row and column context barely affects DSEC (removing it lowers $D_{\rm async}$) but matters on EvSLAM, as shown in the supplementary.
The base density alone increases Fano error, showing the value of event context. We include additional ablations in the supplementary.
\begin{table}[t]
\centering\scriptsize
\setlength{\tabcolsep}{0.9pt}
\caption{Ablations on DSEC test. EAD rows: EventDiffusion-G's EAD with EventDiffusion-FT's DSEC ESA and ETD. ETD rows: EventDiffusion-G's ETD given real per-pixel counts. $D_{\rm async}$ ${\times}10^5$; $D_{\rm temp}$, Same-ts ${\times}10^3$.}
\label{tab:abl_components}
\resizebox{\columnwidth}{!}{%
\begin{tabular}{@{}lccccccc@{}}
\toprule
\textbf{Variant}
& \shortstack{$D_{\rm count}$$\downarrow$}
& \shortstack{$D_{\rm ST}$$\downarrow$}
& \shortstack{$D_{\rm async}$$\downarrow$}
& \shortstack{$D_{\rm temp}$$\downarrow$}
& \shortstack{\textbf{Same-ts}$\downarrow$}
& \shortstack{\textbf{Fano}$\downarrow$}
& \shortstack{\textbf{CEFS}$\uparrow$} \\
\midrule
\multicolumn{8}{@{}l}{\textit{EAD, full pipeline}} \\
Full model                 & 0.109 & 0.257 & 4.00 & 11.1 & 24.8 & 1.35 & 0.638 \\
$-$ event history          & 0.135 & 0.263 & 2.38 & 12.3 & 28.2 & 1.39 & 0.613 \\
$-$ camera information     & 0.191 & 0.267 & 0.78 & 10.5 & 44.6 & 1.52 & 0.571 \\
\midrule
\multicolumn{8}{@{}l}{\textit{ETD, real event marks}} \\
Full ETD                   & 0 & 0.194 & 22.3 & 2.37 & 1.17 & 1.07 & 0.805 \\
$-$ anchor events          & 0 & 0.194 & 28.8 & 2.32 & 1.30 & 1.07 & 0.805 \\
$-$ same-pixel inhibition  & 0 & 0.195 & 25.3 & 2.51 & 1.22 & 1.09 & 0.804 \\
$-$ neighbour features     & 0 & 0.207 & 37.5 & 5.16 & 1.11 & 1.29 & 0.791 \\
$-$ row/column context     & 0 & 0.193 & 6.54 & 2.23 & 1.15 & 1.07 & 0.806 \\
Base density only          & 0 & 0.199 & 2.23 & 3.87 & 1.26 & 1.51 & 0.800 \\
\bottomrule
\end{tabular}%
}
\end{table}

\section{Conclusion}
\label{sec:conclusion}
EventDiffusion generates and causally forecasts native asynchronous event streams through three stages that model activity, spatial allocation, and continuous timing. This factorization makes their assumptions independently testable.
Future work could explore forecasting as a step towards predictive world models operating directly on asynchronous sensor measurements.

\label{sec:main_end}
{\small
\bibliographystyle{ieeenat_fullname}
\bibliography{references}
}

\clearpage
\setcounter{table}{0}
\setcounter{figure}{0}
\setcounter{equation}{0}
\renewcommand{\thetable}{S\arabic{table}}
\renewcommand{\thefigure}{S\arabic{figure}}
\renewcommand{\theequation}{S\arabic{equation}}
\renewcommand{\theHtable}{S\arabic{table}}
\renewcommand{\theHfigure}{S\arabic{figure}}
\renewcommand{\theHequation}{S\arabic{equation}}
\renewcommand{\topfraction}{0.9}
\renewcommand{\dbltopfraction}{0.9}
\renewcommand{\bottomfraction}{0.8}
\renewcommand{\textfraction}{0.07}
\renewcommand{\floatpagefraction}{0.8}
\renewcommand{\dblfloatpagefraction}{0.8}
\setcounter{topnumber}{3}
\setcounter{dbltopnumber}{2}
\setcounter{totalnumber}{4}
\maketitlesupplementary
\section*{Overview}
This supplementary material provides the datasets and evaluation protocol in Sec.~\ref{sec:supp-protocol}, followed by method, training and implementation details in Sec.~\ref{sec:supp-method} and the full CEFS definition in Sec.~\ref{sec:supp-cefs}. Section~\ref{sec:supp-results} presents additional results, Sec.~\ref{sec:supp-ablations} provides further ablations, and Sec.~\ref{sec:supp-qualitative} shows qualitative results. Section~\ref{supp:cefs_validation} gives additional CEFS validation.
\appendix
\section{Datasets and Evaluation Protocol}
\label{sec:supp-protocol}

\subsection{Training and Test Splits}
\label{sec:supp-splits}
In the main paper, we report the performance on DSEC~\cite{gehrig2021dsec}, EDS~\cite{hidalgo2022eds}, BS-ERGB~\cite{tulyakov2022timelenspp} and CEAR~\cite{zhu2024cear}.
EventDiffusion-G is one shared set of EAD, ESA and ETD trained on the training recordings of these four datasets together with HS-ERGB~\cite{tulyakov2021timelens} and EvSLAM.
EventDiffusion-FT continues the shared training of each stage on each dataset separately.

Every split is made at the recording level. All windows, crops and stereo sides of a recording stay in its recording's part. This ensures that we use different recordings for training and
test sets.
Table~\ref{tab:supp_splits} gives the counts.

\begin{table}[h]
\centering
\scriptsize
\setlength{\tabcolsep}{3pt}
\caption{Training and test splits.}
\label{tab:supp_splits}
\begin{tabular*}{\columnwidth}{@{\extracolsep{\fill}}llrrrrr@{}}
\toprule
Dataset & Split rule & Train & Val & Test & Win./rec. & Window \\
\midrule
DSEC & official & 41 & -- & 12 (12) & 20 & 50\,ms \\
EDS & unique scenes & 13 & -- & 2 (2) & 60 & 13--21\,ms \\
BS-ERGB & official & 48 & -- & 26 (21) & 25 & 36--49\,ms \\
CEAR & by location & 93 & -- & 16 (16) & 60 & 15--18\,ms \\
\midrule
HS-ERGB & by sequence & 12 & -- & 3 & 60 & 6.1--6.7\,ms \\
EvSLAM & by platform & 15 & -- & 3 & 30 & 33--34\,ms \\
\bottomrule
\end{tabular*}
\end{table}

\begingroup\raggedright
\paragraph{DSEC.}
We use DSEC's official test split of 12 sequences (\texttt{interlaken\_00\_a}, \texttt{interlaken\_00\_b}, \texttt{interlaken\_01\_a}, \texttt{thun\_01\_a}, \texttt{thun\_01\_b}, \texttt{zurich\_city\_12\_a}, \texttt{zurich\_city\_13\_a/b}, \texttt{zurich\_city\_14\_a/b/c} and \texttt{zurich\_city\_15\_a}). Eleven are recorded by day and \texttt{zurich\_city\_12\_a} at night. All 41 official training sequences are used for training, with no validation split.

\paragraph{EDS.}
EDS has no official split. We hold out \texttt{13\_airplane} and \texttt{15\_apartment\_day}, the two scenes that no other sequence shares, and train on the other 13 sequences. \texttt{10\_office} is excluded because its released events cover only 1\,s of its 66\,s.

\paragraph{BS-ERGB.}
We keep the release's split~\cite{tulyakov2022timelenspp}, with 48 sequences in \texttt{3\_TRAINING}, 19 in \texttt{2\_VALIDATION} and 26 in \texttt{1\_TEST} (\texttt{2\_VALIDATION/horse\_10} is empty). We train on \texttt{3\_TRAINING} only. The \texttt{2\_VALIDATION} sequences are used neither for training nor for testing. We score 21 of the 26 test sequences. The event files of \texttt{horse\_11}, \texttt{horse\_12}, \texttt{horse\_13}, \texttt{horse\_18} and \texttt{juggling\_06} are not time-sorted, so they fail our window index and are left out for every method.

\paragraph{CEAR.}
CEAR has no official split, so we split by location. A location is a recording name without its lighting condition and gait, and all recordings of a location fall in the same part, so no place seen in training is scored. The four test locations (\texttt{backflip/env8}, \texttt{indoor/classroom}, \texttt{outdoor/between-buildings} and \texttt{outdoor/grass1}, 16 recordings and 9\% of CEAR's event time) cover both event cameras (DVXplorer and DAVIS346) and every lighting condition (well lit, blinking, HDR, day and night) except indoor dark, whose six recordings all train. The other 93 recordings are used for training, with no validation split. \texttt{outdoor/downtown1\_day\_comb} is left out of both parts because its RGB--event clock offset could not be measured.

\paragraph{HS-ERGB.}
The HS-ERGB data we use are the release's 15 test sequences~\cite{tulyakov2021timelens}. The release has no validation split. We train on 12 of the sequences and hold out \texttt{close/fountain\_schaffhauserplatz\_02}, \texttt{close/spinning\_umbrella} and \texttt{far/sihl\_03} for testing.

\paragraph{EvSLAM.}
Each EvSLAM recording has two event cameras. We hold out three recordings, one each from the arm, drone and mecanum-car platforms (\texttt{arm\_hdr}, \texttt{drone\_circle\_norm} and \texttt{test}), with both cameras (the slider platform's two recordings both train), and train on the other 15 (30 camera streams).

\par\endgroup

\paragraph{Test windows.}
In every test recording we score a contiguous block of about 1\,s of native RGB frame intervals $[t_k,t_{k+1})$ nearest the recording's middle, given by one shared interval list that is identical for every method and RGB condition. This gives 240 DSEC, 120 EDS, 525 BS-ERGB and 960 CEAR windows. SSM-10$\times$ and Gap-5$\times$ outputs are scored on the same native intervals.

\paragraph{Compared methods.}
We compare ESIM~\cite{rebecq2018esim}, v2e~\cite{hu2021v2e}, IEBCS~\cite{joubert2021characterized}, DVS-Voltmeter~\cite{lin2022voltmeter}, V2CE~\cite{zhang2024v2ce} and FracEvent~\cite{chen2026fracevent} in Tables~1--3 of the main paper.
For each simulator, we calibrate their parameters. V2CE is fine-tuned on each dataset at its native resolution from the released checkpoint.

\paragraph{Visual-information conditions.}
Native RGB uses the recorded frames at their recorded times, after the per-recording frame--event offsets of Sec.~\ref{sec:supp-preprocessing}. SSM-10$\times$ adds fixed $10\times$ Super SloMo interpolation~\cite{jiang2018superslomo}. Gap-5$\times$ retains every fifth native RGB frame, so each gap $[t_k,t_{k+5})$ receives only its two end frames, while the reference event stream and the scored windows stay fixed.

\subsection{Preprocessing and Dataset Fixes}
\label{sec:supp-preprocessing}
Events stay on each event camera's own pixel grid, and the RGB frames are resampled onto that grid. Each dataset also gives a mask of the pixels where both events and RGB are valid, and scoring uses only complete valid $32{\times}32$\,px patches.

\paragraph{DSEC.}
The RGB and event cameras are about 4.4\,cm apart, so mapping the frames into the event grid with the calibrated rotation and rectification alone leaves near objects misaligned.
Therefore we use a depth-aware alignment. We estimate metric depth for each RGB frame with MoGe-2~\cite{wang2025moge2} and, for every event pixel, solve for the RGB pixel that sees the same 3D point, using the calibrated rotation and translation.
Pixels without a depth estimate are treated as far away, surfaces nearer than 1.5\,m are sampled at 1.5\,m, and pixels hidden from the RGB camera take their appearance from the nearby far side of the occluding edge.

\paragraph{EDS.}
EDS records with a beam splitter, so the RGB frames are mapped into the event grid by a depth-free warp through the release's calibration.
A per-recording registration correction and a frame--event clock offset are fitted by aligning frames with events, for every recording.

\paragraph{BS-ERGB and HS-ERGB.}
Both releases store events already warped into RGB image coordinates, BS-ERGB on a $1/32$-pixel lattice and HS-ERGB as floating-point positions.
The warp leaves curved lines and lattice points that can never receive an event. For each recording we instead rebuild an integer sensor lattice from the event positions that occur and a fitted coordinate map, and resample the frames onto it.
Events that do not fall on the rebuilt lattice are excluded. HS-ERGB is split into windows by timestamp, and spans where its event stream drops out are excluded.

\paragraph{CEAR.}
CEAR records with two event cameras, a DVXplorer ($320\times240$) and a DAVIS346 ($346\times260$), and a separate RGB camera with depth registered to the event camera.
We read each recording's event camera from its file header, since the file names do not identify it reliably. The RGB frames are warped into the event grid using the recorded depth (matched in time to each frame) and the calibrated event-to-RGB extrinsics.
Missing depth takes the nearest depth within 6\,px and is otherwise treated as far away, and frames with no depth within 20\,ms are treated as far. Both cameras' event-to-RGB extrinsics and every recording's clock offset are refined by aligning the frames with the events.

\paragraph{EvSLAM.}
EvSLAM records with a stereo pair of event cameras ($640\times480$) and a separate RGB camera, on two rigs with different RGB--event extrinsics. One recording without RGB frames is excluded.
The dataset's depth images are not metric, so, as for DSEC, the frames are warped with MoGe-2 depth scaled by a factor fitted per camera set, using the time offsets from a Kalibr calibration and extrinsics refined by aligning frames with events.
The refinement scans include the test recording \texttt{arm\_hdr}. The RGB camera sees only part of the event cameras' field of view, so only the overlap is valid and scored. Some views show black strips or rolling-shutter effects, which we do not correct.

\subsection{Forecasting Protocol}
\label{sec:supp-forecasting}
Forecasting is evaluated on BS-ERGB's 21 scored test recordings (Sec.~\ref{sec:supp-splits}).
Each recording has one forecast origin $t_0$, the start of the 11th of its 25 test windows, and the forecast covers the five frame intervals that follow, about 187\,ms at BS-ERGB's mean interval of 37.4\,ms.
No frame after $t_0$ is given to any method.

EventDiffusion forecasts by rollout, generating one frame interval at a time in 8 slices, and each later interval reads the stream generated before it as its history. The input settings are:
\begin{itemize}
\item \emph{Single RGB}: the frame at $t_0$, given at every step with its age, and no past events.
\item \emph{RGB + events}: the frame at $t_0$, aged as above, and the real events before $t_0$.
\item \emph{Past RGB}: the events between the two frames before $t_0$, generated by EventDiffusion from those frames, then the frame at $t_0$, aged as above.
\item \emph{Past events}: the real events before $t_0$ and no frame.
\end{itemize}
EAD reads the last slice of the history and ESA and ETD the last 20\,ms, so ``Events (20\,ms)'' in Table~5 of the main paper is the longest span any stage reads. The forecast's event count is not anchored to the past rate.

In order to evaluate the baselines fairly, using their own proposed protocols, each one is provided either with equivalent or greater amounts of conditioning evidence to support its forecasts, and we initialise from their pre-trained weights where available.
\emph{RGB prediction plus simulation.} SimVP~\cite{gao2022simvp,tan2022simvp} (v2, gSTA) is trained on BS-ERGB's training recordings to predict five frames from the two latest observed frames. Cosmos Edge~\cite{nvidia2026cosmos3} predicts the five frames from the five latest observed frames in one call. ESIM and DVS-Voltmeter then simulate events from the predicted frames with their calibrated BS-ERGB parameters, after warming up on the three latest observed frames. In contrast, our own technique only ever receives 1 or 2 RGB frames.
\emph{Event history.} E-TIDE had no released code or checkpoint to support this comparison. E-Motion~\cite{wu2024emotion} is run on 6 voxel frames at 60\,fps, the 100\,ms of real events before $t_0$, with the event camera's lattice resized to $256\times256$ as its code expects. In contrast, our own technique only ever receives 0 or 20\,ms of past events.

CEFS (Table~5 of the main paper) is computed on each of the five forecast intervals against the real events of the same interval and area, and averaged over the 105 intervals. Table~4 of the main paper scores the span $[t_0,t_0+H)$ for $H\in\{5,10,20,50\}$\,ms. E-Motion forecasts voxel frames rather than events, so every method is compared with it in E-Motion's own representation.
 The real events and each method's events are voxelized with E-Motion's voxel grid (3 bins per $1/60$\,s frame, normalized per frame, resized to $256\times256$ and scaled by the clip's largest magnitude), and E-Motion's forecast is taken as stored. F1 and IoU between each method's mask and the real mask are averaged over frames, then over leads within each sequence, then over sequences. The count CRPS of Table~4 compares the event counts of three forecasts (seeds 0--2) over the first 25\,ms with the real count, as the empirical CRPS of the three samples. It is pooled over sequences as the total CRPS divided by the total real count.

\subsection{Timing Diagnostics}
\label{sec:supp-metrics}
The timing diagnostics of Table~3 of the main paper follow TIDES~\cite{thirgood2026tides}. Each is computed per window on the whole time-sorted stream inside the scored area, with integer-microsecond timestamps, and reported as a distance to the paired recording.
Let $\delta$ be the gaps between consecutive events of the stream, of any pixel or polarity.
\begin{itemize}
\item \emph{Same-ts}: the share of gaps $\delta\le\theta$ for $\theta\in\{1\,\mathrm{ns},0.1\,\mu\mathrm{s},1\,\mu\mathrm{s}\}$. We compute the absolute difference between the generated and recorded shares and average over the three thresholds.
\item \emph{ISI-spike}: the share of gaps $\delta\le10\,\mu$s. We compute the absolute difference between the generated and recorded shares.
\item \emph{Fano}: event counts in bins of 10, 50 and $200\,\mu$s from each stream's first event, with Fano factor $F=\mathrm{Var}/\mathrm{mean}$ over the bins. We average the error $|\log(F_{\rm gen}/F_{\rm rec})|$ over the three bin widths to measure how closely the generated stream matches the burstiness of the recording.
\end{itemize}
PECS distances (P-CD) follow~\cite{han2024pecs} and EQS is computed as described in Sec.~\ref{sec:supp-cefs-baselines}.

\section{Method and Implementation Details}
\label{sec:supp-method}
\label{sec:supp-implementation}
EventDiffusion generates a native event stream through a sequence of conditional decisions. EAD samples polarity-specific event counts for each spatial cell and time slice. ESA allocates those counts to pixels while preserving their totals. ETD assigns timestamps to the resulting events while retaining their pixels and polarities. Each stage therefore resolves a distinct part of the stream while preserving the decisions made before it. The same factorization supports both generation between RGB frames and forecasting beyond the last observed frame. Secs.~\ref{sec:supp-ead}--\ref{sec:supp-etd-sampling} give the details of each stage, Sec.~\ref{sec:supp-impl-details} their training, including a few-step version of EAD's sampler, Sec.~\ref{sec:supp-forecast-model} forecasting, and Sec.~\ref{sec:supp-runtime} the time generation takes.

\subsection{Event Activity Diffusion}
\label{sec:supp-ead}
EAD models event activity as integer-valued ON and OFF counts across spatial cells and temporal slices. Conditioning on visual input, interval duration, camera and event history lets it adapt the activity to the observed scene and sensor. Sampling reverses the thinning of the main paper's Eq.~(1), starting from the empty count map at $\xi=1$ and ending at $\xi=0$. For a step from $\xi$ to $\xi'<\xi$, let $p=\pi_{\rm EAD}(\xi)$ and $q=\pi_{\rm EAD}(\xi')$. Each current count is kept with probability $1-\delta$, and new events are drawn from the predicted missing count $\widehat Y_\xi$ of Eq.~(2):
\begin{equation}
\begin{aligned}
\delta&=\epsilon\min\{1,(1-q)/p\},\qquad \beta=\frac{q-(1-\delta)p}{1-p},\\
X_{\xi'}&\sim\operatorname{Bin}(X_\xi,1-\delta)+\operatorname{Bin}\big(\mathrm{RR}(\widehat Y_\xi),\beta\big),
\end{aligned}
\label{eq:supp-ead-reverse}
\end{equation}
where $\mathrm{RR}(y)=\lfloor y\rfloor+\mathrm{Bernoulli}(y-\lfloor y\rfloor)$ rounds without bias. If $X_\xi$ has mean $pA$ and $\widehat Y_\xi$ equals the expected missing count $(1-p)A$, then $X_{\xi'}$ has mean $qA$, the mean count at step $\xi'$. The attrition rate $\epsilon$ lets every step remove a small fraction of the counts placed at earlier steps, so the sampler can revise them. Setting $\epsilon=0$ gives plain reverse thinning. At the last step $q=1$, so $\delta=0$ and no count is removed.
Each step calls EAD's network once, so the number of steps sets EAD's cost: at 128 steps EAD takes about 18\,s per DSEC frame interval on an RTX 3090.
To reduce this cost, we distil EAD's count-generation process into a few-step student (Sec.~\ref{sec:supp-ead-distill}). The student retains the same integer-valued reverse update and supplies counts to ESA in the same way as the teacher.

The RGB frames are encoded using a U-Net whose encoder is BiT's ResNetV2-50x1~\cite{kolesnikov2020bit} with ImageNet-21k weights, its first convolution widened to the ten input channels, and whose decoder is FiLM-modulated by the interval's length and the camera embedding.
It is trained jointly with EAD. Its input is the two RGB frames that bound the interval, their log-luminance images, the absolute difference of the two log-luminances, and a mask of the valid sensor support.

EAD also takes the history of the previous interval as input.
The event history is the ON and OFF counts of the last slice before the interval, read as $\log(1+n)/2$, with that span's duration in units of 20\,ms.
In training, 30\% of intervals keep their real history, so the model also learns to start without one. To reduce exposure bias, training also uses one-frame intervals whose history is the previous interval's last slice, sampled by an earlier EAD checkpoint in stored chained rollouts over the training sequences.
Each camera has a learned baseline event rate $\lambda_{d_{\rm cam}}$, initialized from the count statistics of the training data, and a learned embedding that enters the network. Index 0 is an unknown camera with its own baseline event rate and embedding. In training, the camera is replaced by the unknown one with probability 0.15, so the model can also generate for a camera it has not seen.

\subsection{Event Spatial Allocation}
\label{sec:supp-esa}
ESA turns each sampled cell count into pixel-level events through hierarchical splits. Each split assigns $m$ events to one region and the remaining $n-m$ to the other, preserving the polarity-specific count exactly as spatial detail is introduced. A fully convolutional U-Net predicts the parameters $(\eta,\kappa)$ of Eq.~(5) of the main paper for every split. It works at resolutions of 1, 2, 4 and 8 pixels with widths 32, 64, 96 and 128, two FiLM residual blocks per level in the encoder and one in the decoder, and its head reads the 1\,px features, folded into 2\,px cells, together with the 2\,px features.
Its inputs are EAD's ten RGB input channels on the event grid, the cell mask $m_c$, each cell's counts repeated over its pixels, each pixel's position within its cell, and the pixel counts of the cells already allocated. The event history is, per polarity, $\log(1+n)$ of the counts in the 20\,ms before the slice. FiLM conditioning carries the slice's position and duration, whether frames are given,
and the event history's duration. In training the event history is absent for 60\% of samples, partial for 10\% and complete for 30\%, and with probability 0.5 it comes from a stored stream generated by the full pipeline.

\subsection{Event Timing Diffusion}
\label{sec:supp-etd-sampling}

ETD models the timing of the events placed by ESA, keeping their number, pixels and polarities fixed. Its conditional density combines a background component for temporal variation, anchor kernels for correlations with nearby events, row and column context for timestamp sharing, and same-pixel inhibition for refractory behaviour. Together these terms place individual events in time while retaining the activity and spatial allocation sampled by the preceding stages.

The bounded Fourier series of $\rho^{\rm b}_i$ uses the sines and cosines of the time since the slice start at
25 periods from 1 to 64\,ms, spaced by factors of $2^{1/4}$, and four cosines
$\cos(m\pi\tau)$ of the position in the slice. We bound its log-density by
$15\tanh(\cdot/15)$ and normalize it on 512 points. Each event has $K=16$ anchor events, and the logistic kernels $\rho^{\rm a}_{ij}$ have offsets within
$\pm2$\,ms and widths within 5\,$\mu$s--2\,ms. A sample is a $32\times32$\,px patch with a 16\,px margin, and each event carries a camera embedding (replaced by an unknown camera with probability 0.15 in training). A patch with more than
2{,}048 events in a slice is split into the fewest groups of whole pixels with at
most 2{,}048 events each, and ETD times each group as its own sample. Each sample
reads at most 1{,}024 events from the patch margin, 1{,}024 events in the same rows and columns ($\mathcal C_i$), the latest 512 of its patch's
events from the 20\,ms before the slice, and a 40-bin profile of the whole frame's event rate over that span. In training the real event history is shown for 30\% of samples, and with probability 0.5 the history comes from a stored stream generated by the full pipeline.

We train with $T=100$ steps and sample with 6 of them, spaced so that each step brings in the same share of the events. An event never placed still holds its noise time $U_i$, so we treat
it as noise regardless of $q_i$ and give it confidence $v_m=0$ in the
same-pixel inhibition of other events until it is placed. Events selected for timestamp replacement at the current step also have $v_m=0$ until they are placed again, and every other event has
$v_m=1$. Rejection runs for 16 rounds, in each of which the next waiting
event of each pixel draws one proposal from $\tilde p_i$. We draw the events still
waiting from $\tilde p_ig_i$ on a grid of cells of at most 10\,$\mu$s, choosing a cell
with probability proportional to $\tilde p_ig_i$ at its centre and then a uniform time
inside it. This fallback omits $\rho^{\rm t}$, whose mass is concentrated
within single microseconds and so falls between grid nodes.

Near the start of a slice, $g_i$ also includes the pixel's events before the
slice, each with confidence $v_m=1$. Near its end, the pixel's events in the next slice are not yet generated,
and without them events would pile up at the slice's end. At sampling we
therefore multiply $g_i$ by the probability that none of the pixel's later events,
not yet generated, suppresses the density at $\tau$, taking those events as Poisson at the rate $\lambda_e$ of the pixel's other events in the slice, separately for the same and the opposite polarity $e$ (the event itself excluded):
$\exp\!\big[-\sum_e\lambda_e\int_{(1-\tau)\Delta}^{\infty}h_{e}(u)\,du\big]$.

We draw one step per sample uniformly from $\{1,\dots,T\}$. Events whose nearest
same-pixel event in the corrupted times lies within 1\,ms are weighted by 3 in the
cross-entropy, as the refractory period acts on these events. The inhibition term
$-\log g_i(\tau_{0,i})+\log Z_i$ of the main paper's ETD loss uses at most
1{,}024 corrupted events per batch, takes each other event's predicted $q_m$,
without gradient, as its confidence $v_m$, and estimates $Z_i$ as the mean of $g_i$
over 16 draws from $\tilde p_i$. The suppression parameters are predicted from detached features, so this
term trains $g_i$ alone and leaves $\tilde p_i$ fixed.

\subsection{Training}
\label{sec:supp-impl-details}
\paragraph{Training strategy.} EAD, ESA, and ETD are trained independently using their respective objectives. Each stage is first trained on the pooled training recordings to produce the shared EventDiffusion-G pipeline. EventDiffusion-FT subsequently fine-tunes the stages on each dataset to adapt to its event statistics. During later training phases, a proportion of batches use event histories generated by an earlier version of the pipeline rather than recorded histories. This reduces the mismatch between training and inference, where each generated interval conditions subsequent generation. Table~\ref{tab:supp_training} specifies the optimizer and cumulative training steps for each stage and checkpoint.

Each stage is trained on one GPU with AdamW ($\beta=(0.9,0.99)$, weight decay 0.01), a linear warm-up followed by cosine decay to 10\% of the peak learning rate in every training phase, gradient clipping at 1.0 and bf16 autocast. In half of the batches of the later training phases, the event history each stage reads is generated by an earlier version of the pipeline (Secs.~\ref{sec:supp-ead}--\ref{sec:supp-etd-sampling}).

\begin{table}[t]
\centering\scriptsize
\setlength{\tabcolsep}{2.5pt}
\caption{Training settings. LR: peak learning rate and warm-up steps, from scratch / in each fine-tuning phase. Steps: all phases, for EventDiffusion-G / EventDiffusion-FT (the per-dataset fine-tune). Gap-5$\times$ adds 20k steps of ESA and ETD.}
\label{tab:supp_training}
\begin{tabular*}{\columnwidth}{@{\extracolsep{\fill}}lccc@{}}
\toprule
& \textbf{EAD} & \textbf{ESA} & \textbf{ETD} \\
\midrule
Parameters & 99.2M & 2.21M & 2.43M \\
LR, warm-up & 2e-4, 2k / 1e-4, 500 & 1e-3, 1k / 1e-4, 500 & 3e-4, 2k / 1e-4, 500 \\
Batch & $\le$10 slices (1--3 gaps) & 8 crops & 32 patches \\
Input size & $480\times640$ & $256\times320$ & 32\,px patch \\
EMA decay & 0.9995 & -- & -- \\
\bottomrule
\end{tabular*}
\end{table}

\paragraph{Few-step EAD by step distillation.}
\label{sec:supp-ead-distill}%
EAD's iterative reverse process produces the cell counts that determine the number of events ESA must allocate. To reduce the number of network evaluations while preserving this interface, we train a student to take eight larger reverse steps in place of the teacher's 128. The student retains EAD's network and the integer-valued update of Eq.~\eqref{eq:supp-ead-reverse}, so its sampled ON/OFF counts pass directly to ESA without changing the subsequent stages.

For each training batch, we choose a step $i$ on the student's eight-step grid. The student's moving average samples from an empty map up to that step, exposing training to the count maps the student encounters during generation. From the resulting map, the frozen teacher covers the next student step in four steps of the second-order $\theta$-trapezoidal sampler ($\theta=\tfrac12$). The training target is the teacher's expected count addition over those steps, for each cell, polarity and slice. The student predicts a missing count so that its single reverse step matches that target in expectation, using the count error of Eq.~(3) of the main paper. The loss matches expected additions, rather than the complete joint distribution of the teacher's transitions. Each student step remains a product of binomials and produces integer counts. We train the student for 6{,}000 steps (learning rate $2\times10^{-5}$ after 500 warm-up steps, EMA decay 0.999) on the teacher's training data, including generated histories.

On DSEC, the few-step pipeline achieves slightly higher CEFS and lower Fano error than the full-step configuration while using fewer reverse steps (Sec.~\ref{sec:supp-sampler-allocation}). The two pipeline configurations differ in settings beyond EAD, and matching expected additions does not require the student to reproduce the teacher's full transition distribution or evaluation scores exactly. The comparison identifies an advantageous empirical fidelity--runtime operating point for the few-step pipeline.

\subsection{Forecasting}
\label{sec:supp-forecast-model}
EventDiffusion forecasts individual asynchronous events, including their pixel locations, polarities, and continuous timestamps, from different combinations of past observations. The same generative framework supports a single RGB frame, multiple past RGB frames, event history alone, or RGB combined with event history. Unlike approaches that predict temporally aggregated event representations, the output is a native event stream whose temporal resolution is not fixed by the forecasting model. Unlike RGB-prediction-plus-simulation approaches, EventDiffusion does not require an intermediate prediction of future RGB frames.

Forecasting uses the same activity--allocation--timing factorization as RGB-conditioned simulation. Starting from the observations available at the forecast origin, EAD generates future activity, ESA allocates the sampled counts to native sensor pixels, and ETD generates continuous timestamps for those events. Generation proceeds causally, one frame interval at a time in eight temporal slices, with each subsequent interval conditioned on the previously generated events. No future RGB frames are supplied. Section~\ref{sec:supp-forecasting} specifies the conditioning and evaluation for each forecast setting.

Forecasting requires the stages to operate with the observations available at the origin and with their own generated history. During training, the end frame is dropped in 30\% of batches, and previous events are dropped in another 30\%. The later training phases also use generated histories (Sec.~\ref{sec:supp-impl-details}). These settings expose the model to the missing-frame and history conditions encountered at inference, while retaining the same three-stage generation process.

During a rollout, the history available to EAD changes from an observed event stream to events sampled by the model itself. To make the preparation of this conditioning explicit, the second EAD fine-tuning phase applies Diffusion Forcing~\cite{chen2024diffusionforcing} to the sequence of slices and the event history. Its noise process is EAD's own count thinning (Eq.~(1) of the main paper). In half of the gaps, each slice draws its own noise level. In half of the gaps with a history, the history is thinned to a random level and that level is supplied to the network. The model thus learns to predict future activity with context at different known thinning levels.

At inference, an observed event history is supplied unchanged at level 0. A history generated by EventDiffusion is thinned to level 0.5 and supplied with that level. This gives EAD access to previous activity for continuity while distinguishing generated context from an unthinned observation. ESA and ETD then allocate and time the sampled activity as in the rest of the pipeline.

The forecasts of Tables~4 and~5 and Fig.~4 of the main paper come from this model. Sec.~\ref{sec:supp-forecast-timing} examines their timing and temporal resolution, and Fig.~\ref{fig:forecast_horizon} shows forecasts across horizons.

\subsection{Inference and Runtime}
\label{sec:supp-runtime}
Table~\ref{tab:supp_runtime} compares generation time on DSEC with fidelity. We use the distilled EAD with 8 steps (Sec.~\ref{sec:supp-ead-distill}), ESA allocating every cell at once, and ETD with 6 steps. EAD draws the next frame interval while ESA and ETD process the current one, so an interval takes about as long as ETD (4.4\,s) rather than the sum of the three stages.
The faster baselines incur a substantial loss in event fidelity. On DSEC, V2CE generates a frame interval in 0.141\,s but reaches a CEFS of 0.502, while DVS-Voltmeter takes 1.17\,s with interpolated frames and reaches 0.576. EventDiffusion takes 4.8\,s per interval and achieves 0.700 using native RGB frames. Its additional computation therefore provides substantially higher measured event fidelity, making it attractive when generation is performed offline and the resulting streams are reused.

\begin{table}[t]
\centering\footnotesize
\setlength{\tabcolsep}{4pt}
\caption{Generation time and fidelity on DSEC test, RTX 3090. EventDiffusion: EventDiffusion-FT with the few-step settings, mean over three seeds. Times are in seconds per 50\,ms frame interval. Baseline timings cover six full DSEC recordings. Recording: a DSEC training recording of average length (1{,}285 intervals). DVS-Voltmeter on frame pairs without interpolation takes 0.16\,s.}
\label{tab:supp_runtime}
\begin{tabular}{@{}llccc@{}}
\toprule
\textbf{Method} & \textbf{Frames} & \textbf{CEFS}$\uparrow$ & \textbf{Time (s)} & \textbf{Recording} \\
\midrule
ESIM          & SSM-10$\times$ & 0.575 & 0.236 & 5.1\,min \\
DVS-Voltmeter & SSM-10$\times$ & 0.576 & 1.17  & 25\,min \\
V2CE          & native         & 0.502 & 0.141 & 3.0\,min \\
\midrule
EventDiffusion (few-step) & native & 0.700 & 4.8 & 103\,min \\
\bottomrule
\end{tabular}
\end{table}

\section{Composite Event Fidelity Score}
\label{sec:supp-cefs}
This section gives the complete definition of CEFS (Sec.~4 of the main paper), how its bin-size weights are set, and the evidence behind the main paper's statements about PECS and EQS.

\subsection{Full Definition}
\label{sec:supp-cefs-definition}

\paragraph{Inputs.}
A real stream $r$ and a generated stream $g$ are scored over the same sensor rectangle and the same half-open window $[t_0,t_1)$ of length $L=t_1-t_0$. Scoring does not resize or register the streams, stretch time, match counts, denoise, or jitter timestamps. Timestamps are kept as integer nanoseconds so that the recorded clock and its ties are preserved. In the main paper each window is one native RGB frame interval, and only events inside complete valid $32{\times}32$\,px patches are scored.

\paragraph{Activity and polarity.}
$D_{\rm count}$ and $D_{\rm pol}$ are given by Eq.~(11) of the main paper. $D_{\rm count}$ depends only on $\log_2(N_g/N_r)$, so producing half or twice the real count gives $D_{\rm count}=0.393$, a quarter or four times gives $0.865$, and ten times gives $0.996$. If both streams are empty CEFS is $1$, and if exactly one is empty it is $0$.

\paragraph{Space--time histograms.}
The 30 bin sizes $k=(s,\Delta)$ combine $s\in\{1,2,4,8,16\}$\,px with $\Delta\in\{10,50,100,500\,\mu{\rm s},1,5\,{\rm ms}\}$. At each bin size, every event is splatted trilinearly from $\big(x/s,\,y/s,\,(t-t_0)/\Delta\big)$ onto the $2^3$ surrounding nodes. With fractional coordinates $u_x,u_y,u_t$, corner $(a,b,c)\in\{0,1\}^3$ receives
\begin{equation}
\omega_{abc}=u_x^a(1-u_x)^{1-a}\,u_y^b(1-u_y)^{1-b}\,u_t^c(1-u_t)^{1-c}.
\label{eq:supp-cefs-splat}
\end{equation}
The grid has $\lceil W/s\rceil{+}1\times\lceil H/s\rceil{+}1\times\lceil L/\Delta\rceil{+}1$ nodes, so mass at the window edges is kept. ON and OFF events go to separate histograms, and each bin size is splatted from the original events rather than pooled from a finer one.

\paragraph{Structure.}
Each histogram is normalized to a distribution, compared by the squared Hellinger distance $h_{k,\sigma}$ of Eq.~(12), and pooled into
\begin{equation}
H_{\rm ST}=\textstyle\sum_k\widetilde w_k\sum_\sigma\frac{N_r^\sigma}{N_r}\,h_{k,\sigma},\qquad D_{\rm ST}=H_{\rm ST}^2,
\label{eq:supp-cefs-dst}
\end{equation}
which is Eq.~(13). The weights $\widetilde w_k$ are given in Sec.~\ref{sec:supp-cefs-weights}. A polarity that has events in $r$ but none in $g$ receives the maximum distance. A polarity absent from $r$ carries zero weight. Extra generated polarity is penalized by $D_{\rm pol}$ and $D_{\rm count}$.

\paragraph{Combination.}
The discrepancies are merged with $B_\lambda$ (Eq.~14) as in Eq.~(15), with
\begin{equation}
\lambda_T=0.3164,\qquad \lambda_C=0.5226 .
\label{eq:supp-cefs-lambda}
\end{equation}
Both were fixed once from corrupted real recordings (MVSEC \texttt{outdoor\_day1}~\cite{zhu2018mvsec}, DAVIS \texttt{boxes\_6dof}~\cite{mueggler2017davis} and the DSEC training recording \texttt{zurich\_city\_05\_b}), and no generated stream was used. For each discrepancy we record its increase at every step of every corruption ladder of Sec.~\ref{sec:supp-cefs-validation}, normalize these increases to unit mass, and average over the three recordings to obtain a response profile $p$. Then $\lambda_T=1-\sum_e\min(p_{{\rm async},e},p_{{\rm temp},e})$, the share of responses that $D_{\rm async}$ and $D_{\rm temp}$ do not have in common, and $\lambda_C$ follows in the same way from $D_{\rm ST}$ and $D_{\rm timing}$. The same two values are used for every dataset.

\paragraph{Properties.}
Every discrepancy lies in $[0,1]$, and for $x,y,\lambda\in[0,1]$
\begin{equation}
\max(x,y)\le B_\lambda(x,y)\le 1 ,
\label{eq:supp-cefs-bound}
\end{equation}
so $0\le\mathrm{CEFS}\le(1-D_{\rm count})(1-D_{\rm pol})\le 1$. $B_\lambda(1,y)=1$, so a complete failure in structure or timing gives CEFS $0$ regardless of the other terms, and $B_\lambda(x,0)=x$. When $x>y$, $\partial B_\lambda/\partial x=1-\lambda y$ and $\partial B_\lambda/\partial y=\lambda(1-x)$ are both positive, so reducing either discrepancy raises CEFS, unlike a plain maximum (Table~\ref{tab:supp-cefs-aggregation}). A $k$-fold duplicate of $r$ has the same normalized histograms and timestamp profile as $r$, so its score is $1-D_{\rm count}(k)$, whereas its nearest-neighbour distance to $r$ is zero.

\begin{table}[t]
\centering\small
\setlength{\tabcolsep}{4pt}
\caption{Combining discrepancies with a plain maximum and with CEFS ($D_{\rm count}=D_{\rm pol}=0$). The maximum ignores every discrepancy but the largest. CEFS also charges a bounded share of the others.}
\label{tab:supp-cefs-aggregation}
\begin{tabular}{lcc}
\toprule
$(D_{\rm ST},D_{\rm async},D_{\rm temp})$ & $1-\max$ & $1-D_{\rm core}$ \\
\midrule
$(0.40,0.40,0.40)$ & 0.600 & 0.415 \\
$(0.40,0.05,0.05)$ & 0.600 & 0.580 \\
$(0.80,0.02,0.02)$ & 0.200 & 0.197 \\
$(0.20,0.60,0.60)$ & 0.400 & 0.290 \\
\bottomrule
\end{tabular}
\end{table}

\subsection{Bin-Size Weights}
\label{sec:supp-cefs-weights}
At fine bin sizes a sparse window holds too few events per bin for two recordings of the same scene to agree. The weights $\widetilde w_k$ therefore down-weight bin sizes at which perturbed copies of $r$ already disagree with $r$. They are computed from $r$ alone, once per window, and every method is scored with the same weights.

\paragraph{Perturbed copies.}
For each real window we build 16 Bernoulli-$0.5$ thinnings (random halves, rescaled to the original mass), 16 bootstrap resamples of the occupied $16\,{\rm px}\times16\,{\rm px}\times5\,{\rm ms}$ blocks (keeping each block's events and both polarities), and six small shifts: uniform spatial jitter of $0.25$\,px, four $0.25$\,px translations along the signed axes, and uniform timing jitter of $10\,\mu$s, all clipped to the sensor and window.

\paragraph{Stability.}
Each copy $r'$ is compared with $r$ at every bin size using the variance-stabilized distance
\begin{equation}
\begin{aligned}
A_k(r,r')&=\frac{1}{8N_r}\sum_v\big[\mathcal A(p_v)-\mathcal A(q_v)\big]^2,\\
\mathcal A(x)&=2\sqrt{N_rx+3/8},
\end{aligned}
\label{eq:supp-cefs-anscombe}
\end{equation}
where $p$ and $q$ are the normalized histograms of $r$ and $r'$ over both polarities. The stability of bin size $k$ is the smallest of three values: the 10th percentile of $1-A_k$ over the thinnings, the same over the block resamples, and the minimum of $1-A_k$ over the six shifts.

\paragraph{Weights.}
Let $w^{(s)}_i$ and $w^{(t)}_j$ be trapezoidal weights on the logarithmic spatial and temporal bin-size axes, each normalized to sum to one, and $R_{ij}$ the stability of bin size $(s_i,\Delta_j)$. With
\begin{equation}
u_{ij}=\mathrm{clip}\!\left(\tfrac{R_{ij}-0.90}{0.09},0,1\right),\qquad \rho_{ij}=u_{ij}^2(3-2u_{ij}),
\label{eq:supp-cefs-taper}
\end{equation}
the weights are $\widetilde w_{ij}\propto w^{(s)}_iw^{(t)}_j\rho_{ij}$, normalized to sum to one. A bin size thus gets no weight below stability $0.90$ and its full weight above $0.99$. If every weight is zero, the window cannot be scored. The thresholds $0.90$ and $0.99$ are fixed design choices.

\subsection{Timing Terms}
\label{sec:supp-cefs-temporal}

\paragraph{Asynchrony.}
$D_{\rm async}$ compares how strongly events share timestamps in $g$ and in $r$. It uses 25 clock views per polarity: the exact timestamps, and each $\Delta$ of the bin sizes above at four phase offsets $q\Delta/4$, $q\in\{0,1,2,3\}$, which group events by $\lfloor(t-t_0+q\Delta/4)/\Delta\rfloor$. In a view where a group holds $m$ events of stream $e$ (with $n_e$ events of that polarity, and $n_r$ in $r$), the group is given the value
\begin{equation}
z_e(m)=\frac{\log\max\{1,\,m\min(1,n_r/n_e)\}}{\log\max(2,n_r)} ,
\label{eq:supp-cefs-async-z}
\end{equation}
which rescales group sizes when $g$ has more events than $r$, so over-generation alone does not count as batching. Weighting each group by its share of events gives a distribution of $z$ with CDF $F_e$. The one-sided distance $d_{v,\sigma}=\int_0^1[F_r(z)-F_g(z)]_+\,dz$ grows only when $g$ concentrates more events per timestamp than $r$. With $d_v=\sum_\sigma\frac{N_r^\sigma}{N_r}d_{v,\sigma}$,
\begin{equation}
D_{\rm async}=\Big[\max\Big\{d_{\rm exact},\ \max_{\Delta}\tfrac14\textstyle\sum_{q=0}^{3}d_{\Delta,q}\Big\}\Big]^2 .
\label{eq:supp-cefs-async}
\end{equation}
A missing generated polarity has $d_{v,\sigma}=1$. $D_{\rm async}$ only penalizes excess sharing of timestamps. A value of zero does not certify inter-event intervals or refractory behaviour.

\paragraph{Temporal distribution.}
For each $\Delta$, the histograms are summed over space and normalized per polarity, which places each polarity's mass $\nu_{e,\Delta,\sigma}$ on the temporal nodes $m\Delta$, $m=0,\dots,\lceil L/\Delta\rceil$. Then
\begin{equation}
D_{\rm temp}=\sum_j\widetilde v_j\sum_\sigma\frac{N_r^\sigma}{N_r}\,\frac{W_1(\nu_{r,\Delta_j,\sigma},\nu_{g,\Delta_j,\sigma})}{L+5\,{\rm ms}},\quad \widetilde v_j=\sum_i\widetilde w_{ij},
\label{eq:supp-cefs-temp}
\end{equation}
where $W_1$ is the exact one-dimensional Wasserstein distance. The last node lies at most one bin, and so at most $5\,$ms, beyond $t_1$, so no transport exceeds $L+5\,$ms and each term is at most $1$. A missing generated polarity receives distance $1$. $D_{\rm temp}$ measures when activity occurs within the window. $D_{\rm ST}$ measures whether that activity occurs at the right place and time.
\subsection{Failure Cases of PECS and EQS}
\label{sec:supp-cefs-baselines}

Existing event-stream metrics capture complementary aspects of
fidelity but can overlook important discrepancies.

\paragraph{PECS.}
PECS~\cite{han2024pecs} independently min--max normalizes
each event stream before computing Chamfer and Gaussian distances.
Consequently, a constant spatial translation or temporal shift
can leave the normalized point sets unchanged, producing zero
distance despite misalignment in the original sensor coordinates.
Nearest-neighbour matching can also overlook duplicated events,
since multiple generated events may match the same reference event.
CEFS instead evaluates streams in their shared physical coordinates
and explicitly penalizes count discrepancies.

\paragraph{EQS.}
EQS~\cite{chanda2025eqs} compares learned event representations.
Our evaluation uses a pretrained RVT-Small~\cite{gehrig2023rvt}
encoder with ten temporal count bins. This adaptation can be
insensitive to fine timestamp changes and missing activity.
CEFS directly evaluates event counts, multiscale space--time
structure, and temporal fidelity without relying on a learned
feature representation.

Section~\ref{supp:cefs_comparators} provides controlled comparisons with PECS and EQS,
including absolute-alignment, event-duplication, and timestamp
quantization experiments.

\subsection{Validation under Controlled Corruptions}
\label{sec:supp-cefs-validation}

We evaluate CEFS using controlled perturbations of real event
streams. Seven reference windows are selected from MVSEC, the
DAVIS Event-Camera Dataset, and a DSEC training recording
(Table~\ref{tab:supp-cefs-windows}). These recordings were used
during metric development and are separate from the
EventDiffusion test recordings.

Each corruption is applied at predefined severity levels using
four seeds. All metrics receive identical corrupted
streams, and all events are evaluated without subsampling.
The eight corruption families in Table~\ref{tab:supp-cefs-corruptions}
cover event count, polarity, spatial allocation, space--time
correspondence, and timestamp fidelity. Separate controls use spatial
jitter (1, 2, 5, 10\,px) and exact duplication ($2\times$, $4\times$).

\begin{table}[t]
\centering\small
\setlength{\tabcolsep}{3pt}
\caption{Real reference windows used for controlled corruption
experiments. Sensor resolutions are $346\times260$ (MVSEC),
$240\times180$ (DAVIS), and $640\times480$ (DSEC).}
\label{tab:supp-cefs-windows}
\begin{tabular}{llrr}
\toprule
Dataset & Recording & Window & Events \\
\midrule
MVSEC & \texttt{outdoor\_day1} & 100\,ms & 673 \\
MVSEC & \texttt{outdoor\_day2} & 100\,ms & 299 \\
MVSEC & \texttt{outdoor\_night1} & 100\,ms & 11{,}554 \\
DAVIS & \texttt{boxes\_6dof} & 100\,ms & 99{,}052 \\
DAVIS & \texttt{dynamic\_6dof} & 100\,ms & 50{,}057 \\
DAVIS & \texttt{shapes\_6dof} & 100\,ms & 8{,}194 \\
DSEC & \texttt{zurich\_city\_05\_b} & 5\,ms & 86{,}582 \\
\bottomrule
\end{tabular}
\end{table}

\begin{table*}[t]
\centering\small
\caption{The eight development corruption families plotted in
Fig.~\ref{fig:cefs_corruption_compact}.
Each severity sequence includes the unmodified reference.}
\label{tab:supp-cefs-corruptions}
\begin{tabular}{p{.24\textwidth}p{.33\textwidth}p{.34\textwidth}}
\toprule
Corruption & Severities & Main effect \\
\midrule
Random deletion & 10, 25, 50, 75, 100\% removed
& Event count and structural stability \\

Random addition & 10, 25, 50, 100\% added
& Count and false space--time activity \\

Polarity inversion & 10, 25, 50, 100\% flipped
& Polarity balance and signed structure \\

Spatial replacement & 10, 25, 50, 75, 100\% of coordinates replaced uniformly
& Spatial distribution \\

Spatial/temporal permutation & 10, 25, 50, 100\% of coordinates shuffled
& Space--time correspondence with the spatial marginal preserved \\

Timestamp quantization & 10 and 100\,$\mu$s, and 1 and 10\,ms
& Temporal precision and timestamp sharing \\

Timestamp collapse & 10, 25, 50, 100\% moved to $t_0$
& Temporal distribution and timestamp sharing \\

Temporal shuffle & 10, 25, 50, 100\% of timestamps shuffled
& Space--time correspondence with the timestamp marginal preserved \\
\bottomrule
\end{tabular}
\end{table*}

For each corruption, we measure the change in CEFS error
$1-\mathrm{CEFS}$ relative to the unmodified reference.
Results are averaged over corruption seeds within each recording.
We assess sensitivity to corruption severity, monotonicity,
and tolerance to small spatial and temporal perturbations.
The quantitative results, response curves, and comparisons against existing
metrics are presented in Sections~\ref{supp:cefs_corruptions}--
\ref{supp:cefs_comparators}.

\section{Additional Results}
\label{sec:supp-results}
The per-dataset, per-sequence and Gap-5$\times$ results in this section come from EventDiffusion-FT (Sec.~\ref{sec:supp-splits}), each dataset's own EAD, ESA and ETD are scored on the test windows of that section. As for the main paper, no real events are given: the first frame interval of a recording is generated from its frames alone, and each later one follows the stream generated before it. Table~\ref{tab:supp_ed_native_datasets} gives every metric for EventDiffusion-FT on each dataset under Native RGB. Table~\ref{tab:supp_ed_native_sequences} breaks Table~\ref{tab:supp_ed_native_datasets} down by test recording. Tables~\ref{tab:supp_ed_gap5_datasets} and~\ref{tab:supp_ed_gap5_sequences} give the per-dataset and per-sequence scores behind EventDiffusion-FT's Gap-5$\times$ row in Table~2 of the main paper.
\begin{table*}[tp]
\centering\scriptsize
\setlength{\tabcolsep}{3.5pt}
\caption{\textbf{EventDiffusion-FT under Native RGB, per dataset.} Each dataset's fine-tuned EAD, ESA and ETD. Each window is one native frame interval, scored as in the main paper over complete valid $32{\times}32$\,px patches. We average across windows within each sequence, then across sequences within each dataset. CEFS and its components are defined in Sec.~\ref{sec:supp-cefs}. P-CD is the PECS Chamfer distance by the paper's equations. Same-ts, ISI-spike and Fano are the timing errors against the real stream of Table~3 of the main paper. $N_g/N_r$ is the generated over the real event count (best at~1). Mean: unweighted mean over the four datasets, as in Tables~2 and~3 of the main paper, with Seq.\ and Win.\ totals.}
\label{tab:supp_ed_native_datasets}
\begin{tabular*}{\textwidth}{@{\extracolsep{\fill}}lrr*{12}{c}@{}}
\toprule
& & & \multicolumn{6}{c}{\textbf{CEFS}} & & & \multicolumn{3}{c}{\textbf{Timing} $|\Delta_{\rm GT}|$} & \\
\cmidrule(lr){4-9}\cmidrule(lr){12-14}
Dataset & Seq. & Win. & CEFS$\uparrow$ & $D_{\rm count}\downarrow$ & $D_{\rm pol}\downarrow$ & $D_{\rm ST}\downarrow$ & $D_{\rm async}\downarrow$ & $D_{\rm temp}\downarrow$ & EQS$\uparrow$ & P-CD$\downarrow$ & Same-ts$\downarrow$ & ISI-spike$\downarrow$ & Fano$\downarrow$ & $N_g/N_r$ \\
\midrule
DSEC & 12 & 240 & 0.685 & 0.058 & 0.029 & 0.250 & $4.24{\times}10^{-5}$ & 0.0104 & 0.991 & 3.60 & 0.0135 & $3.03{\times}10^{-6}$ & 1.23 & 0.862 \\
EDS & 2 & 120 & 0.660 & 0.045 & 0.024 & 0.287 & $2.44{\times}10^{-4}$ & 0.0194 & 0.992 & 5.75 & 0.0411 & $3.49{\times}10^{-4}$ & 0.468 & 1.20 \\
BS-ERGB & 21 & 525 & 0.590 & 0.085 & 0.032 & 0.337 & $5.82{\times}10^{-5}$ & 0.0191 & 0.990 & 4.89 & 0.0565 & 0.00714 & 0.970 & 1.04 \\
CEAR & 16 & 960 & 0.512 & 0.100 & 0.034 & 0.420 & $9.57{\times}10^{-4}$ & 0.0285 & 0.986 & 12.2 & 0.106 & 0.0280 & 0.685 & 0.891 \\
\midrule
Mean & 51 & 1845 & 0.612 & 0.072 & 0.030 & 0.324 & $3.25{\times}10^{-4}$ & 0.0194 & 0.990 & 6.61 & 0.0543 & 0.00887 & 0.838 & 0.998 \\
\bottomrule
\end{tabular*}
\end{table*}

\begin{table*}[tp]
\centering\scriptsize
\setlength{\tabcolsep}{3.5pt}
\caption{\textbf{EventDiffusion-FT under Gap-5$\times$, per dataset.} Each dataset's own EAD and its ESA and ETD fine-tuned for five-frame gaps (Sec.~\ref{sec:supp-impl-details}), seed~0. Each window is one native frame interval, five per gap, scored as in the main paper over complete valid $32{\times}32$\,px patches. We average across windows within each sequence, then across sequences within each dataset. CEFS and its components are defined in Sec.~\ref{sec:supp-cefs}. P-CD is the PECS Chamfer distance by the paper's equations. Same-ts, ISI-spike and Fano are the timing errors against the real stream of Table~3 of the main paper. $N_g/N_r$ is the generated over the real event count (best at~1). Mean: unweighted mean over the four datasets, as in Table~2 of the main paper, with Seq.\ and Win.\ totals. The Native RGB results are in Table~\ref{tab:supp_ed_native_datasets}.}
\label{tab:supp_ed_gap5_datasets}
\begin{tabular*}{\textwidth}{@{\extracolsep{\fill}}lrr*{12}{c}@{}}
\toprule
& & & \multicolumn{6}{c}{\textbf{CEFS}} & & & \multicolumn{3}{c}{\textbf{Timing} $|\Delta_{\rm GT}|$} & \\
\cmidrule(lr){4-9}\cmidrule(lr){12-14}
Dataset & Seq. & Win. & CEFS$\uparrow$ & $D_{\rm count}\downarrow$ & $D_{\rm pol}\downarrow$ & $D_{\rm ST}\downarrow$ & $D_{\rm async}\downarrow$ & $D_{\rm temp}\downarrow$ & EQS$\uparrow$ & P-CD$\downarrow$ & Same-ts$\downarrow$ & ISI-spike$\downarrow$ & Fano$\downarrow$ & $N_g/N_r$ \\
\midrule
DSEC & 12 & 240 & 0.520 & 0.214 & 0.046 & 0.307 & $3.22{\times}10^{-6}$ & 0.0100 & 0.984 & 3.89 & 0.0318 & $3.03{\times}10^{-6}$ & 1.83 & 0.697 \\
EDS & 2 & 120 & 0.549 & 0.130 & 0.040 & 0.348 & $7.98{\times}10^{-5}$ & 0.0222 & 0.986 & 6.27 & 0.0671 & $4.10{\times}10^{-4}$ & 0.655 & 1.27 \\
BS-ERGB & 21 & 525 & 0.201 & 0.540 & 0.056 & 0.561 & $2.32{\times}10^{-6}$ & 0.0252 & 0.981 & 6.91 & 0.192 & 0.00852 & 1.36 & 0.497 \\
CEAR & 16 & 960 & 0.439 & 0.187 & 0.034 & 0.450 & $1.89{\times}10^{-4}$ & 0.0282 & 0.983 & 12.8 & 0.101 & 0.0371 & 0.790 & 0.836 \\
\midrule
Mean & 51 & 1845 & 0.427 & 0.268 & 0.044 & 0.416 & $6.87{\times}10^{-5}$ & 0.0214 & 0.983 & 7.46 & 0.0980 & 0.0115 & 1.16 & 0.825 \\
\bottomrule
\end{tabular*}
\end{table*}

\subsection{Forecast Timing and Temporal Resolution}
\label{sec:supp-forecast-timing}
Both tables use EventDiffusion's RGB + events forecasts of Table~5 of the main paper, from the forecasting model of Sec.~\ref{sec:supp-forecast-model} (BS-ERGB, 21 test sequences, five frame intervals from the origin, three seeds). Table~\ref{tab:forecast_timing} isolates timestamp assignment by keeping every event's pixel, polarity and slice fixed and varying only its time within the slice. ETD lowers burst Fano error to $1.08$, compared with $1.33$ for uniform times and $1.29$ for a causal per-patch rate. It achieves lower Fano error in 54 and 52 of the 63 sequence--seed pairs, respectively. This improvement is obtained on the same forecast counts and locations, showing that ETD contributes temporal structure beyond the activity and allocation stages.

Table~\ref{tab:forecast_rebin} scores one forecast at bin widths from 20 to 1\,ms over its whole horizon, aggregating the same sampled stream each time. The count profile and polarity agreement stay nearly constant across widths, while occupancy IoU falls as each cell shrinks, as it does for any stream.
\begin{table}[tbp]
\centering\scriptsize
\caption{Forecast timing on BS-ERGB (RGB + events input, five intervals from the origin, 105 windows), mean ${\pm}$ std over three seeds. All rows use the same forecast events, with the same pixel, polarity and slice. Only their times within each slice differ. Uniform: uniform in the slice. Causal rate: per 32\,px patch, a rate linear in time with the slice's own rate at its middle and the slope from the previous slice's rate. ETD: the forecast as generated.}
\label{tab:forecast_timing}
\setlength{\tabcolsep}{3pt}
\begin{tabular*}{\columnwidth}{@{\extracolsep{\fill}}lcc@{}}
\toprule
\textbf{Timestamps} & \textbf{Same-ts}$\downarrow$ & \textbf{Fano}$\downarrow$ \\
\midrule
Uniform in slice & 0.115{\tiny$\pm$0.006} & 1.33{\tiny$\pm$0.01} \\
Causal rate      & 0.115{\tiny$\pm$0.006} & 1.29{\tiny$\pm$0.01} \\
ETD              & 0.113{\tiny$\pm$0.006} & 1.08{\tiny$\pm$0.03} \\
\bottomrule
\end{tabular*}
\end{table}

\begin{table}[tbp]
\centering\scriptsize
\caption{One forecast scored at several bin widths over its whole horizon (BS-ERGB, RGB + events input, about 187\,ms). Each seed's forecast is scored at all widths without resampling. Values are means over three seeds (EventDiffusion ${\pm}$ std). Occupancy IoU: of the (pixel, bin) cells holding an event. Count profile: total variation distance between the per-bin event counts, each stream normalized by its own total. Polarity: agreement of the majority polarity in cells occupied in both.}
\label{tab:forecast_rebin}
\setlength{\tabcolsep}{3pt}
\begin{tabular*}{\columnwidth}{@{\extracolsep{\fill}}lccc@{}}
\toprule
\textbf{Bin} & \textbf{Occupancy IoU}$\uparrow$ & \textbf{Count profile}$\downarrow$ & \textbf{Polarity}$\uparrow$ \\
\midrule
20\,ms & 0.236{\tiny$\pm$0.003} & 0.102{\tiny$\pm$0.005} & 0.581{\tiny$\pm$0.005} \\
10\,ms & 0.167{\tiny$\pm$0.003} & 0.106{\tiny$\pm$0.005} & 0.585{\tiny$\pm$0.005} \\
5\,ms  & 0.109{\tiny$\pm$0.002} & 0.107{\tiny$\pm$0.005} & 0.592{\tiny$\pm$0.006} \\
2\,ms  & 0.054{\tiny$\pm$0.001} & 0.108{\tiny$\pm$0.005} & 0.601{\tiny$\pm$0.007} \\
1\,ms  & 0.028{\tiny$\pm$0.001} & 0.109{\tiny$\pm$0.005} & 0.606{\tiny$\pm$0.006} \\
\bottomrule
\end{tabular*}
\end{table}

\subsection{Downstream Models Trained on Generated Events}
\label{sec:supp-downstream-trained}
Table~\ref{tab:downstream_trained} asks whether generated events can replace real recordings as training data. EventDiffusion-FT generates DSEC's 41 training sequences whole from their RGB frames, each interval following the stream generated before it.
E2VID~\cite{rebecq2019e2vid} is trained from scratch on these events with its authors' recipe (320k iterations, LPIPS and temporal-consistency losses, the final checkpoint) and scored on the real events of the 12 test recordings against their RGB frames warped onto the event camera.
ESS~\cite{sun2022ess} trains its segmentation decoder with the authors' settings on the events of DSEC-Semantic's 8 training sequences, over a frozen E2VID trained on generated events, and is scored after the 50th epoch on the real events of its 3 test sequences. We train E2VID and ESS in the same way on events generated separately by ESIM, DVS-Voltmeter and V2CE.

\begin{table}[tbp]
\centering\scriptsize
\caption{Downstream models trained on generated events and tested on real events. Top: E2VID trained from scratch on the events of DSEC's 41 training sequences, scored on the 12 DSEC test recordings against their frames without normalization. Bottom: the ESS segmentation decoder trained on the events of DSEC-Semantic's 8 training sequences over a frozen E2VID trained on the same kind of events, scored on its 3 test recordings. EventDiffusion-FT: mean ${\pm}$ std over three training seeds. Each reference row uses one seed. The released E2VID was trained on ESIM events from COCO images. The published ESS model was trained on real events with the released E2VID.}
\label{tab:downstream_trained}
\setlength{\tabcolsep}{4pt}
\begin{tabular*}{\columnwidth}{@{\extracolsep{\fill}}lccc@{}}
\toprule
\textbf{E2VID training events} & \textbf{MSE}$\downarrow$ & \textbf{SSIM}$\uparrow$ & \textbf{LPIPS}$\downarrow$ \\
\midrule
Released E2VID (trained on ESIM) & 0.0842 & 0.423 & 0.525 \\
Real events & 0.0183 & 0.800 & 0.206 \\
\midrule
ESIM & 0.1666 & 0.294 & 0.673 \\
DVS-Voltmeter & 0.1026 & 0.304 & 0.669 \\
V2CE & 0.0632 & 0.564 & 0.410 \\
EventDiffusion-FT & 0.0485{\tiny$\pm$0.0115} & 0.725{\tiny$\pm$0.017} & 0.313{\tiny$\pm$0.015} \\
\bottomrule
\end{tabular*}

\vspace{6pt}
\begin{tabular*}{\columnwidth}{@{\extracolsep{\fill}}lcc@{}}
\toprule
\textbf{ESS training events} & \textbf{mIoU (\%)}$\uparrow$ & \textbf{Acc.\ (\%)}$\uparrow$ \\
\midrule
Real events (released E2VID(ESIM)) & 51.1 & 88.7 \\
Real events & 55.9 & 91.1 \\
\midrule
ESIM & 34.5 & 78.2 \\
DVS-Voltmeter & 33.6 & 76.7 \\
V2CE & 45.1 & 85.3 \\
EventDiffusion-FT & 54.3{\tiny$\pm$0.6} & 90.2{\tiny$\pm$0.4} \\
\bottomrule
\end{tabular*}
\end{table}

\begin{figure*}[t]
\centering\scriptsize
\setlength{\tabcolsep}{1pt}
\newcommand{\dssw}[2]{\tikz[baseline=0.1ex]\fill[fill={rgb,255:#1},draw=black!30,line width=0.2pt] (0,0) rectangle (1.4ex,1.4ex);\,#2}
\begin{tabular}{@{}cc@{\hspace{8pt}}cc@{}}
\multicolumn{2}{@{}c}{\textbf{E2VID}, against the frame (LPIPS$\downarrow$)} & \multicolumn{2}{c@{}}{\textbf{ESS}, against the label (mIoU\,\%$\uparrow$)} \\[2pt]
\multicolumn{2}{@{}l}{(a) \texttt{zurich\_city\_14\_c}, frame 459} & \multicolumn{2}{l@{}}{label 492, 1.65\,s later} \\
\includegraphics[width=0.240\textwidth]{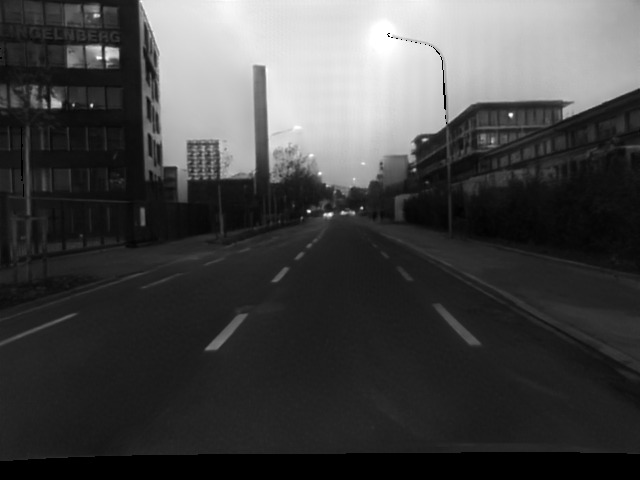} & \includegraphics[width=0.240\textwidth]{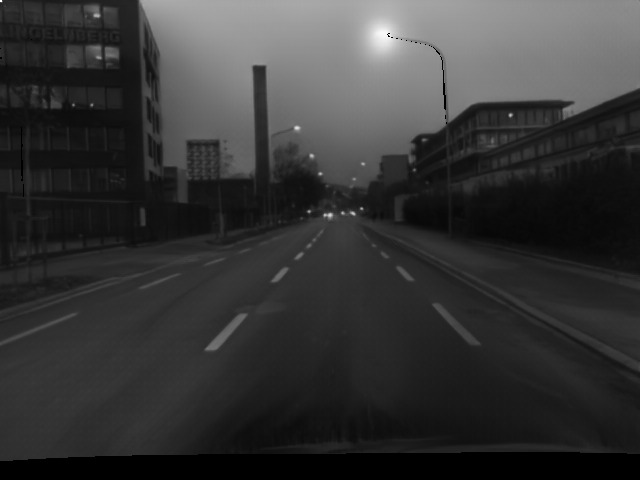} & \includegraphics[width=0.240\textwidth]{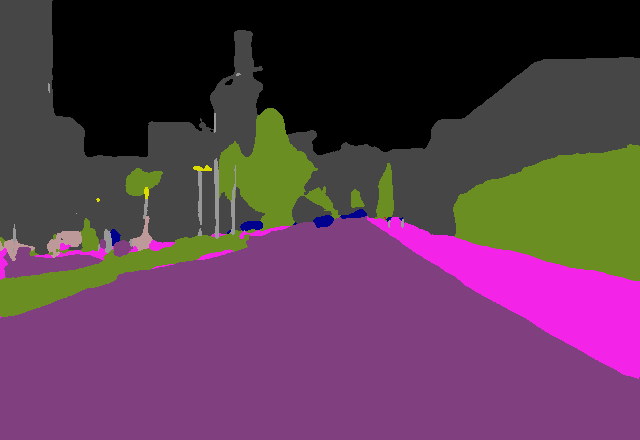} & \includegraphics[width=0.240\textwidth]{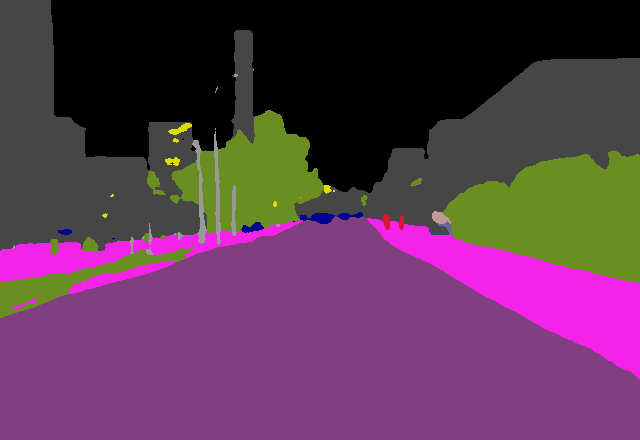} \\
Real events: \textbf{0.147} & EventDiffusion-FT: 0.186 & Real events: 38.9 & EventDiffusion-FT: \textbf{45.1} \\
\includegraphics[width=0.240\textwidth]{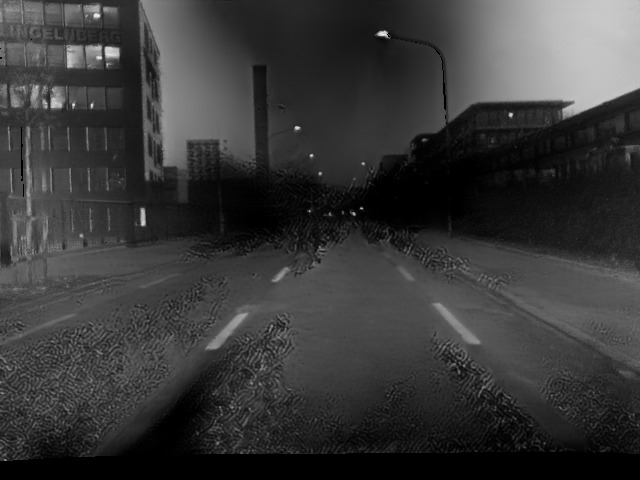} & \includegraphics[width=0.240\textwidth]{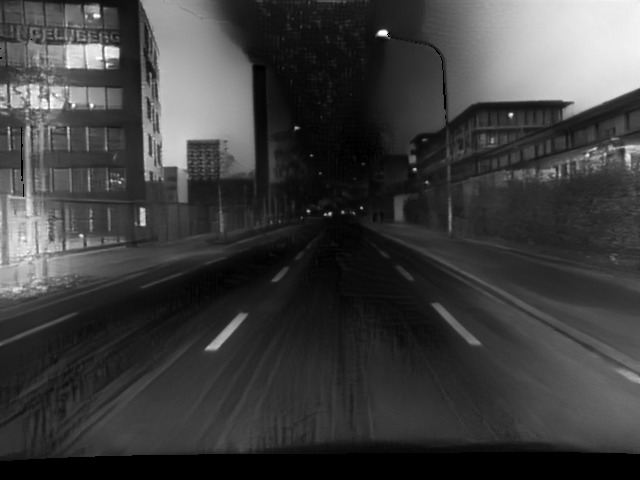} & \includegraphics[width=0.240\textwidth]{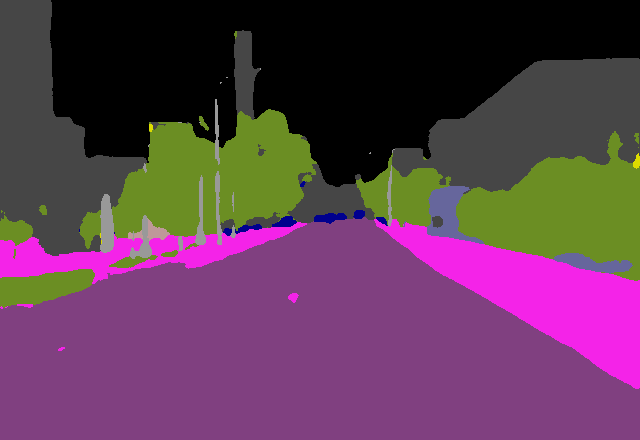} & \includegraphics[width=0.240\textwidth]{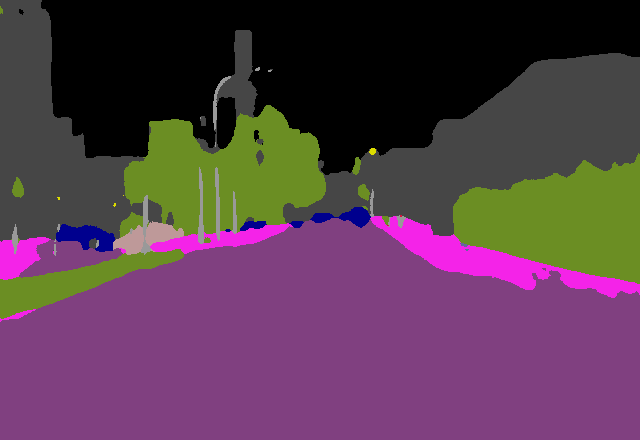} \\
Released: 0.491 & V2CE: 0.416 & Released: 37.6 & V2CE: 35.1 \\[4pt]
\multicolumn{2}{@{}l}{(b) \texttt{zurich\_city\_14\_c}, frame 833} & \multicolumn{2}{l@{}}{label 1084, 12.55\,s later} \\
\includegraphics[width=0.240\textwidth]{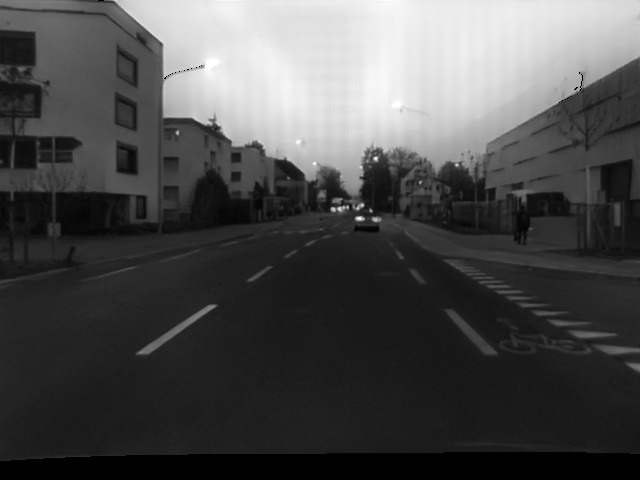} & \includegraphics[width=0.240\textwidth]{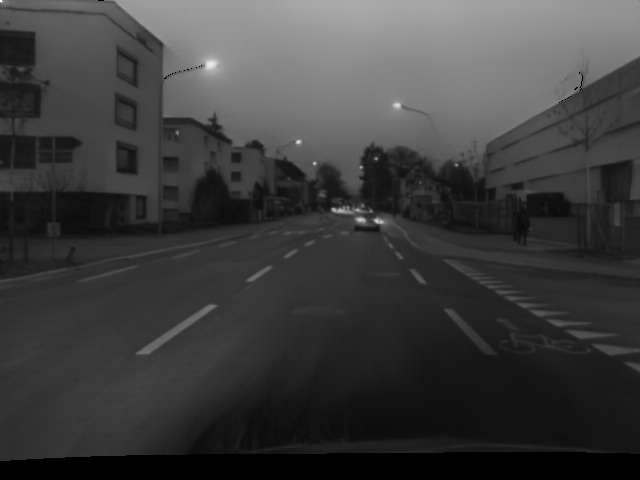} & \includegraphics[width=0.240\textwidth]{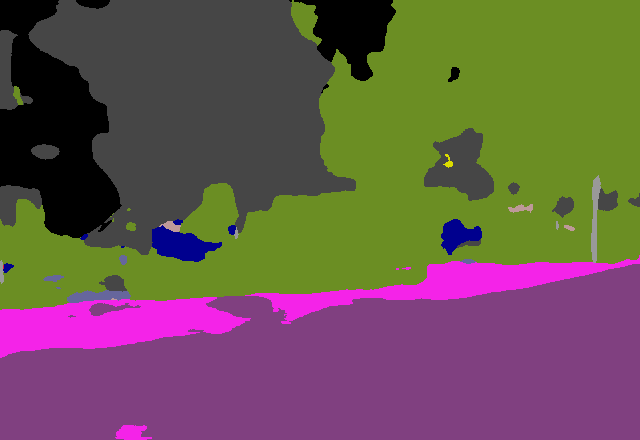} & \includegraphics[width=0.240\textwidth]{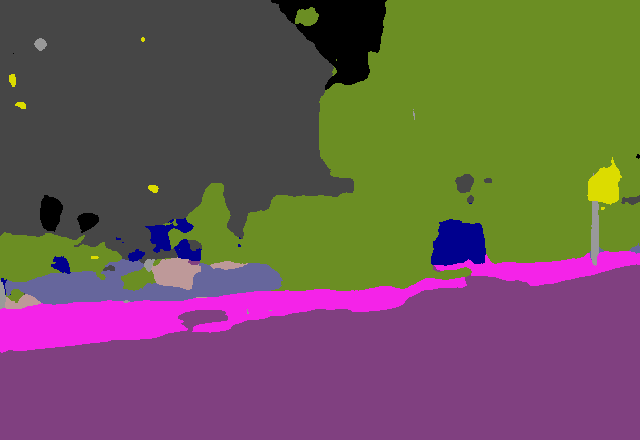} \\
Real events: \textbf{0.157} & EventDiffusion-FT: 0.188 & Real events: 34.9 & EventDiffusion-FT: \textbf{56.5} \\
\includegraphics[width=0.240\textwidth]{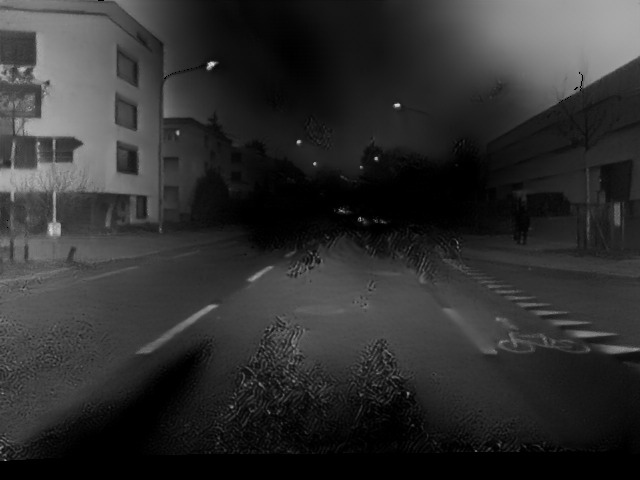} & \includegraphics[width=0.240\textwidth]{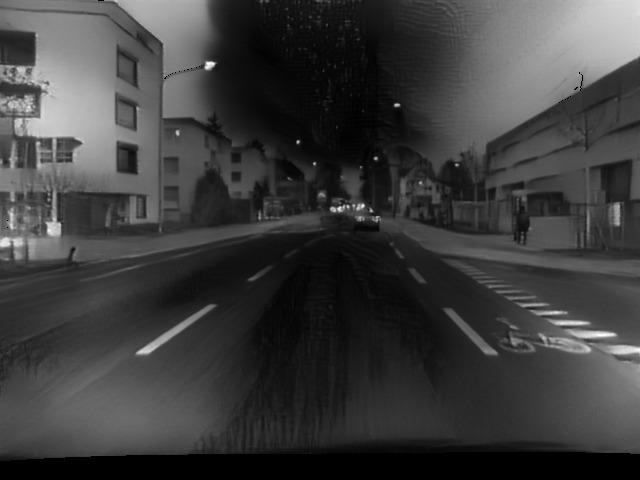} & \includegraphics[width=0.240\textwidth]{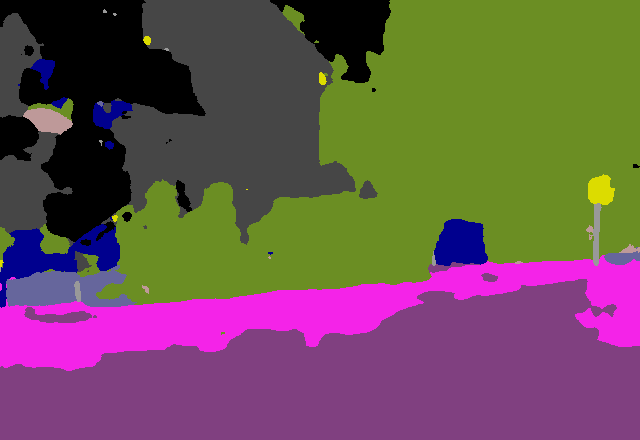} & \includegraphics[width=0.240\textwidth]{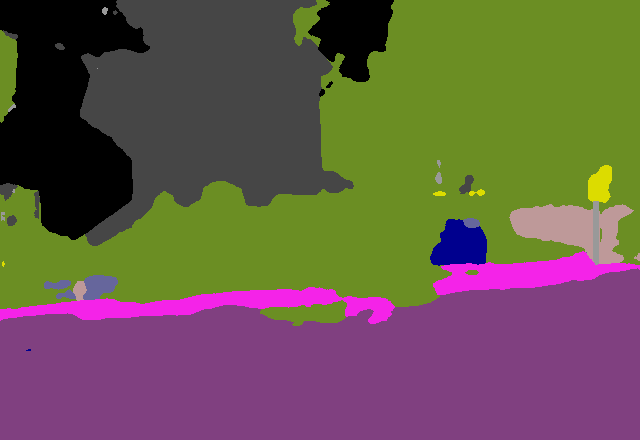} \\
Released: 0.487 & V2CE: 0.405 & Released: 39.8 & V2CE: 41.9 \\
\end{tabular}

\vspace{3pt}
\dssw{red,0;green,0;blue,0}{background}\quad
\dssw{red,70;green,70;blue,70}{building}\quad
\dssw{red,190;green,153;blue,153}{fence}\quad
\dssw{red,220;green,20;blue,60}{person}\quad
\dssw{red,153;green,153;blue,153}{pole}\quad
\dssw{red,128;green,64;blue,128}{road}\quad
\dssw{red,244;green,35;blue,232}{sidewalk}\quad
\dssw{red,107;green,142;blue,35}{vegetation}\quad
\dssw{red,0;green,0;blue,142}{car}\quad
\dssw{red,102;green,102;blue,156}{wall}\quad
\dssw{red,220;green,220;blue,0}{traffic sign}
\caption{Downstream outputs on DSEC's real test events in \texttt{zurich\_city\_14\_c}, in E2VID (left) and ESS (right). The ESS label shows the same street 1.65\,s after the E2VID frame in (a) and 12.55\,s after it in (b), further along the drive. The panel labels identify the training source: real recordings, EventDiffusion-FT, the released checkpoints or V2CE. The EventDiffusion-FT panels use E2VID seed 1 and the ESS model that uses it as a frozen encoder. The released E2VID weights were trained on ESIM events from COCO images. The released ESS decoder was trained on real events with that E2VID model. E2VID outputs are shown without normalization and black outside the scored area. Numbers are per-frame LPIPS (E2VID) and mIoU over the classes in the label or the prediction (ESS). The best score in each grid is in bold. Both ESS labels come from frames where all three EventDiffusion-FT ESS seeds score above the real-events, V2CE, DVS-Voltmeter and ESIM models. For E2VID, the model trained on real events stays closer to the frame.}
\label{fig:supp_downstream_scenes}
\end{figure*}

Figure~\ref{fig:supp_downstream_scenes} shows two test moments through both tasks, for models trained on real events, EventDiffusion-FT's events, the released checkpoint and V2CE's events.

\begin{table*}[p]
\centering\scriptsize
\setlength{\tabcolsep}{3pt}
\caption{\textbf{EventDiffusion-FT under Native RGB, per sequence.} Metrics and scoring as in Table~\ref{tab:supp_ed_native_datasets}. Each value is the sequence's mean over its windows (Win.). The \textit{Mean} row averages over sequences in the dataset. We use the original sequence names. The BS-ERGB sequences are in the \texttt{1\_TEST} folder. $^{\dagger}$Night recording.}
\label{tab:supp_ed_native_sequences}
\begin{tabular*}{\textwidth}{@{\extracolsep{\fill}}lr*{10}{c}@{}}
\toprule
Sequence & Win. & CEFS$\uparrow$ & $D_{\rm count}\downarrow$ & $D_{\rm pol}\downarrow$ & $D_{\rm ST}\downarrow$ & $D_{\rm async}\downarrow$ & $D_{\rm temp}\downarrow$ & EQS$\uparrow$ & P-CD$\downarrow$ & Fano$\downarrow$ & $N_g/N_r$ \\
\midrule
\multicolumn{12}{@{}l}{\textbf{DSEC}} \\
\texttt{interlaken\_00\_a} & 20 & 0.699 & 0.066 & 0.072 & 0.190 & $1.31{\times}10^{-5}$ & 0.00774 & 0.991 & 2.54 & 1.08 & 0.775 \\
\texttt{interlaken\_00\_b} & 20 & 0.674 & 0.061 & 0.039 & 0.250 & $1.82{\times}10^{-5}$ & 0.00609 & 0.985 & 3.26 & 1.77 & 1.01 \\
\texttt{interlaken\_01\_a} & 20 & 0.717 & 0.015 & 0.046 & 0.233 & $3.62{\times}10^{-5}$ & 0.0112 & 0.989 & 3.01 & 1.41 & 0.925 \\
\texttt{thun\_01\_a} & 20 & 0.742 & 0.041 & 0.039 & 0.193 & $7.47{\times}10^{-5}$ & 0.00444 & 0.994 & 2.51 & 0.550 & 0.823 \\
\texttt{thun\_01\_b} & 20 & 0.687 & 0.083 & 0.019 & 0.235 & $6.84{\times}10^{-6}$ & 0.00604 & 0.991 & 3.03 & 1.39 & 0.753 \\
\texttt{zurich\_city\_13\_a} & 20 & 0.719 & 0.006 & 0.010 & 0.267 & $4.01{\times}10^{-5}$ & 0.00683 & 0.994 & 3.22 & 2.63 & 0.954 \\
\texttt{zurich\_city\_13\_b} & 20 & 0.718 & 0.072 & 0.010 & 0.216 & $6.21{\times}10^{-6}$ & 0.00527 & 0.993 & 2.84 & 1.79 & 0.769 \\
\texttt{zurich\_city\_14\_a} & 20 & 0.580 & 0.085 & 0.006 & 0.358 & $3.39{\times}10^{-7}$ & 0.0144 & 0.992 & 3.38 & 1.74 & 0.749 \\
\texttt{zurich\_city\_14\_b} & 20 & 0.687 & 0.014 & 0.008 & 0.296 & $2.28{\times}10^{-5}$ & 0.00691 & 0.992 & 3.91 & 0.301 & 1.08 \\
\texttt{zurich\_city\_14\_c} & 20 & 0.748 & 0.002 & 0.011 & 0.239 & $1.97{\times}10^{-4}$ & 0.00835 & 0.992 & 2.88 & 0.896 & 1.03 \\
\texttt{zurich\_city\_15\_a} & 20 & 0.782 & 0.024 & 0.010 & 0.189 & $8.40{\times}10^{-5}$ & 0.00391 & 0.994 & 2.37 & 0.861 & 0.862 \\
\texttt{zurich\_city\_12\_a}$^{\dagger}$ & 20 & 0.466 & 0.228 & 0.074 & 0.330 & $9.45{\times}10^{-6}$ & 0.0431 & 0.981 & 10.3 & 0.315 & 0.618 \\
\textit{Mean} & 240 & 0.685 & 0.058 & 0.029 & 0.250 & $4.24{\times}10^{-5}$ & 0.0104 & 0.991 & 3.60 & 1.23 & 0.862 \\
\midrule
\multicolumn{12}{@{}l}{\textbf{EDS}} \\
\texttt{13\_airplane} & 60 & 0.648 & 0.068 & 0.027 & 0.283 & $4.63{\times}10^{-4}$ & 0.0249 & 0.991 & 6.21 & 0.588 & 1.27 \\
\texttt{15\_apartment\_day} & 60 & 0.673 & 0.023 & 0.021 & 0.292 & $2.48{\times}10^{-5}$ & 0.0138 & 0.993 & 5.30 & 0.349 & 1.13 \\
\textit{Mean} & 120 & 0.660 & 0.045 & 0.024 & 0.287 & $2.44{\times}10^{-4}$ & 0.0194 & 0.992 & 5.75 & 0.468 & 1.20 \\
\midrule
\multicolumn{12}{@{}l}{\textbf{BS-ERGB}} \\
\texttt{acquarium\_08} & 25 & 0.642 & 0.014 & 0.034 & 0.319 & $4.99{\times}10^{-6}$ & 0.0174 & 0.990 & 3.80 & 0.812 & 0.936 \\
\texttt{ball\_05} & 25 & 0.703 & 0.008 & 0.017 & 0.273 & $1.75{\times}10^{-5}$ & 0.0125 & 0.991 & 6.32 & 0.849 & 1.05 \\
\texttt{ball\_06} & 25 & 0.675 & 0.045 & 0.015 & 0.275 & $4.69{\times}10^{-5}$ & 0.0196 & 0.987 & 3.07 & 0.806 & 1.23 \\
\texttt{basket\_08} & 25 & 0.704 & 0.010 & 0.027 & 0.264 & $8.10{\times}10^{-5}$ & 0.0139 & 0.992 & 2.79 & 0.925 & 0.953 \\
\texttt{basket\_09} & 25 & 0.691 & 0.011 & 0.025 & 0.278 & $9.06{\times}10^{-5}$ & 0.0167 & 0.989 & 2.93 & 0.807 & 0.950 \\
\texttt{candies\_03} & 25 & 0.520 & 0.062 & 0.033 & 0.419 & $1.28{\times}10^{-5}$ & 0.0234 & 0.991 & 4.46 & 0.673 & 1.16 \\
\texttt{eggs\_04} & 25 & 0.659 & 0.016 & 0.030 & 0.303 & $3.67{\times}10^{-6}$ & 0.0198 & 0.993 & 5.45 & 0.994 & 0.893 \\
\texttt{elastic\_bands\_01} & 25 & 0.249 & 0.520 & 0.048 & 0.456 & $1.43{\times}10^{-7}$ & 0.0418 & 0.989 & 3.61 & 3.40 & 0.434 \\
\texttt{fire\_02} & 25 & 0.613 & 0.037 & 0.029 & 0.334 & $1.65{\times}10^{-4}$ & 0.0275 & 0.986 & 6.28 & 0.686 & 1.09 \\
\texttt{football\_04} & 25 & 0.715 & 0.011 & 0.023 & 0.255 & $2.02{\times}10^{-5}$ & 0.0122 & 0.993 & 2.66 & 0.909 & 1.08 \\
\texttt{horse\_20} & 25 & 0.600 & 0.118 & 0.023 & 0.299 & $4.54{\times}10^{-5}$ & 0.0159 & 0.992 & 7.14 & 0.530 & 1.41 \\
\texttt{jacket\_03} & 25 & 0.696 & 0.030 & 0.040 & 0.247 & $4.61{\times}10^{-5}$ & 0.0136 & 0.994 & 5.21 & 1.39 & 0.845 \\
\texttt{may28\_axe\_01} & 25 & 0.302 & 0.244 & 0.049 & 0.592 & $2.55{\times}10^{-5}$ & 0.0346 & 0.993 & 15.3 & 0.198 & 0.663 \\
\texttt{may29\_redbull\_01} & 25 & 0.639 & 0.021 & 0.021 & 0.329 & $3.01{\times}10^{-5}$ & 0.0125 & 0.987 & 3.84 & 0.430 & 1.07 \\
\texttt{may29\_water\_tank\_pouring\_02} & 25 & 0.710 & 0.034 & 0.018 & 0.246 & $1.03{\times}10^{-5}$ & 0.0118 & 0.991 & 2.51 & 1.57 & 0.842 \\
\texttt{orange\_juice\_02} & 25 & 0.578 & 0.041 & 0.050 & 0.362 & $1.10{\times}10^{-5}$ & 0.0165 & 0.987 & 4.71 & 0.981 & 1.03 \\
\texttt{paprika\_1000\_gain\_control\_02} & 25 & 0.667 & 0.054 & 0.026 & 0.272 & $1.34{\times}10^{-4}$ & 0.00945 & 0.983 & 2.78 & 1.37 & 1.24 \\
\texttt{pen\_03} & 25 & 0.528 & 0.040 & 0.050 & 0.415 & $4.12{\times}10^{-4}$ & 0.0202 & 0.986 & 3.96 & 0.853 & 1.07 \\
\texttt{rope\_jumping\_01} & 25 & 0.504 & 0.110 & 0.069 & 0.389 & $1.69{\times}10^{-5}$ & 0.0104 & 0.995 & 7.25 & 0.612 & 1.39 \\
\texttt{street\_crossing\_08} & 25 & 0.633 & 0.056 & 0.020 & 0.309 & $1.23{\times}10^{-5}$ & 0.0204 & 0.986 & 3.96 & 0.777 & 0.807 \\
\texttt{watermelon\_01} & 25 & 0.359 & 0.312 & 0.030 & 0.450 & $3.49{\times}10^{-5}$ & 0.0320 & 0.984 & 4.78 & 0.799 & 1.80 \\
\textit{Mean} & 525 & 0.590 & 0.085 & 0.032 & 0.337 & $5.82{\times}10^{-5}$ & 0.0191 & 0.990 & 4.89 & 0.970 & 1.04 \\
\midrule
\multicolumn{12}{@{}l}{\textbf{CEAR}} \\
\texttt{backflip/env8\_backflip1} & 60 & 0.455 & 0.130 & 0.019 & 0.462 & $4.13{\times}10^{-4}$ & 0.0407 & 0.985 & 10.7 & 0.437 & 0.770 \\
\texttt{backflip/env8\_backflip2} & 60 & 0.301 & 0.440 & 0.051 & 0.572 & $3.59{\times}10^{-4}$ & 0.0640 & 0.978 & 21.8 & 1.80 & 0.644 \\
\texttt{indoor/classroom\_blinking\_comb} & 60 & 0.543 & 0.044 & 0.018 & 0.412 & 0.00205 & 0.0354 & 0.986 & 10.1 & 0.706 & 1.00 \\
\texttt{indoor/classroom\_blinking\_trot} & 60 & 0.598 & 0.025 & 0.018 & 0.373 & 0.00210 & 0.0229 & 0.989 & 10.6 & 0.884 & 1.02 \\
\texttt{indoor/classroom\_hdr\_comb} & 60 & 0.429 & 0.107 & 0.051 & 0.492 & $4.78{\times}10^{-4}$ & 0.0239 & 0.983 & 13.4 & 0.736 & 0.816 \\
\texttt{indoor/classroom\_hdr\_trot} & 60 & 0.305 & 0.164 & 0.129 & 0.582 & $2.20{\times}10^{-4}$ & 0.0205 & 0.982 & 18.2 & 0.642 & 0.720 \\
\texttt{indoor/classroom\_well-lit\_comb} & 60 & 0.561 & 0.063 & 0.031 & 0.379 & 0.00116 & 0.0189 & 0.987 & 10.1 & 0.597 & 0.820 \\
\texttt{indoor/classroom\_well-lit\_trot} & 60 & 0.583 & 0.021 & 0.016 & 0.389 & 0.00185 & 0.0290 & 0.988 & 10.9 & 0.810 & 1.01 \\
\texttt{outdoor/between-buildings\_day\_comb} & 60 & 0.658 & 0.040 & 0.010 & 0.295 & 0.00227 & 0.0336 & 0.991 & 8.05 & 0.714 & 0.910 \\
\texttt{outdoor/between-buildings\_day\_trot} & 60 & 0.422 & 0.241 & 0.028 & 0.433 & $1.89{\times}10^{-6}$ & 0.0336 & 0.984 & 12.5 & 0.488 & 0.631 \\
\texttt{outdoor/between-buildings\_night\_comb} & 60 & 0.569 & 0.015 & 0.030 & 0.398 & $1.71{\times}10^{-5}$ & 0.0224 & 0.990 & 11.3 & 0.288 & 0.966 \\
\texttt{outdoor/between-buildings\_night\_trot} & 60 & 0.557 & 0.025 & 0.025 & 0.409 & $2.60{\times}10^{-5}$ & 0.0237 & 0.989 & 11.4 & 0.306 & 0.974 \\
\texttt{outdoor/grass1\_day\_comb} & 60 & 0.615 & 0.099 & 0.009 & 0.305 & 0.00169 & 0.0316 & 0.987 & 8.26 & 0.649 & 0.895 \\
\texttt{outdoor/grass1\_day\_trot} & 60 & 0.655 & 0.076 & 0.010 & 0.274 & 0.00171 & 0.0252 & 0.988 & 7.66 & 0.666 & 0.837 \\
\texttt{outdoor/grass1\_night\_comb} & 60 & 0.536 & 0.016 & 0.045 & 0.426 & $7.64{\times}10^{-4}$ & 0.0132 & 0.981 & 13.1 & 0.729 & 0.964 \\
\texttt{outdoor/grass1\_night\_trot} & 60 & 0.411 & 0.090 & 0.048 & 0.524 & $2.00{\times}10^{-4}$ & 0.0174 & 0.984 & 16.3 & 0.503 & 1.28 \\
\textit{Mean} & 960 & 0.512 & 0.100 & 0.034 & 0.420 & $9.57{\times}10^{-4}$ & 0.0285 & 0.986 & 12.2 & 0.685 & 0.891 \\
\bottomrule
\end{tabular*}
\end{table*}

\begin{table*}[p]
\centering\scriptsize
\setlength{\tabcolsep}{3pt}
\caption{\textbf{EventDiffusion-FT under Gap-5$\times$, per sequence.} Metrics and scoring as in Table~\ref{tab:supp_ed_gap5_datasets}. Each value is the sequence's mean over its windows (Win.). The \textit{Mean} row averages over sequences in the dataset. We use the original sequence names. The BS-ERGB sequences are in the \texttt{1\_TEST} folder. $^{\dagger}$Night recording.}
\label{tab:supp_ed_gap5_sequences}
\begin{tabular*}{\textwidth}{@{\extracolsep{\fill}}lr*{10}{c}@{}}
\toprule
Sequence & Win. & CEFS$\uparrow$ & $D_{\rm count}\downarrow$ & $D_{\rm pol}\downarrow$ & $D_{\rm ST}\downarrow$ & $D_{\rm async}\downarrow$ & $D_{\rm temp}\downarrow$ & EQS$\uparrow$ & P-CD$\downarrow$ & Fano$\downarrow$ & $N_g/N_r$ \\
\midrule
\multicolumn{12}{@{}l}{\textbf{DSEC}} \\
\texttt{interlaken\_00\_a} & 20 & 0.416 & 0.360 & 0.137 & 0.250 & $5.97{\times}10^{-8}$ & 0.00806 & 0.979 & 2.98 & 1.97 & 0.524 \\
\texttt{interlaken\_00\_b} & 20 & 0.568 & 0.114 & 0.023 & 0.341 & $1.84{\times}10^{-5}$ & 0.00749 & 0.977 & 3.68 & 1.89 & 1.11 \\
\texttt{interlaken\_01\_a} & 20 & 0.611 & 0.091 & 0.018 & 0.312 & $1.83{\times}10^{-6}$ & 0.0117 & 0.979 & 3.51 & 1.96 & 0.757 \\
\texttt{thun\_01\_a} & 20 & 0.631 & 0.130 & 0.051 & 0.234 & $1.14{\times}10^{-7}$ & 0.00396 & 0.991 & 2.73 & 0.902 & 0.697 \\
\texttt{thun\_01\_b} & 20 & 0.503 & 0.258 & 0.027 & 0.302 & $3.74{\times}10^{-8}$ & 0.00606 & 0.986 & 3.40 & 1.65 & 0.589 \\
\texttt{zurich\_city\_13\_a} & 20 & 0.470 & 0.252 & 0.017 & 0.358 & $1.54{\times}10^{-7}$ & 0.00746 & 0.985 & 3.87 & 3.11 & 0.594 \\
\texttt{zurich\_city\_13\_b} & 20 & 0.640 & 0.128 & 0.017 & 0.251 & $4.13{\times}10^{-6}$ & 0.00502 & 0.988 & 3.02 & 1.86 & 0.712 \\
\texttt{zurich\_city\_14\_a} & 20 & 0.509 & 0.124 & 0.011 & 0.408 & $1.35{\times}10^{-6}$ & 0.0137 & 0.985 & 3.64 & 2.15 & 0.710 \\
\texttt{zurich\_city\_14\_b} & 20 & 0.630 & 0.006 & 0.012 & 0.356 & $8.02{\times}10^{-6}$ & 0.00574 & 0.987 & 4.22 & 0.257 & 1.00 \\
\texttt{zurich\_city\_14\_c} & 20 & 0.603 & 0.086 & 0.021 & 0.323 & $2.00{\times}10^{-6}$ & 0.00957 & 0.985 & 3.40 & 0.865 & 0.764 \\
\texttt{zurich\_city\_15\_a} & 20 & 0.541 & 0.223 & 0.049 & 0.267 & $2.84{\times}10^{-9}$ & 0.00677 & 0.987 & 2.83 & 0.735 & 0.616 \\
\texttt{zurich\_city\_12\_a}$^{\dagger}$ & 20 & 0.120 & 0.794 & 0.167 & 0.277 & $2.59{\times}10^{-6}$ & 0.0350 & 0.972 & 9.40 & 4.62 & 0.288 \\
\textit{Mean} & 240 & 0.520 & 0.214 & 0.046 & 0.307 & $3.22{\times}10^{-6}$ & 0.0100 & 0.984 & 3.89 & 1.83 & 0.697 \\
\midrule
\multicolumn{12}{@{}l}{\textbf{EDS}} \\
\texttt{13\_airplane} & 60 & 0.523 & 0.176 & 0.039 & 0.353 & $1.41{\times}10^{-4}$ & 0.0289 & 0.985 & 6.79 & 0.971 & 1.30 \\
\texttt{15\_apartment\_day} & 60 & 0.574 & 0.084 & 0.041 & 0.342 & $1.90{\times}10^{-5}$ & 0.0154 & 0.986 & 5.76 & 0.339 & 1.24 \\
\textit{Mean} & 120 & 0.549 & 0.130 & 0.040 & 0.348 & $7.98{\times}10^{-5}$ & 0.0222 & 0.986 & 6.27 & 0.655 & 1.27 \\
\midrule
\multicolumn{12}{@{}l}{\textbf{BS-ERGB}} \\
\texttt{acquarium\_08} & 25 & 0.223 & 0.584 & 0.027 & 0.456 & $3.60{\times}10^{-11}$ & 0.0134 & 0.983 & 5.01 & 1.11 & 0.399 \\
\texttt{ball\_05} & 25 & 0.257 & 0.418 & 0.046 & 0.556 & $1.79{\times}10^{-6}$ & 0.0197 & 0.987 & 7.96 & 1.20 & 0.532 \\
\texttt{ball\_06} & 25 & 0.214 & 0.487 & 0.062 & 0.588 & $2.81{\times}10^{-6}$ & 0.0344 & 0.981 & 5.62 & 1.16 & 0.496 \\
\texttt{basket\_08} & 25 & 0.136 & 0.678 & 0.068 & 0.549 & $1.11{\times}10^{-9}$ & 0.0203 & 0.982 & 4.98 & 1.31 & 0.350 \\
\texttt{basket\_09} & 25 & 0.115 & 0.733 & 0.059 & 0.578 & $1.23{\times}10^{-6}$ & 0.0281 & 0.976 & 5.41 & 1.36 & 0.333 \\
\texttt{candies\_03} & 25 & 0.137 & 0.569 & 0.066 & 0.677 & $3.52{\times}10^{-7}$ & 0.0332 & 0.985 & 7.50 & 0.916 & 0.423 \\
\texttt{eggs\_04} & 25 & 0.482 & 0.184 & 0.025 & 0.391 & $3.49{\times}10^{-6}$ & 0.0160 & 0.986 & 5.45 & 1.13 & 0.734 \\
\texttt{elastic\_bands\_01} & 25 & 0.0191 & 0.9393 & 0.075 & 0.678 & $6.70{\times}10^{-10}$ & 0.0404 & 0.984 & 6.00 & 4.01 & 0.183 \\
\texttt{fire\_02} & 25 & 0.191 & 0.614 & 0.082 & 0.475 & $2.25{\times}10^{-8}$ & 0.0298 & 0.983 & 7.28 & 1.86 & 0.386 \\
\texttt{football\_04} & 25 & 0.147 & 0.600 & 0.042 & 0.628 & $7.35{\times}10^{-9}$ & 0.0205 & 0.983 & 6.30 & 1.33 & 0.397 \\
\texttt{horse\_20} & 25 & 0.137 & 0.594 & 0.106 & 0.667 & $2.11{\times}10^{-7}$ & 0.0252 & 0.971 & 9.48 & 0.421 & 0.405 \\
\texttt{jacket\_03} & 25 & 0.343 & 0.367 & 0.027 & 0.437 & $2.88{\times}10^{-6}$ & 0.0185 & 0.989 & 6.12 & 1.82 & 0.547 \\
\texttt{may28\_axe\_01} & 25 & 0.181 & 0.303 & 0.047 & 0.728 & $8.63{\times}10^{-6}$ & 0.0361 & 0.994 & 15.9 & 0.198 & 0.629 \\
\texttt{may29\_redbull\_01} & 25 & 0.0899 & 0.762 & 0.085 & 0.620 & $4.16{\times}10^{-10}$ & 0.0221 & 0.977 & 6.07 & 0.468 & 0.302 \\
\texttt{may29\_water\_tank\_pouring\_02} & 25 & 0.121 & 0.811 & 0.045 & 0.344 & $5.13{\times}10^{-8}$ & 0.0168 & 0.975 & 3.25 & 2.45 & 0.275 \\
\texttt{orange\_juice\_02} & 25 & 0.131 & 0.600 & 0.059 & 0.647 & $1.68{\times}10^{-8}$ & 0.0244 & 0.981 & 7.17 & 1.49 & 0.389 \\
\texttt{paprika\_1000\_gain\_control\_02} & 25 & 0.245 & 0.580 & 0.045 & 0.404 & $1.76{\times}10^{-9}$ & 0.0179 & 0.965 & 4.00 & 1.57 & 0.406 \\
\texttt{pen\_03} & 25 & 0.0587 & 0.777 & 0.047 & 0.727 & $7.21{\times}10^{-6}$ & 0.0422 & 0.971 & 10.8 & 0.871 & 0.945 \\
\texttt{rope\_jumping\_01} & 25 & 0.452 & 0.015 & 0.032 & 0.525 & $8.53{\times}10^{-6}$ & 0.00981 & 0.991 & 8.25 & 0.669 & 0.990 \\
\texttt{street\_crossing\_08} & 25 & 0.346 & 0.324 & 0.065 & 0.441 & $1.40{\times}10^{-6}$ & 0.0190 & 0.978 & 4.65 & 1.52 & 0.593 \\
\texttt{watermelon\_01} & 25 & 0.190 & 0.409 & 0.059 & 0.659 & $1.02{\times}10^{-5}$ & 0.0409 & 0.975 & 7.90 & 1.61 & 0.714 \\
\textit{Mean} & 525 & 0.201 & 0.540 & 0.056 & 0.561 & $2.32{\times}10^{-6}$ & 0.0252 & 0.981 & 6.91 & 1.36 & 0.497 \\
\midrule
\multicolumn{12}{@{}l}{\textbf{CEAR}} \\
\texttt{backflip/env8\_backflip1} & 60 & 0.424 & 0.169 & 0.021 & 0.468 & $4.25{\times}10^{-5}$ & 0.0377 & 0.982 & 10.7 & 0.643 & 0.757 \\
\texttt{backflip/env8\_backflip2} & 60 & 0.107 & 0.782 & 0.069 & 0.676 & $1.39{\times}10^{-5}$ & 0.0617 & 0.972 & 28.2 & 2.25 & 0.334 \\
\texttt{indoor/classroom\_blinking\_comb} & 60 & 0.461 & 0.139 & 0.031 & 0.440 & $3.92{\times}10^{-4}$ & 0.0362 & 0.982 & 10.5 & 0.641 & 0.945 \\
\texttt{indoor/classroom\_blinking\_trot} & 60 & 0.525 & 0.098 & 0.026 & 0.398 & $6.46{\times}10^{-4}$ & 0.0243 & 0.985 & 10.8 & 0.648 & 1.01 \\
\texttt{indoor/classroom\_hdr\_comb} & 60 & 0.370 & 0.126 & 0.064 & 0.543 & $1.43{\times}10^{-4}$ & 0.0242 & 0.983 & 13.4 & 1.01 & 0.853 \\
\texttt{indoor/classroom\_hdr\_trot} & 60 & 0.287 & 0.271 & 0.050 & 0.588 & $1.52{\times}10^{-4}$ & 0.0194 & 0.983 & 18.5 & 0.762 & 0.599 \\
\texttt{indoor/classroom\_well-lit\_comb} & 60 & 0.492 & 0.110 & 0.034 & 0.425 & $2.11{\times}10^{-4}$ & 0.0217 & 0.984 & 10.6 & 0.682 & 0.851 \\
\texttt{indoor/classroom\_well-lit\_trot} & 60 & 0.519 & 0.121 & 0.020 & 0.398 & $5.67{\times}10^{-4}$ & 0.0275 & 0.986 & 10.8 & 0.646 & 1.19 \\
\texttt{outdoor/between-buildings\_day\_comb} & 60 & 0.535 & 0.176 & 0.012 & 0.332 & $1.62{\times}10^{-4}$ & 0.0338 & 0.985 & 8.49 & 0.875 & 0.782 \\
\texttt{outdoor/between-buildings\_day\_trot} & 60 & 0.381 & 0.246 & 0.027 & 0.478 & $2.04{\times}10^{-5}$ & 0.0341 & 0.981 & 12.8 & 0.580 & 0.756 \\
\texttt{outdoor/between-buildings\_night\_comb} & 60 & 0.511 & 0.069 & 0.041 & 0.423 & $3.13{\times}10^{-5}$ & 0.0239 & 0.988 & 11.4 & 0.373 & 1.02 \\
\texttt{outdoor/between-buildings\_night\_trot} & 60 & 0.502 & 0.082 & 0.039 & 0.429 & $3.12{\times}10^{-5}$ & 0.0209 & 0.988 & 11.4 & 0.318 & 1.04 \\
\texttt{outdoor/grass1\_day\_comb} & 60 & 0.473 & 0.260 & 0.013 & 0.338 & $1.08{\times}10^{-4}$ & 0.0369 & 0.982 & 8.69 & 0.970 & 0.705 \\
\texttt{outdoor/grass1\_day\_trot} & 60 & 0.501 & 0.268 & 0.011 & 0.297 & $1.16{\times}10^{-4}$ & 0.0208 & 0.983 & 8.02 & 0.713 & 0.642 \\
\texttt{outdoor/grass1\_night\_comb} & 60 & 0.500 & 0.040 & 0.042 & 0.453 & $2.60{\times}10^{-4}$ & 0.0130 & 0.981 & 13.5 & 0.984 & 0.914 \\
\texttt{outdoor/grass1\_night\_trot} & 60 & 0.437 & 0.034 & 0.050 & 0.519 & $1.34{\times}10^{-4}$ & 0.0160 & 0.982 & 16.4 & 0.533 & 0.976 \\
\textit{Mean} & 960 & 0.439 & 0.187 & 0.034 & 0.450 & $1.89{\times}10^{-4}$ & 0.0282 & 0.983 & 12.8 & 0.790 & 0.836 \\
\bottomrule
\end{tabular*}
\end{table*}

\section{Additional Ablations}
\label{sec:supp-ablations}

\subsection{Sampler and Allocation Ablations}
\label{sec:supp-sampler-allocation}
These ablations vary one stage at a time around a few-step version of EventDiffusion-FT on DSEC test. Table~\ref{tab:abl_ead_sampler} samples EAD at 2 to 32 steps, with ESA and ETD fixed. Table~\ref{tab:abl_etd_sampler} runs ETD at 3 to 25 reverse steps, with the distilled EAD at its 8 steps and ESA fixed. The two tables share their default row.
The EAD comparison shows how the activity sampler affects the complete stream. Increasing the number of steps from 2 to 16 brings the generated-to-real count ratio from $0.932$ to $0.996$, while 32 steps produce a ratio of $1.039$. CEFS and $D_{\rm count}$ improve across the tested settings. Fano error falls from 1.234 to 1.170 even though ETD is fixed, showing that the upstream activity affects the timing statistics of the completed stream. EAD time increases by about $0.16\,$s per step.

The ETD comparison provides a separate control over the cost of timestamp generation. Increasing the number of steps from 3 to 25 lowers Fano error from 1.359 to 1.119, while ETD runtime rises from 2.8 to 13.1\,s per interval. CEFS changes from 0.701 to 0.697 and Same-ts changes by about 0.0002, so the Fano error provides the clearest distinction among these timing settings. The default of six ETD steps balances this timing result with runtime. With the default settings, the few-step pipeline reaches CEFS 0.700 and Fano error 1.217. EventDiffusion-FT at its full settings (Sec.~\ref{sec:supp-method}) reaches 0.684 and 1.234 on the same windows over three seeds, in about four times the time per interval.
\begin{table}[tbp]
\centering\footnotesize
\setlength{\tabcolsep}{5pt}
\caption{Effect of varying the distilled EAD's sampler steps on DSEC test (8 steps in training), with ESA and ETD fixed. Values are means over three seeds (std at most 0.001 for CEFS and 0.003 for Fano). $N_g/N_r$: generated over real event count. Fano: the timing error of Table~3 of the main paper. Time: EAD seconds per frame interval on an RTX 3090. Underlined: default.}
\label{tab:abl_ead_sampler}
\begin{tabular}{@{}lccccc@{}}
\toprule
\textbf{Steps} & \textbf{CEFS}$\uparrow$ & $D_{\rm count}$$\downarrow$ & $N_g/N_r$ & \textbf{Fano}$\downarrow$ & \textbf{Time (s)} \\
\midrule
2  & 0.695 & 0.049 & 0.932 & 1.234 & 0.26 \\
4  & 0.698 & 0.047 & 0.950 & 1.223 & 0.65 \\
\underline{8}  & 0.700 & 0.044 & 0.964 & 1.217 & 1.31 \\
16 & 0.705 & 0.039 & 0.996 & 1.198 & 2.54 \\
32 & 0.711 & 0.035 & 1.039 & 1.170 & 5.14 \\
\bottomrule
\end{tabular}
\end{table}

\begin{table}[tbp]
\centering\footnotesize
\setlength{\tabcolsep}{5pt}
\caption{Effect of varying ETD sampler steps on the full pipeline on DSEC test, with the 8-step distilled EAD and ESA fixed. Values are means over three seeds (std at most 0.001 for CEFS and 0.005 for Fano). Same-ts and Fano are the timing errors in Table~3 of the main paper. ISI-spike is omitted because it is $3.03{\times}10^{-6}$ in every row, as nearly all gaps on DSEC are under 10\,$\mu$s. Time: ETD seconds per frame interval on an RTX 3090. Underlined: default.}
\label{tab:abl_etd_sampler}
\begin{tabular}{@{}lcccc@{}}
\toprule
\textbf{Steps} & \textbf{CEFS}$\uparrow$ & \textbf{Same-ts}$\downarrow$ & \textbf{Fano}$\downarrow$ & \textbf{Time (s)} \\
\midrule
3  & 0.701 & 0.0126 & 1.359 & 2.83 \\
4  & 0.701 & 0.0126 & 1.290 & 3.35 \\
\underline{6}  & 0.700 & 0.0125 & 1.217 & 4.43 \\
12 & 0.698 & 0.0125 & 1.143 & 7.10 \\
25 & 0.697 & 0.0124 & 1.119 & 13.11 \\
\bottomrule
\end{tabular}
\end{table}

Table~\ref{tab:abl_esa_allocation} compares three ways of allocating each cell's events to its pixels. ESA is given each slice's real cell counts, so only the allocation differs, and ETD times the result as in the pipeline. The even spread places the events uniformly over the cell's pixels. The binomial split is ESA trained with $\kappa=1$ in every split (the Conway--Maxwell-binomial of Eq.~(5) of the main paper reduced to a binomial), from the same initialization and with the same data and schedule as EventDiffusion-FT's ESA. The learned dispersion is that ESA. The last row runs the full pipeline, with the distilled EAD's counts in place of the real ones.
Because every allocation starts from the same real cell counts, this comparison isolates how the events are distributed within cells. On the DSEC test windows, the learned CMB allocation achieves CEFS 0.8035, compared with 0.8029 for the learned binomial split and 0.7997 for the even spread. After ETD assigns timestamps, the completed stream's Fano error is 1.073 with CMB, compared with 1.118 for the binomial split and 1.116 for the even spread. We also evaluate allocation likelihood separately on held-out gaps from two DSEC test recordings. There, CMB saves 0.134 bits per event over the even spread, compared with 0.080 for the binomial split. These comparisons show how the allocation of a fixed event budget affects spatial fidelity and the timing statistics of the completed stream.
\begin{table}[tbp]
\centering\footnotesize
\setlength{\tabcolsep}{4pt}
\caption{ESA allocation on real coarse counts from DSEC test, with ETD fixed. Values are means over three seeds. Standard deviations for $D_{\rm ST}$ and CEFS are below $10^{-4}$ on real counts and below 0.001 in the last row. Fano standard deviations are at most 0.004. Even spread allocates each cell's events uniformly over its pixels. Binomial dispersion learns the pixel shares with $\kappa=1$ in the CMB split. Learned dispersion also predicts $\kappa$. Bits saved: allocation log-likelihood gain per event over even spread, on held-out gaps of two DSEC test recordings. Fano: ETD's timing error of Table~3 of the main paper. The last row runs the full pipeline on the distilled EAD's counts.}
\label{tab:abl_esa_allocation}
\begin{tabularx}{\columnwidth}{@{}>{\raggedright\arraybackslash}Xcccc@{}}
\toprule
\textbf{Allocation}
& \shortstack{\textbf{Bits saved}\\per event $\uparrow$}
& \shortstack{$D_{\rm ST}$\\$\downarrow$}
& \shortstack{\textbf{CEFS}\\$\uparrow$}
& \shortstack{\textbf{Fano}\\$\downarrow$} \\
\midrule
Even spread                         & $0$   & 0.1992 & 0.7997 & 1.116 \\
Binomial dispersion ($\kappa=1$)    & 0.080 & 0.1960 & 0.8029 & 1.118 \\
Learned dispersion (CMB)            & 0.134 & 0.1955 & 0.8035 & 1.073 \\
\midrule
Learned dispersion, full pipeline   & 0.134 & 0.2426 & 0.6998 & 1.217 \\
\bottomrule
\end{tabularx}
\end{table}

\subsection{Row and Column Context in ETD}
\label{sec:supp-rowcol}
Table~\ref{tab:supp_rowcol} repeats the row/column comparison of Table~6 of the main paper on three more test sets, with the same protocol. The context lowers the Fano error on EvSLAM, on all six of its test streams, and on EDS, and leaves it unchanged on DSEC and CEAR. On EvSLAM it also lowers Same-ts.
\begin{table}[t]
\centering\scriptsize
\setlength{\tabcolsep}{3pt}
\caption{ETD with and without row/column context, real event marks, mean over three seeds. Brackets give the number of test streams. EvSLAM's six streams come from two cameras on each of three test recordings, with 30 windows per stream. Same-ts ${\times}10^3$.}
\label{tab:supp_rowcol}
\begin{tabular}{@{}llcc@{}}
\toprule
\textbf{Dataset} & \textbf{Row/col.}
& \shortstack{\textbf{Same-ts}$\downarrow$}
& \shortstack{\textbf{Fano}$\downarrow$} \\
\midrule
DSEC (12)   & with    & 1.17 & 1.07 \\
            & without & 1.15 & 1.07 \\
\midrule
EDS (2)     & with    & 9.84 & 0.296 \\
            & without & 9.73 & 0.397 \\
\midrule
CEAR (16)   & with    & 90.3 & 0.609 \\
            & without & 20.1 & 0.599 \\
\midrule
EvSLAM (6)  & with    & 36.2 & 1.30 \\
            & without & 57.7 & 1.70 \\
\bottomrule
\end{tabular}
\end{table}

\section{Qualitative Results}
\label{sec:supp-qualitative}
\begin{figure*}[tbp]
\centering
\includegraphics[width=\textwidth]{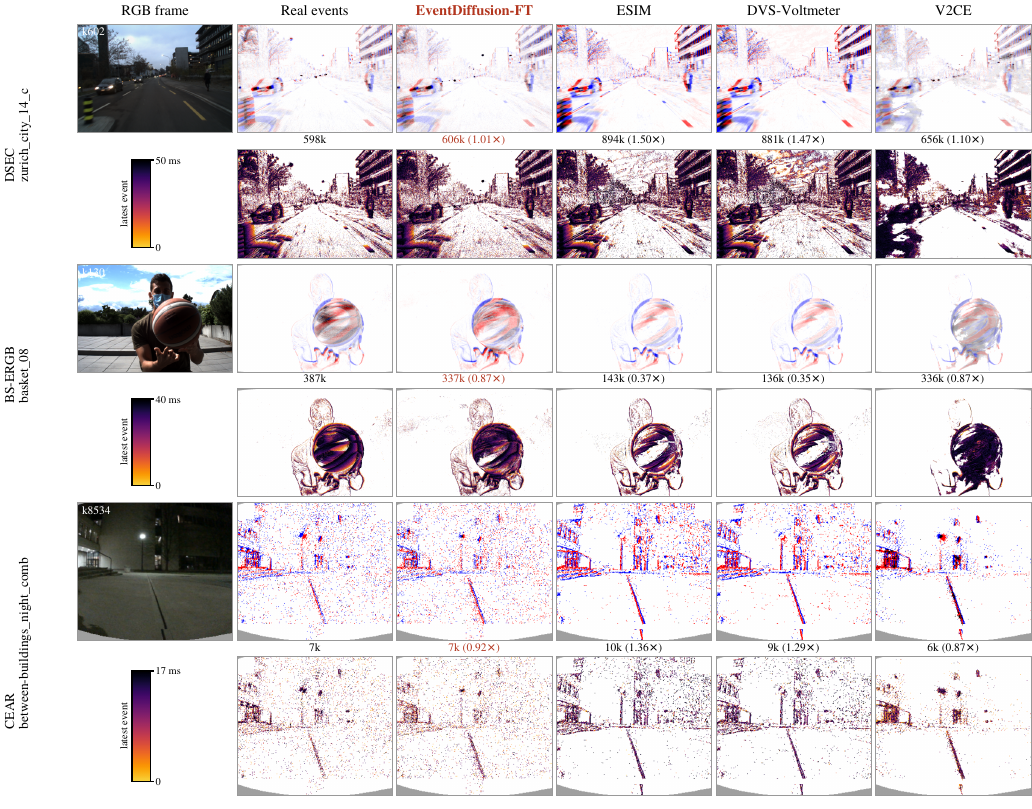}
\caption{Real events, EventDiffusion-FT and the baselines over one frame interval of three test scenes, with native RGB
frames. ESIM and DVS-Voltmeter run on SSM-10$\times$ frames at their calibrated thresholds and V2CE with the dataset's
fine-tuned model, as in Table~1 of the main paper. Upper row of each scene: ON (red) and OFF (blue) events, every tile
on the real tile's colour scale, with the event count and its ratio to the real count. Lower row: each pixel coloured
by the time of its latest event in the interval (white: none).}
\label{fig:supp_baselines}
\end{figure*}

\begin{figure*}[tbp]
\centering
\includegraphics[width=\textwidth]{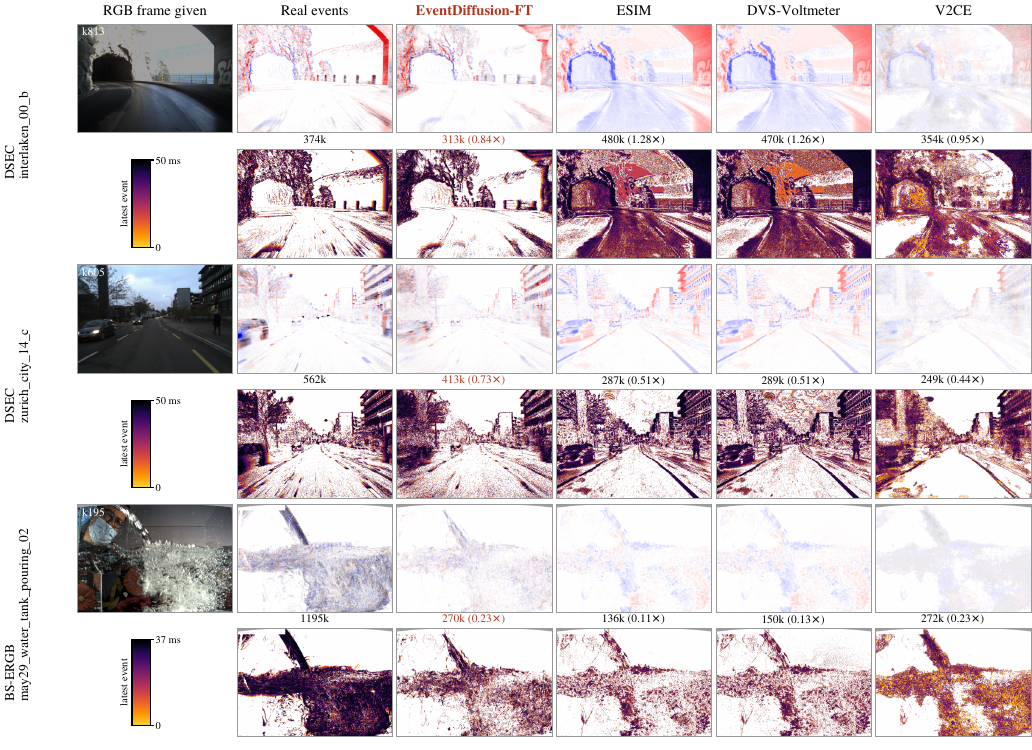}
\caption{Three test scenes under Gap-5$\times$, laid out as Fig.~\ref{fig:supp_baselines}: each over the middle
interval of a five-frame gap, from EventDiffusion-FT's Gap-5$\times$ model and from the baselines given only every fifth
frame (left: the frame given at the gap's start). ESIM and DVS-Voltmeter simulate between the given frames at their
thresholds calibrated on native frames, as in Table~2 of the main paper.}
\label{fig:supp_baselines_gap5}
\end{figure*}

\begin{figure*}[tp]
\centering
\includegraphics[width=\textwidth]{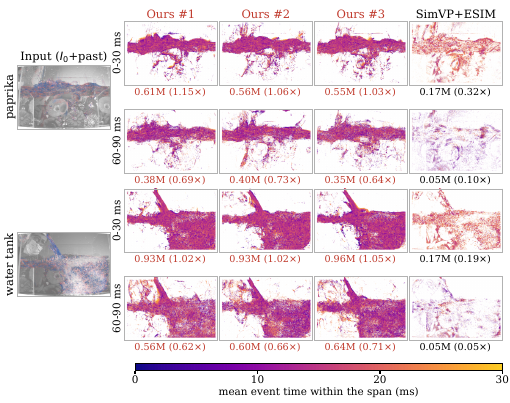}
\caption{Forecasting on two more BS-ERGB scenes, as in Fig.~4 of the main paper: from $I_0$ and the preceding events
(left), three EventDiffusion samples and SimVP+ESIM over the first 30\,ms and 60--90\,ms of the forecast. Pixels are
coloured by mean event time in the span (white indicates no events). Below each panel are the event count and its ratio to the real event count. The
two scenes have the lowest Chamfer distance (P-CD) of EventDiffusion's forecasts at the fifth interval among the 21
BS-ERGB test windows, averaged over the three samples.}
\label{fig:forecast_horizon}
\end{figure*}

Figure~\ref{fig:supp_baselines} shows one frame interval of three test scenes, from DSEC, BS-ERGB and CEAR, for the
real events, EventDiffusion-FT and the baselines with native RGB frames, and Fig.~\ref{fig:supp_baselines_gap5} three
scenes under Gap-5$\times$, each over the middle interval of a five-frame gap (its second row: the DSEC scene of
Fig.~\ref{fig:supp_baselines}, 0.25\,s later). The baselines use the calibrated
parameters and the per-dataset V2CE models of Tables~1 and~2 of the main paper. ESIM and DVS-Voltmeter run on
SSM-10$\times$ frames with native RGB and on the given frames under Gap-5$\times$. V2CE reads the native or the given
frames. Figure~\ref{fig:forecast_horizon} adds two forecasting scenes to Fig.~4 of the main paper.

\section{Additional CEFS Validation}
\label{supp:cefs_validation}

We evaluate the Composite Event Fidelity Score (CEFS) on controlled
perturbations of real event streams, compare it with geometric and learned
alternatives, and examine its relation to a fixed downstream event-perception
model. The metric definition, reference-derived multiscale weights, timestamp
terms, and aggregation coefficients are specified in Sec.~\ref{sec:supp-cefs}.
All comparisons use identical paired event arrays within each experiment, and the CEFS
parameters are held fixed.

\subsection{Sensitivity to Controlled Event Corruptions}
\label{supp:cefs_corruptions}
We first perturb seven real reference windows from MVSEC, the DAVIS
Event-Camera Dataset, and DSEC (three, three, and one recordings,
respectively), spanning sparse and dense event activity. The MVSEC/DAVIS
windows are 100\,ms, and the DSEC window is 5\,ms. The same generated
perturbations are used for every metric. No events are subsampled during
scoring. We test the eight corruption families defined in
Table~\ref{tab:supp-cefs-corruptions}, including the unmodified reference
as the identity case, using three corruption seeds.

\begin{figure*}[tp]
\centering
\includegraphics[width=\textwidth]{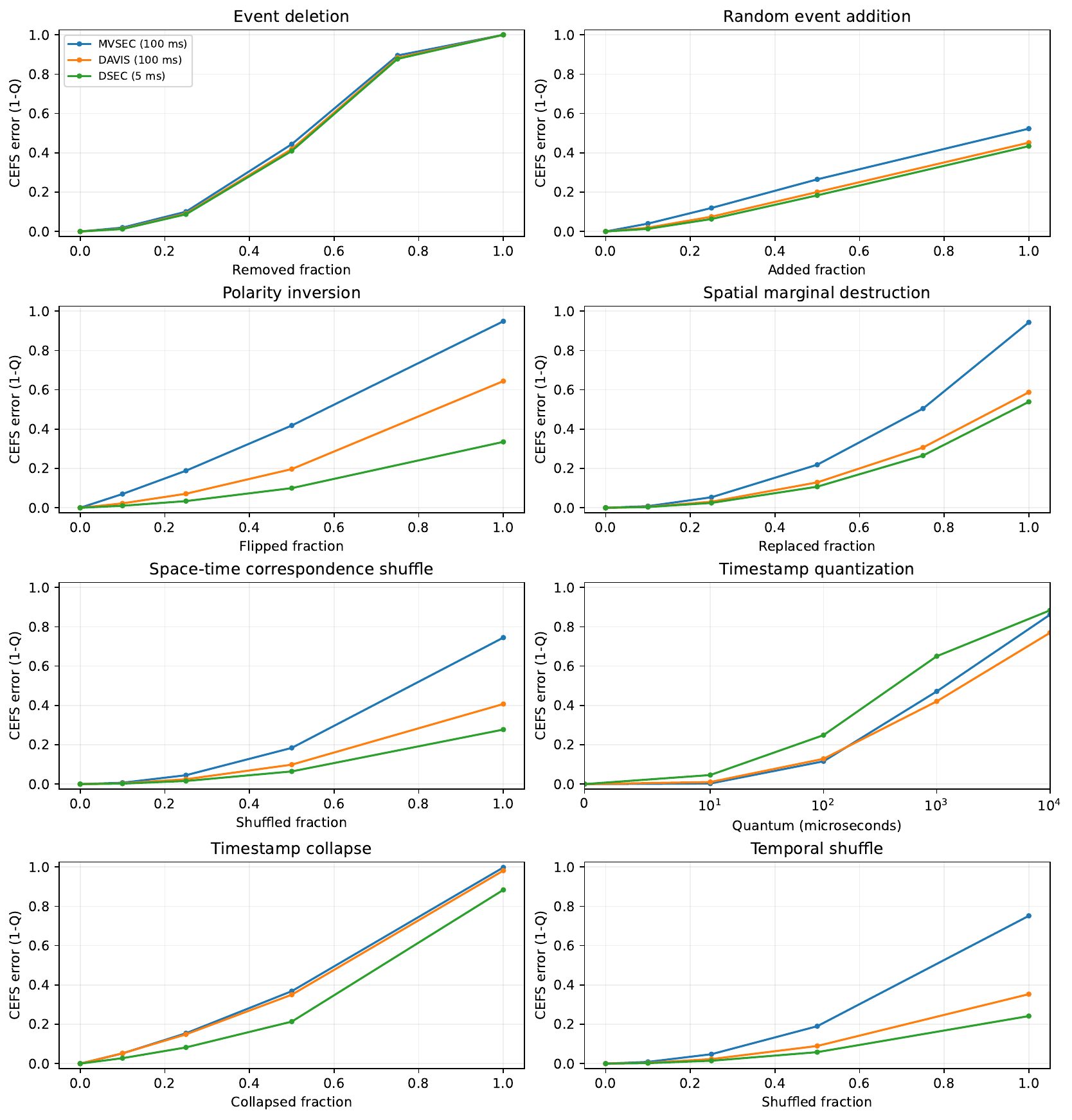}
\caption{CEFS error versus severity for the eight development corruption
families defined in Table~\ref{tab:supp-cefs-corruptions}. Curves are
summarized within recording before domain-level
averaging. Different source-window durations should not be interpreted as
matched-duration comparisons between sensors.}
\label{fig:cefs_corruption_compact}
\end{figure*}

Across these eight corruption families and seven development
references, CEFS error $1-\mathrm{CEFS}$ increases monotonically at every
tested severity (mean Spearman rank correlation $1$ up to rounding, with no
mean or individual-seed reversals). Table~\ref{tab:cefs_corruption_compact} and
Fig.~\ref{fig:cefs_corruption_compact} summarize the endpoint changes and
the full severity curves, respectively.
Complete deletion gives CEFS $0$ by construction. Polarity
inversion, spatial replacement, temporal quantization and timestamp
collapse produce substantial responses. Small predefined spatial and
temporal perturbations produce much smaller penalties than their respective
destructive endpoints: the maximum benign/destructive ratio is $0.0614$
on these development windows.

\begin{table}[t]
\centering\small
\setlength{\tabcolsep}{4pt}
\caption{Increase in CEFS error $1-\mathrm{CEFS}$ from identity to the
maximum tested severity for each of the eight families in
Fig.~\ref{fig:cefs_corruption_compact}, summarized over seven development recordings.
Ranges are across recordings, not confidence intervals. Different native
window durations preclude direct sensor rankings.}
\label{tab:cefs_corruption_compact}
\begin{tabularx}{\columnwidth}{@{}Xcc@{}}
\toprule
Corruption & Median error rise & Across-reference range \\
\midrule
Event deletion & 1.000 & 1.000--1.000 \\
Random addition & 0.462 & 0.434--0.555 \\
Polarity inversion & 0.823 & 0.335--0.996 \\
Spatial replacement & 0.844 & 0.386--0.971 \\
Spatial/temporal permutation & 0.495 & 0.277--0.881 \\
Timestamp quantization & 0.838 & 0.766--0.883 \\
Timestamp collapse & 0.994 & 0.883--0.999 \\
Temporal shuffle & 0.378 & 0.242--0.887 \\
\bottomrule
\end{tabularx}
\end{table}

\subsection{Comparison with Existing Fidelity Metrics}
\label{supp:cefs_comparators}
CEFS complements geometric and feature-based event comparisons. For the
same saved event arrays, the tested RVT-Small, ten-bin adaptation of EQS
scores an empty output at $0.937$--$0.967$, while CEFS scores it at zero.
Quantization to 10\,ms leaves this EQS adaptation unchanged on all 36
tested pairs, whereas CEFS error increases by $0.838$--$0.876$.
These experiments isolate the tested feature representation and do not
claim that every implementation of EQS behaves identically.

\begin{figure*}[tp]
\centering
\includegraphics[width=\textwidth]{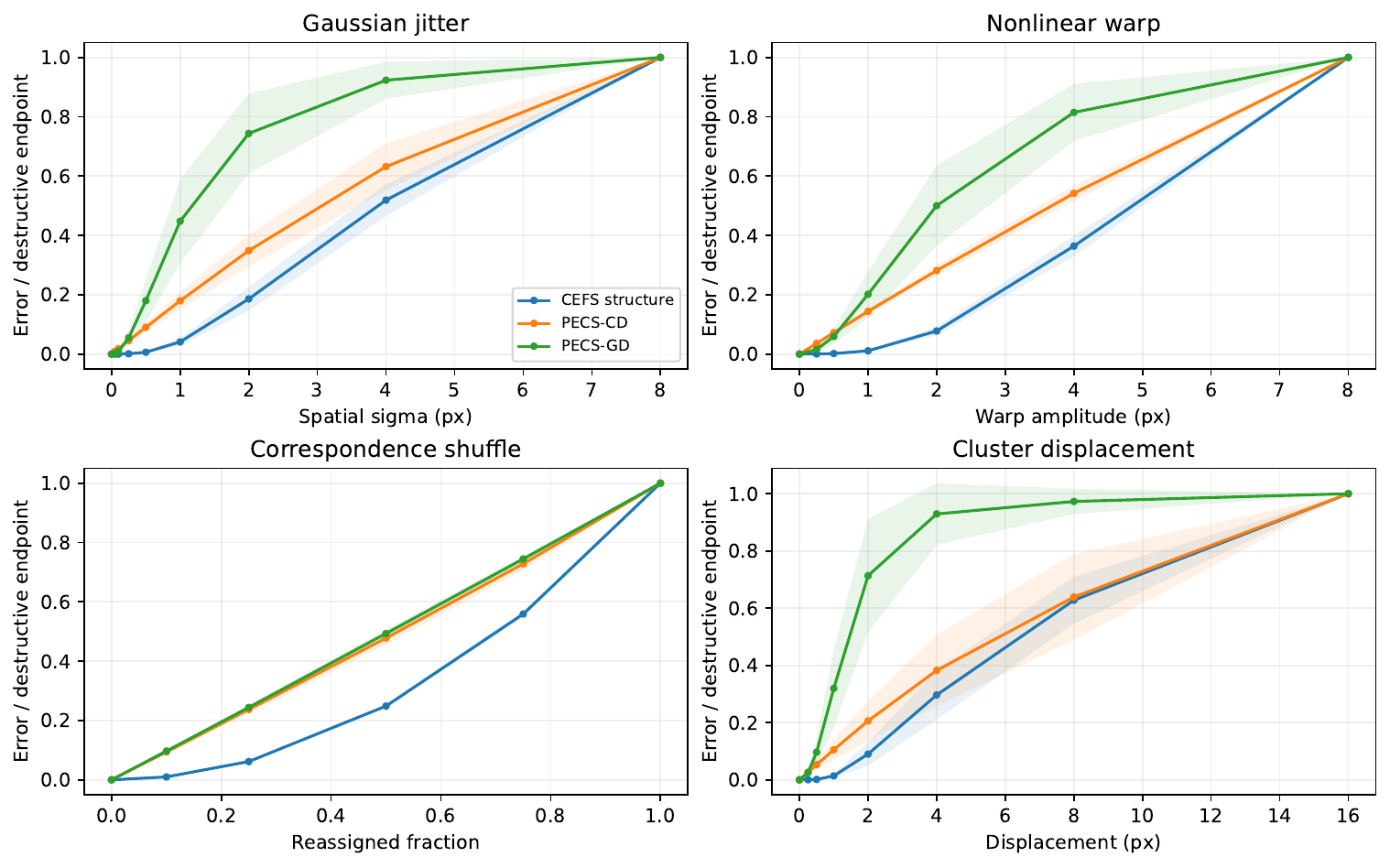}
\caption{Normalized structural error responses of CEFS, PECS Chamfer distance
(CD), and PECS Gaussian distance (GD) to geometric perturbations. Each
recording is normalized by its own predetermined destructive endpoint for
visual comparison. The underlying metrics are not redefined by this plot.}
\label{fig:cefs_geometry_compact}
\end{figure*}

\begin{figure*}[tp]
\centering
\includegraphics[width=\textwidth]{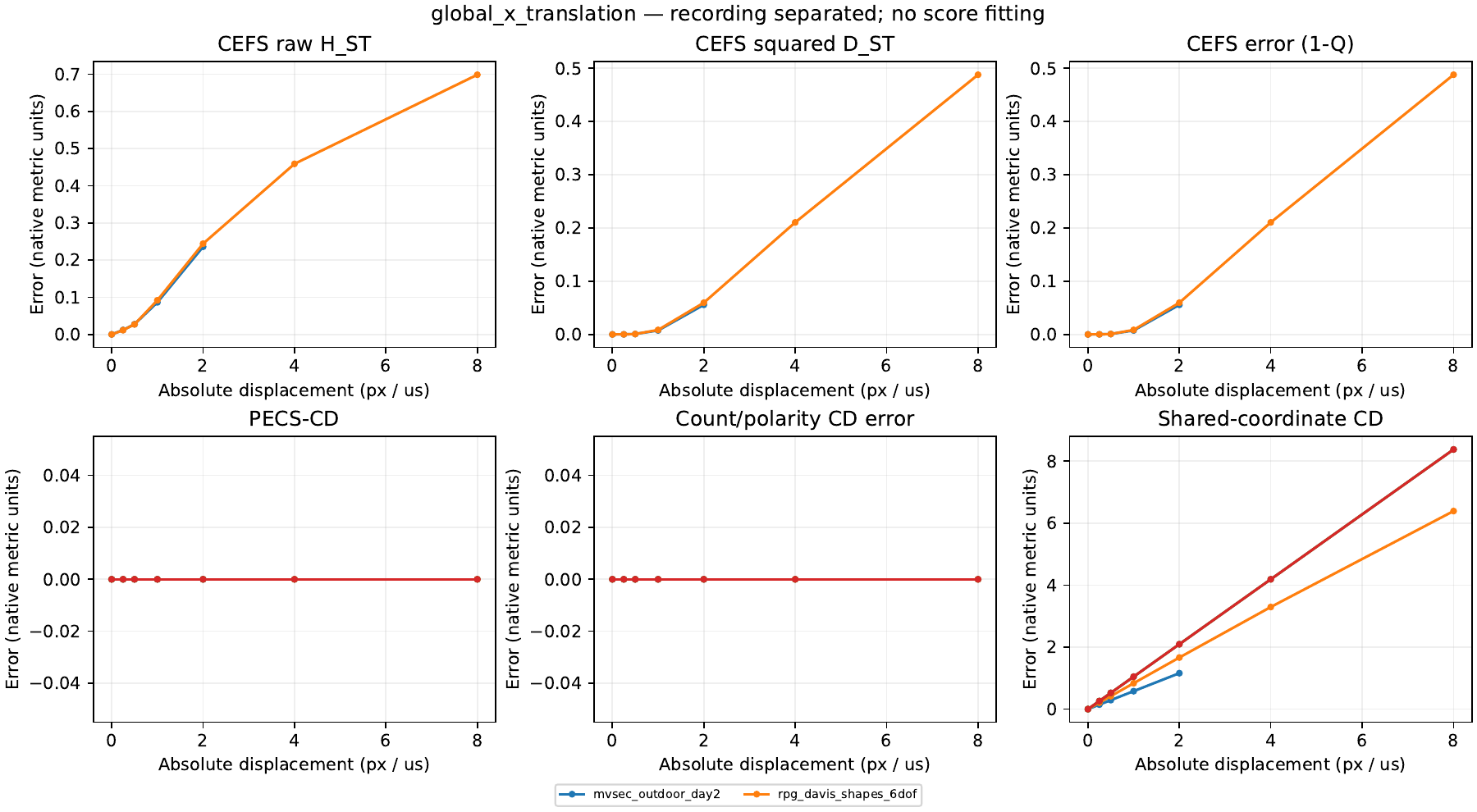}
\caption{Absolute spatial translations within the original sensor support.
The original PECS per-stream normalization removes a global shift, whereas
CEFS and shared-coordinate Chamfer detect it. Only geometrically valid
translations are included.}
\label{fig:cefs_alignment_compact}
\end{figure*}

\begin{figure*}[tp]
\centering
\includegraphics[width=\textwidth]{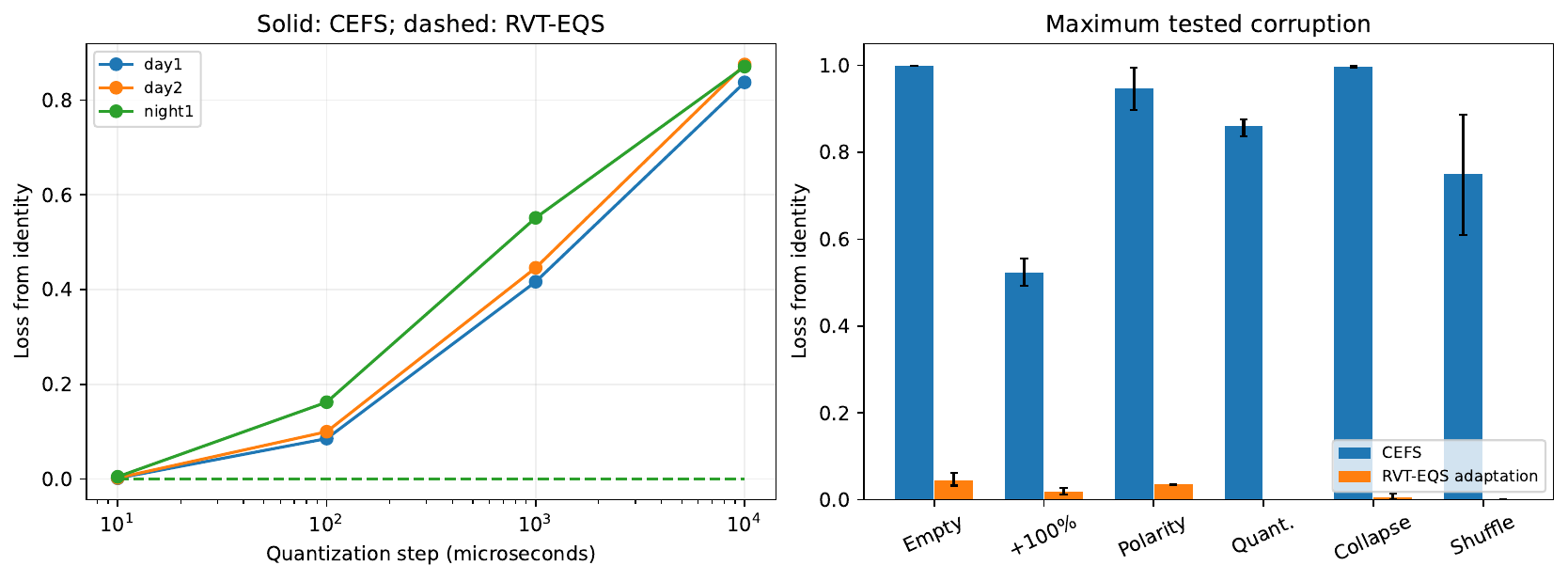}
\caption{CEFS and the tested RVT-based EQS adaptation evaluated on identical
real/corrupted event streams. CEFS responds to timestamp quantization and
other controlled errors that the tested temporally aggregated EQS features
can overlook.}
\label{fig:cefs_eqs_compact}
\end{figure*}

We also compare CEFS structure with PECS Chamfer (CD) and Gaussian (GD)
distances in a separate experiment with seven count-preserving geometric
perturbations and seven reference recordings. Both CEFS
structure and paper-equation CD achieve perfect tested benign-versus-
destructive rank separation (AUC $1$). CEFS's squared multiscale
structure term has lower benign/destructive error ratios than CD and GD
on these geometric perturbations, while CD remains sensitive to fine point
displacements. In the original PECS evaluation preprocessing, independent
per-stream min--max normalization can erase a global shift or delay:
all 25 physically valid strict alignment tests give CD $=0$ although
CEFS responds. A common-coordinate Chamfer implementation also detects
these shifts, identifying the preprocessing choice rather than a universal
limitation of nearest-neighbour distances. Figures~\ref{fig:cefs_geometry_compact}--\ref{fig:cefs_eqs_compact}
visualize the geometric and timestamp tests. CEFS therefore contributes a
single interpretable score spanning count, signed space--time structure,
and timing, not an across-the-board replacement for specialized metrics.

\paragraph{Excess-activity control.}
To isolate an event-count error from a large geometric displacement, we duplicate every event with a $1\,\mu$s timestamp offset. Across 36 DSEC windows, CEFS is $0.607$ on average, while PECS-CD and PECS-GD average $0.010$ and $0.0005$ and the tested EQS adaptation remains at $1.000$. The repeated events lie close to their references, so nearest-neighbour distances change little even though the event count doubles. The tested learned features also show little response in this example. The change is relevant to a downstream task: the LPIPS error of a real-event-trained E2VID increases in 34 of the 36 windows, by $0.094$ on average. This control illustrates why event activity should be evaluated alongside normalized structure.

\begin{figure*}[tp]
\centering
\includegraphics[width=\textwidth]{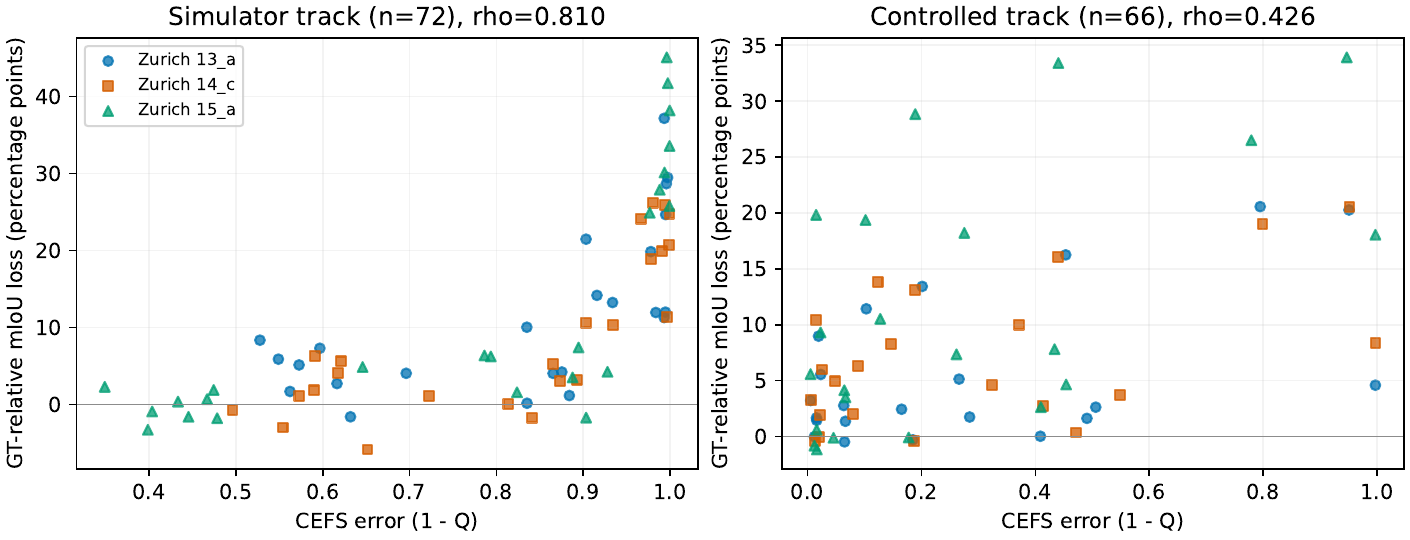}
\caption{Association between paired event-stream fidelity error and
GT-relative segmentation loss for the frozen ESS evaluation. The simulator
and controlled-corruption tracks are kept separate, and dependent windows
are not treated as independent recordings.}
\label{fig:cefs_task_scatter_compact}
\end{figure*}

\subsection{Validation on Additional Recordings}
\label{supp:cefs_extra_recordings}
After fixing the CEFS definition, we evaluated four additional DSEC
recordings (\texttt{interlaken\_01\_a},
\texttt{zurich\_city\_12\_a}, \texttt{zurich\_city\_14\_a}, and
\texttt{zurich\_city\_15\_a}) using fixed 5/20/50/100\,ms windows
and independent corruption seeds. Nineteen of the twenty planned windows
were executable. One 100\,ms high-density window exceeded a predefined
2.5-million-event execution ceiling and was recorded as unavailable,
not assigned a score. This experiment uses a broader suite of fourteen
corruption families, including spatial jitter and exact duplication.
At 50\,ms, all fourteen family summaries have Spearman correlation $1$
and zero severity reversals.
Further evaluations on an additional MVSEC motorcycle recording and a
subsequently selected DSEC recording also show monotone responses with the
fixed formulation. These tests broaden coverage without treating nested
windows or corruption seeds as independent physical recordings.

\subsection{Downstream Task Compatibility}
\label{supp:cefs_task}
To check whether CEFS is informative beyond synthetic perturbations, we
compare paired event-stream fidelity with predictions from a frozen ESS
event-based semantic segmenter on three DSEC-Semantic test recordings
(\texttt{zurich\_city\_13\_a}, \texttt{14\_c}, \texttt{15\_a}).
For each recording, eight fixed 50\,ms event windows are evaluated under
identical preprocessing. Simulator outputs are generated by ESIM,
DVS-Voltmeter and v2e. The segmentation checkpoint is not fine-tuned or
selected using CEFS. With the primary events-supervised checkpoint and
20-packet input policy, mean mIoU across recordings is $43.11\%$ for
real events, $42.10\%$ for ESIM, $36.23\%$ for DVS-Voltmeter and
$14.78\%$ for v2e. Their respective mean CEFS values are $1$,
$0.463$, $0.143$ and $0.00889$.

Across the 72 simulator/window pairs, CEFS error has a positive
Spearman association ($\rho=0.810$) with same-window segmentation loss.
For context, paper-equation PECS-CD obtains $0.771$, the RVT-EQS
adaptation $0.752$, and a count/polarity-aware CD control $0.814$.
Figure~\ref{fig:cefs_task_scatter_compact} visualizes the separate simulator
and corruption tracks. CEFS addresses both event activity and native space--time fidelity. In the duplication control, it detects the excess events that the tested PECS and EQS scores largely overlook, while E2VID reconstruction quality also deteriorates. In the downstream perturbation analysis, CEFS remains the strongest of the tested metrics for the timestamp-based CMax task even when its direct count factor is removed. Its mean correlation is 0.803, compared with 0.722 for PECS-GD, 0.666 for PECS-CD and 0.371 for the tested EQS adaptation (Table~\ref{tab:cefs_noncount_tasks}). This indicates that CEFS captures information relevant to native-event processing beyond event-count agreement. On the simulator-to-ESS comparison, CEFS and the strengthened count/polarity-aware Chamfer control have similar associations with segmentation loss (Table~\ref{tab:cefs_task_compact}).

\begin{table}[t]
\centering\small
\setlength{\tabcolsep}{5pt}
\caption{Simulator-track Spearman correlation between paired-fidelity
error and GT-relative frozen ESS mIoU loss ($72$ simulator/window pairs
from three recordings). Values are descriptive. Dependent windows do
not constitute 72 independent recordings.}
\label{tab:cefs_task_compact}
\begin{tabular}{lc}
\toprule
Fidelity metric & $\rho$ with task loss $\uparrow$ \\
\midrule
CEFS & 0.810 \\
PECS-CD & 0.771 \\
PECS-GD & 0.668 \\
RVT-EQS adaptation & 0.752 \\
Count/polarity-aware CD & 0.814 \\
\bottomrule
\end{tabular}
\end{table}

\paragraph{Downstream association beyond event count.}
The duplication experiment tests the activity component. We also evaluate what the other CEFS components contribute. Alongside the unmodified stream, we evaluate random polarity flips (probabilities $0.05$, $0.2$ and $0.5$), rightward spatial shifts (1, 4 and 16\,px), timestamp delays (1, 5 and 20\,ms), Gaussian timestamp jitter ($1\,$ms standard deviation), and timestamp shuffling within 50\,ms blocks. We compare the resulting fidelity errors with changes in four downstream tasks: E2VID reconstruction, ESS segmentation, E-RAFT flow and contrast-maximization flow (CMax). E2VID and ESS use our real-event-trained checkpoints. E-RAFT uses its released DSEC-trained checkpoint. These three models are frozen and use temporally aggregated representations. CMax estimates motion using individual event timestamps. Each task receives the same perturbed stream used for the fidelity metrics. To remove the direct contribution of count disagreement, we divide CEFS by its count factor and use one minus this non-count score as the fidelity error. Spatial shifts discard events outside the valid mask, and temporal perturbations can move events across window boundaries, so this is a count-factor ablation rather than an exactly count-matched experiment.

For each window we compute Spearman correlation across the selected perturbation conditions, and then average over windows. On CMax, the non-count CEFS error has mean correlation $0.803$, compared with $0.666$ for PECS-CD, $0.722$ for PECS-GD and $0.371$ for the tested EQS adaptation. CMax degradation is measured against its motion estimates on the corresponding real events, rather than ground-truth flow. The result shows that CEFS's polarity and space--time components can be informative about a timestamp-dependent task even when its direct count factor is removed. The tested EQS adaptation gives stronger associations on E2VID and E-RAFT, illustrating that learned feature and native-stream comparisons can capture different aspects of downstream behaviour. These selected conditions are a follow-up analysis of 36 windows from nine DSEC recordings. Windows from the same recording are not independent recordings.

\begin{table}[t]
\centering\small
\setlength{\tabcolsep}{3pt}
\caption{Mean within-window Spearman correlation between fidelity error and downstream degradation across selected perturbations. E2VID and ESS use real-event-trained checkpoints. E-RAFT uses its released DSEC-trained checkpoint. E2VID, E-RAFT and CMax use 36 windows from nine DSEC recordings. ESS uses the four windows with segmentation labels. The CEFS calculation excludes the count factor. Some perturbations change event counts at the scored boundaries. CMax and E-RAFT errors are relative to outputs on the real stream.}
\label{tab:cefs_noncount_tasks}
\begin{tabular}{@{}lrrrr@{}}
\toprule
Fidelity error & E2VID & ESS & E-RAFT & CMax \\
\midrule
CEFS, non-count & 0.249 & 0.243 & 0.490 & \textbf{0.803} \\
PECS-CD & 0.002 & $-$0.049 & 0.257 & 0.666 \\
PECS-GD & 0.043 & 0.031 & 0.303 & 0.722 \\
EQS adaptation & \textbf{0.868} & \textbf{0.736} & \textbf{0.943} & 0.371 \\
\bottomrule
\end{tabular}
\end{table}

\end{document}